\documentclass{article}

\usepackage[preprint,nonatbib]{neurips_2021}
    
\usepackage[usenames,dvipsnames]{xcolor}
\usepackage{subfig}

\usepackage[utf8]{inputenc} % allow utf-8 input
\usepackage[T1]{fontenc}    % use 8-bit T1 fonts
\usepackage{hyperref}       % hyperlinks
\usepackage{url}            % simple URL typesetting
\usepackage{booktabs}       % professional-quality tables
\usepackage{amsfonts}       % blackboard math symbols
\usepackage{nicefrac}       % compact symbols for 1/2, etc.
\usepackage{todonotes}
\usepackage{multicol}
\usepackage{multirow}
\usepackage{comment}
\usepackage{wrapfig}
\usepackage{acronym}
\acrodef{IC}{integrated circuit}
\acrodef{EDA}{electronic design automation}
\acrodef{HDL}{hardware description language}
\acrodef{AIG}{and-inverter-graph}
\acrodef{RL}{reinforcement learning}
\acrodef{ML}{machine learning}
\acrodef{IP}{intellectual property}
\acrodef{RTL}{register transfer level}
\acrodef{DAG}{directed acyclic graph}
\acrodef{GCN}{graph convolutional network}

\newcommand{\rt}[1]{\textcolor{red}{#1}}
\newcommand{\gt}[1]{{\color{NavyBlue}{#1}}}
\newcommand{\vt}[1]{{\color{Fuchsia}{#1}}}
\newcommand{\bt}[1]{{\color{Sepia}{#1}}}
\newcommand{\mat}[1]{{\color{PineGreen}{#1}}}
\newcommand{\bst}[1]{{\color{Bittersweet}{#1}}}
\newcommand{\fgt}[1]{{\color{WildStrawberry}{#1}}}

\usepackage{graphicx}
\usepackage[export]{adjustbox}
\usepackage{pdfpages}
\def\hlinewd#1{%
\noalign{\ifnum0=`}\fi\hrule \@height #1 %
\futurelet\reserved@a\@xhline} 

\title{OpenABC-D: A Large-Scale Dataset For Machine Learning Guided Integrated Circuit Synthesis}

\author{%
  Animesh Basak Chowdhury\\%\thanks{Github links} \\
  New York University\\
  \And
   Benjamin Tan \\
   New York University\\
  \And
  Ramesh Karri \\
  New York University\\
  \And
  Siddharth Garg \\
  New York University\\
}

\begin{document}

\maketitle

\begin{abstract}
%Modern integrated circuit (IC) design is heavily reliant on electronic design automation (EDA) algorithms --- software tools that transform a high-level specification into a physical IC layout ready for fabrication. The first step in this process, called circuit (or logic) synthesis, transforms a specification described in a hardware programming language like Verilog into an optimized digital circuit, a network of interconnected Boolean logic gates. 
Logic synthesis is a challenging and widely-researched combinatorial optimization problem during integrated circuit (IC) design. It transforms a high-level description of hardware in a programming language like Verilog into an optimized digital circuit netlist, a network of interconnected Boolean logic gates, that implements the function. Spurred by the success of ML in solving combinatorial and graph problems in other domains, there is growing interest in the design of ML-guided logic synthesis tools. Yet,  there are no standard datasets or prototypical learning tasks defined for this problem domain. Here, we describe OpenABC-D,  a large-scale, labeled dataset produced by synthesizing open source designs with a  leading open-source logic synthesis tool and illustrate its use in developing, evaluating and benchmarking ML-guided logic synthesis. OpenABC-D has intermediate and final outputs in the form of 870,000 And-Inverter-Graphs (AIGs) produced from 1500 synthesis runs plus labels such as the optimized node counts, and delay. %longest path, area, and timing of the AIGs.
We define a generic learning problem on this dataset and benchmark existing 
solutions for it. The codes related to dataset creation and benchmark models are available at \url{https://github.com/NYU-MLDA/OpenABC.git}. The dataset generated is available at \url{https://archive.nyu.edu/handle/2451/63311}.

\end{abstract}

\section{Introduction}
\label{sec:intro}
 
Complex \acp{IC} can have over a billion transistors making hand-design impossible. 
Hence, the \ac{IC} industry relies on \ac{EDA} tools that progressively transform a high-level hardware into a layout ready for \ac{IC} fabrication. 
EDA tools let designers focus on describing function at a high-level using a \ac{HDL} like Verilog, without worrying about low-level implementation of the \ac{IC}. 
Increasing design complexity and scalability challenges in the design flow has raised interest in \ac{ML} for \ac{EDA}~\cite{huang_machine_2021}. 

The first step in \ac{EDA} is logic synthesis. Logic synthesis transforms an \ac{HDL} program into a functionally equivalent graph (netlist) of Boolean logic gates while attempting to minimize metrics such as area, power, and delay. Since logic synthesis is the first in a sequence of \ac{EDA} steps that yields the final \ac{IC} layout, the quality of its output impacts the size, power, and speed of the final \ac{IC}. Even the simplest version of this problem, logic minimization, is $\Sigma^{2}_{p}$-Hard~\cite{umans1999hardness,buchfuhrer2011complexity}.\footnote{ $\Sigma^{2}_{p}$-Hard problems are hard even with access to an oracle solver for NP-complete problems.} Commercial logic synthesis tools use heuristics developed  by academia and industry~\cite{amaru2017logic}.
%https://users.cs.duke.edu/~reif/courses/complectures/books/AB/phchap.png
%Optimal logic synthesis has been shown to be an $\mathcal{NP}^\mathcal{NP}$-Hard combinatorial optimization problem
%and is one of the most fundamental and well-studied problems in \ac{EDA}~\cite{}.
%In that sense, circuit/logic synthesis is akin to software compilation, allowing designers to use high-level  
%https://ieeexplore.ieee.org/abstract/document/1270347?casa_token=AoMcstu-fQYAAAAA:K5bVJietP2s4zeGsuYw3eE28EOaEyQzqq4B9-fcMen8dmXN11UA8f-n7lpJRbAE4184i2dotrbP6
%https://ieeexplore.ieee.org/abstract/document/52213?casa_token=-cb8o-KZoxIAAAAA:dbKLxqcMgWm806Ineg0Z0Q8Ogf25DvLIhwk7G3SFBHzGqp2B0uY387TuqNsfnLohYZzMdlfLKF43
State-of-the-art in logic synthesis applies a \emph{sequence} of logic minimization heuristics to  transform a sum-of-products (SOP) or an \ac{AIG} representation. Common heuristics remove redundant nodes, refactor Boolean formulas, and simplify node representations. The \emph{order} in which these heuristics are applied -- the \emph{synthesis recipe} -- is critical to the quality of results. Designers use synthesis recipes that %have become part of folklore because they 
either work well for a range of inputs or have to hand-tune them by trial-and-error.  
%https://ieeexplore.ieee.org/stamp/stamp.jsp?arnumber=1688855&casa_token=kl0WyfWhlVsAAAAA:duCwDtrh_Qw1vcg5kghypjZLZGWiu1J8R4GClC36coR1qHL8DHCmnYVEVet3gNSlOzSPtUjRSA&tag=1

%The aim is to create an optimized gate-level representation of the design function that satisfy constraints on area, delay and performance. Algorithmic scaling challenges arise when applying logic synthesis on industrial designs. In a typical industrial setting, domain experts create logic synthesis scripts --sequences of optimization heuristics-- based on prior experience targeting classes of hardware designs.

%\begin{wrapfigure}{r}{0.8\textwidth}
\begin{figure}
\centering
%\vspace*{-0.2in}
\includegraphics[width=0.8\textwidth]{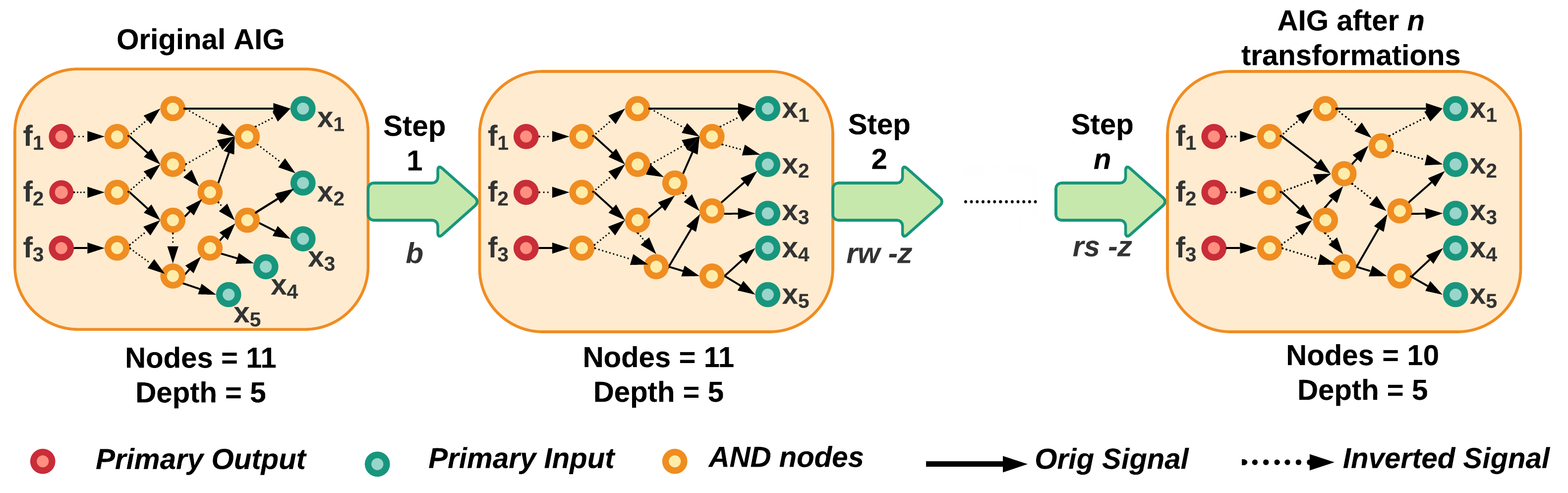}
\caption{Logic synthesis optimizations on And-Inverter Graphs (AIG).}
\label{fig:synthesisflow}
%\end{wrapfigure}
\end{figure}

The success of deep learning methods in solving a range of combinatorial and graph problems has spurred interest in \ac{ML}-guided logic synthesis~\cite{cunxi_dac,drills,firstWorkDL_synth,lsoracle,cunxi_iccad,mlcad_abc}. 
%For example, logic synthesis can be modelled as a Markov decision process; \ac{RL} can be used to generate a synthesis flow customized for every IP.Recent work \cite{firstWorkDL_synth,drills,mlcad_abc} has developed ``intelligent'' synthesis recipes using deep \ac{RL} agents. 
However, they report results on small datasets and solve different versions of the problem. As a consequence, benchmarking and comparing SoTA solutions, especially on "real-world" designs, is challenging. This is because there is no comprehensive, labeled dataset of publicly available and prototype problems that can serve as benchmarks.

%with a few designs, and the performance of such \ac{ML}/RL algorithms are unclear on large, ``real-world'' designs. This is because there is a lack of a comprehensive, labelled dataset of publicly available, realistic designs. 
%So far, the \ac{EDA} community has not created/standardized a high-quality dataset with which to train or evaluate \ac{ML} models for problems related to logic synthesis. 
%using which one can generate custom, domain-specific, pre-trained models for a variety of tasks. 
%{\bf FIXME: so what happened to these attempts; if they failed why did they fail?} No point in failure as there has not been any attempt. 
%Such models can later be used to solve open problems in logic synthesis. 
This motivates us to create a realistic and feature-rich dataset for logic synthesis that is of interest to both \ac{EDA} researchers (as the first large-scale dataset in this application domain) and \ac{ML} researchers (who could be interested dealing with a specialized domain of structured graph data).
Our work presents a dataset of 29 open source designs with 870,000 data samples. The \ac{EDA} community can use this data to train \ac{ML} models for a range of logic synthesis optimization tasks. This dataset can augment data for other problems in \ac{EDA} and hopefully be useful to the graph datasets community.

\label{sec:background}

%\begin{figure}[t]

\subsection{Overview of Logic Synthesis}
\label{subsec:background-overview}
A digital hardware \ac{IP} block is designed using an \ac{HDL} like Verilog or VHDL. Specifying functionality at this abstraction is typically called behavior-level or \ac{RTL} design. Logic synthesis %is the next step which 
takes an RTL implementation %of the hardware block 
and outputs an (optimized) gate-level netlist representation of the design % followed by technology mapping using a standard cell library. 
that can be mapped to a standard cell library (i.e., technology mapping). 
To meet a designer-specified area and delay overhead, % of the synthesized design, 
optimization takes place before and after technology mapping. In this work, we focus on technology independent logic optimization. This combinatorial optimization problem uses simple Boolean logic gates. 

A gate-level netlist is a Boolean function with binary-valued inputs and uses logical operations like AND, NOT, and XOR. This netlist can be represented canonically (e.g., as a truth table) % sum-of-products, binary decision diagrams, 
or in other formats, like \acp{AIG} and majority-inverter graphs (MIG). \ac{AIG} is a \ac{DAG} representation with 2-input AND function (nodes) and NOT function (dotted edges). %Out of well known representation techniques, 
The \ac{AIG} is popular since it scales and can compactly represent industrial-sized designs; %. AIG 
it is used in the state-of-the-art open source logic synthesis tool, ABC~\cite{abc}. \ac{AIG} allows structural optimizations like cut enumeration, Boolean implication and \ac{DAG}-based heuristics. 
% We describe fundamental sub-graph optimizations supported by ABC used as sequence of steps during logic minimization.
The following are fundamental sub-graph optimizations supported by ABC (ABC's commands are in parentheses): 

%\begin{enumerate}
\noindent\textbf{1. Balance (b)} is a depth-optimization step to minimize the delay of a design. Given an \ac{AIG} representation in the form of a \ac{DAG}, \textit{balance} applies tree-balancing transformations using associative and commutative properties on logic function. \\ 
\noindent\textbf{2. Rewrite (rw, rw -z)} is  \ac{DAG}-aware logic rewriting heuristic that does template pattern matching on sub-trees and replaces them with equivalent logic functions. Rewrite heuristic algorithm uses a k-way cut enumeration (k varies from 4-8) with the objective of finding an optimized representation of the sub-tree. In ABC, zero-cost variant (rw -z) does not  immediately reduce the number of nodes of the \ac{DAG}. The transformed structure can be optimized using other heuristics. \\
\noindent\textbf{3. Refactor (rf, rf -z)} can potentially change a large part of the netlist without caring about logic sharing. The method traverses iteratively on all nodes in the netlist, computes maximum fan out free cones and replaces them with equivalent functions if it improves cost (e.g., reduce number of nodes). Zero cost variant is available. \\
\noindent\textbf{4. Re-substitution (rs, rs -z)} optimizes by representing the logical function of a node using logic functions of existing nodes. Typically, \textit{k} nodes are introduced to represent the function and compared against the number of redundant nodes that are no longer required. %Positive cost reduces the number of nodes. 
k determines the size of the sub-circuit that can be replaced. Re-substitution improves logic sharing. 

%\end{enumerate}

Logic synthesis of a gate-level netlist is a sequential decision process applying sub-graph optimization heuristics in a non-trivial combination to obtain an optimized design. 
We term the fundamental sub-graph level optimizations as \textit{synthesis transformations}. \ac{IC} designers develop a sequence of synthesis transformations to get an optimized netlist. Technology mapping then results in a circuit satisfying quality of result (QoR) in terms of area, delay, and power consumption. We call sequences of synthesis transformations a \textit{synthesis recipe}. The objective of one  recipe is to reduce the number of nodes and depth of the \ac{DAG} network to directly correlate to minimizing area and delay~\cite{lsoracle}.

\subsection{Motivation} %existing problems}
\label{subsec:background-related}
\noindent{\bf Problem statement:} Given a gate-level netlist as \ac{AIG} graph representing a set of boolean functionalities, determine the sequence of sub-graph optimization steps generating the optimal \ac{AIG} representing same functionalities. The computational complexity of the problem is $\Sigma^{2}_{p}$-Hard.

Success of \ac{ML} algorithms has prompted researchers to re-examine logic synthesis, where most of the fundamental research happened in 1990s and 2000s~\cite{brayton1987mis,brayton1990multilevel,dag_mischenko}. 
The overarching question is: \textit{can past experience lead to informed decision-making for future problem instances?} 
So far, \ac{EDA} engineers use years of experience from past synthesis runs to intuit a good synthesis recipe for new \acp{IP}. To scale up designs and enable faster design sign-off, researchers formulate learning tasks in the logic synthesis domain and propose \ac{ML} algorithms to solve them, e.g., \\
\textbf{To predict synthesis recipe quality}: \cite{cunxi_dac} proposes a classification model to determine ``angel'' and ``devil'' synthesis recipes, i.e., identify if a recipe will generate a good quality design for an IP by training a model on data from a few synthesis runs. \\
\textbf{To predict the ``best'' optimizer}:  \cite{lsoracle} proposes an \ac{ML} model that identifies how a sub-circuit should be represented (AIG/MIG) to  match sub-circuit and synthesis recipes. % to use to optimize a subgraph from the entire \ac{AIG}. 
The work demonstrated state-of-the-art results compared to baseline ABC synthesis results.  \\
\textbf{To predict the best synthesis recipes (\ac{RL}-guided)}: Recent work~\cite{firstWorkDL_synth,drills,mlcad_abc,cunxi_iccad} formulates the problem as a Markov Decision Process (MDP), where  the future \ac{AIG} depends on current \ac{AIG} (the state) and synthesis transformation (the action). Past synthesis transformations do not impact the transition to new \ac{AIG} state. 
The state-action transition yields a deterministic result for an action.  However for the agent to be effective across diverse hardware IPs, it needs to explore a wide range of state-action pairs before being used in "exploitation" phase on unseen data. 

% All such works exhibit the potential of \ac{ML} in synthesis.
% \subsection{Motivation}
\label{subsec:background_motivation}
% A recent promising work~\cite{googlerl} using \ac{RL} for chip design have shown a wide range of  opportunity available for applying \ac{ML} in design automation problems. However qualitative \ac{ML} research can only be possible with the availability of qualitative dataset. 

While promising, prior work suffers from the inappropriateness of an 
% Currently, there is no standardized \ac{EDA} dataset on which various \ac{ML} techniques for a particular problem can be evaluated. This hurts the progress of \ac{ML} research in the area since there are no 
apples-to-apples comparison of the various proposed techniques %. Recent line  of works~\cite{firstWorkDL_synth,cunxi_dac,lsoracle,cunxi_iccad,mlcad_abc,drills} have used \ac{ML} for problems in logic synthesis; and claims achieving state-of-the-art performance. However, lack of a 
in the absence of a standardized dataset, configurations and benchmarks. 
We observed that prior work is evaluated on different data sets, making it difficult for researchers to understand the effectiveness of each approach (\autoref{sec:priorSotabenchmarks}). 
Additionally, there is no clear explanation on the choice of benchmarks considered for experiments. %leading to an important question: \textit{Are the performance of proposed \ac{ML} models consistent amongst diverse set of benchmarks?}
% for comparing \ac{ML} techniques  is hurting the progress of \ac{ML} in the area.
To address this shortcoming, %and establish a standardized method in the domain akin to Image and Natural language processing, 
we thus present a three-fold contribution: \\
%\begin{enumerate}
\noindent\textbf{1. OpenABC-Dataset (OpenABC-D)}: We release  OpenABC-D, a large-scale synthesis dataset comprising of 870,000 data samples by running 1500 synthesis recipes on 29 open source hardware IPs and preserving intermediate stage and final \ac{AIG}s. \\
\noindent\textbf{2. Data Generation Framework}: We provide an open source framework that can generate labeled data by performing synthesis runs on hardware IPs using different synthesis recipes. \\
\noindent\textbf{3. Benchmarking \ac{ML} models}: We benchmark the performance of simple graph models on our learning task using OpenABC-D. %(out of 4 described) 
%\end{enumerate}

%\todo[inline,color=red!20]{Logic synthesis as MDP}
%\todo[inline,color=red!20]{Logic synthesis using ABC}
%\todo[inline,color=red!20]{Steps of ABC: rewrite, resub, refactor etc (show diagram)}
%\todo[inline,color=red!20]{Placement}

% \section{Data generation pipeline}
% \todo[inline,color=red!20]{Lack of data in various tasks leading to over-fitting}
% \todo[inline,color=red!20]{Transferability of learned models from circuits accross various sizes}
% \todo[inline,color=red!20]{More opportunities for data-driven approaches}
% \todo[inline,color=red!20]{Who all are involved in creating of dataset}
\section{The Data Generation Pipeline}
\label{sec:datagen-pipeline}
\begin{figure}[t]
%\begin{wrapfigure}{r}{0.6\textwidth}
\centering
\vspace*{-0.3in}
\includegraphics[width=0.8\textwidth]{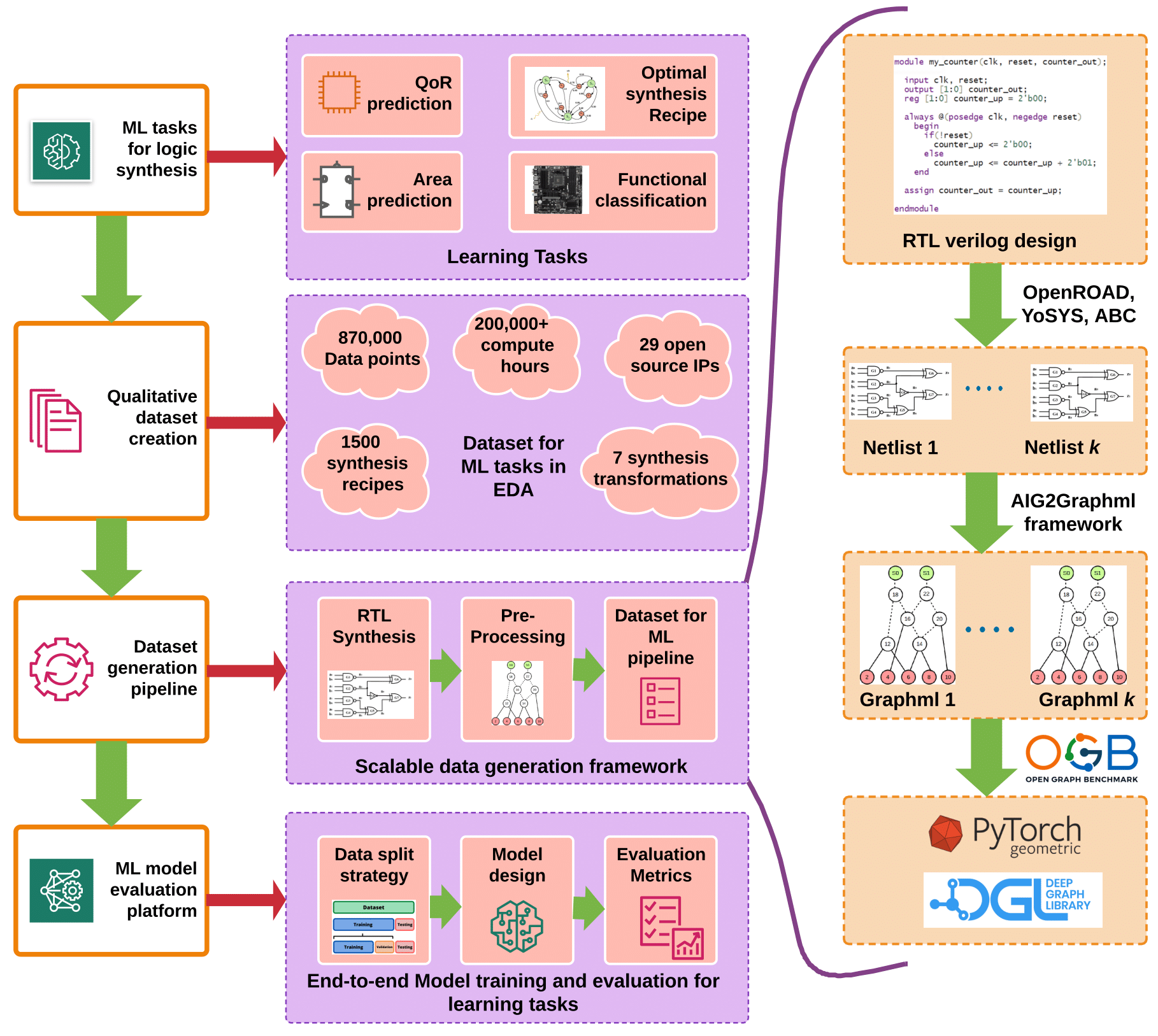}
\caption{OpenABC-D framework}
\label{fig:openabcdframework}
\end{figure}

For wider access, we focus on open source EDA platforms as key to advancing \ac{ML} research in EDA.
Thus, we %This motivated us to 
propose the OpenABC-D framework using open source EDA tools, thus making it freely available for anyone to generate data. 
% Note, however, the task is not simple and straightforward. 
Note, however, that this framework requires substantial compute hours to generate synthesis data, considerable preprocessing of intermediate-stage data, and converting it to compatible formats for applying \ac{ML} models. 
Thus, we applied the OpenABC-D framework on a set of open source IPs and make this data available to the community. 
We highlight our contributions and challenges for developing OpenABC-D (\autoref{fig:openabcdframework}): an end-to-end large-scale data generation framework for augmenting \ac{ML} research in circuit synthesis. 

\subsection{Open Source Tools}% and IPs} 
%In OpenABC-D, 
We use OpenROAD v1.0~\cite{ajayi2019openroad} EDA to perform logic synthesis; it uses 
% For logic synthesis, OpenROAD uses 
Yosys~\cite{Yosys} as the front-end engine (currently v0.9). 
Yosys performs logic synthesis in conjunction with ABC~\cite{abc}. It can generate a logic minimized netlist for a desired QoR. We use networkx v2.6 for graph processing. For \ac{ML} frameworks on graph structured data, we use pytorch v1.9 and pytorch-geometric v1.7.0. 
We collect area and timing of the AIG post-technology mapping using NanGate 45nm technology library and ``5K\_heavy'' wireload model. The dataset generation pipeline has three stages: (1) RTL synthesis, (2) Graph-level processing, and (3) Preprocessing for \ac{ML}. 
%Yosys performs sequential logic synthesis followed by ABC~\cite{abc} passes.
%Yosys used ABC as a backend engine for performing combinational circuit optimization and technology mapping using a pre-defined target cell library. Post combinational technology mapping, Yosys maps sequential elements to the flip-flops defined in the technology library and generates the final hardware design for further processing down the EDA stack. Our synthesis data generation framework has three steps 1) Register Transfer Level synthesis 2) Netlist pre-processing 3) Dataset generation for \ac{ML} pipeline. We have outlined the OpenABC-D framework in ~\autoref{fig:openabcdframework}.

%\end{wrapfigure}

\subsection{Register Transfer Level (RTL) Synthesis}
\label{subsec:datagen-rtlsynthesis}
In this stage, we take the specification of IPs (in Verilog/VHDL) and perform logic synthesis. % using Yosys v0.9 integrated in OpenROAD v1.0 framework. Yosys performs logic synthesis in conjunction with ABC~\cite{abc}. 
Yosys  performs optimization on the sequential part of the IP and passes on the combinational part to ABC for logic optimization and technology mapping. First, ABC structurally hashes the combinational design to create AIGs. 
Post-hashing, a user-provided synthesis recipe performs tech-independent optimization. 
Our framework allows two options: 1) automated synthesis script generation and 2) user-defined synthesis scripts. A synthesis script is a synthesis recipe with additional operations to save intermediate-stage AIGs. The AIGs generated by ABC are in the \texttt{BENCH} file format~\cite{brglez_combinational_1989}. %, a technique to represent a gate level netlist. 
The intermediate and final AIGs have different graph structures despite having the same function. Synthesis transformations affect parts of the graph differently, yielding a diverse collection of graph structured data. We prepared $K = 1500$ synthesis recipes each having $L = 20$ synthesis transformations.

%Each output circuit represent same IP functionality but different netlist structure obtained  by applying various transformations. Each circuit generated in the intermediate steps of synthesis flow represents impact of various synthesis transformations on circuits having different structural characteristics. In our work, we considered 1500 synthesis flows each having a step of 20 synthesis transformations.

%\autoref{fig:openabcdframework} shows a magnified view of data generation pipeline from RTL synthesis to processed data readily available to be used for \ac{ML} pipeline.  Yosys invokes ABC for combinational logic optimization and technology mapping. Initially, 

\subsection{Graph-Level Processing}
\label{subsec:datagen-graphprocessing}
%We develop an AIG2Grahpml framework where 
We wrote a gate-level netlist parser that takes a \texttt{BENCH} file (containing the AIG representation of an IP) as input and generates a corresponding \texttt{GRAPHML} file. % with the design as a DAG of the AIG representation of the design in \texttt{GRAPHML} format. 
%The \texttt{GRAPHML} file is an AIG representation of the design. 
While generating the \texttt{GRAPHML} format of the design, we preserve the fundamental characteristics of the AIGs. 
Nodes in the AIG are 2-input AND gates and the edges in the AIG are either inverters or buffers. We define Node type and number of incoming inverted edges as two node-based features and the edge type as the one edge feature. %The depth of the AIG is the longest path on the graph. 
For each  IP, we save $K \times L = 30,000$ graph structures.

\begin{table}[!tb]
\centering
\footnotesize
\setlength\tabcolsep{1.5pt}
%\resizebox{\textwidth}{!}{%
\begin{tabular}{@{}lllllllc@{}}
\toprule
 & \multicolumn{7}{c}{Characteristics of Benchmarks} \\ 
 \cmidrule(l){2-8} 
 \multirow{-2}{*}{IP} & PI & PO & N & E & I & D & Function \\ \midrule
\rt{spi~\cite{opencores}}  & \rt{254} & \rt{238} & \rt{4219} & \rt{8676} & \rt{5524} & \rt{35} &  \rt{Serial peripheral interface}\\
\rt{i2c\cite{opencores}}  & \rt{177} & \rt{128} & \rt{1169} & \rt{2466} & \rt{1188} & \rt{15} & \rt{Bidirectional serial bus protocol}\\
\rt{ss\_pcm\cite{opencores}} & \rt{104} & \rt{90} & \rt{462} & \rt{896} & \rt{434} & \rt{10} & \rt{Single slot PCM}\\
\rt{usb\_phy\cite{opencores}} & \rt{132} & \rt{90} & \rt{487} & \rt{1064} & \rt{513} & \rt{10} & \rt{USB PHY 1.1}\\
\rt{sasc\cite{opencores}} & \rt{135} & \rt{125} & \rt{613} & \rt{1351} & \rt{788} & \rt{9} &  \rt{Simple asynch serial controller}\\
\rt{wb\_dma\cite{opencores}} & \rt{828} & \rt{702} & \rt{4587} & \rt{9876} & \rt{4768} & \rt{29} & \rt{Wishbone DMA/Bridge} \\
\rt{simple\_spi\cite{opencores}} & \rt{164} & \rt{132} & \rt{930} & \rt{1992} & \rt{1084} & \rt{12} & \rt{MC68HC11E based SPI interface}\\
\rt{pci\cite{opencores}} &	\rt{3429} &	\rt{3157} &	\rt{19547} & \rt{42251}	& \rt{25719} &\rt{29} & \rt{PCI controller}\\
 \rt{wb\_conmax\cite{opencores}} &	\rt{2122} &	\rt{2075} &	\rt{47840} & \rt{97755}	& \rt{42138} &	\rt{24} & \rt{WISHBONE Conmax}  \\
\rt{ethernet\cite{opencores}} &	\rt{10731} & \rt{10422} & \rt{67164} & \rt{144750}	& \rt{86799} &	\rt{34} & \rt{Ethernet IP core}\\
 \midrule
 \gt{ac97\_ctrl\cite{opencores}} &	\gt{2339} &	\gt{2137} &	\gt{11464} & \gt{25065} & \gt{14326} &	\gt{11} &  \gt{Wishbone ac97} \\
 \gt{mem\_ctrl\cite{opencores}} &	\gt{1187} &	\gt{962}& \gt{16307} & \gt{37146} & \gt{18092} &	\gt{36} & \gt{Wishbone mem controller}\\
 \gt{bp\_be\cite{bpsoc}} &	\gt{11592} & \gt{8413} & \gt{82514} & \gt{173441}	& \gt{109608} &	\gt{86} & \gt{Black parrot RISCV processor engine}\\
 \gt{vga\_lcd\cite{opencores}} & \gt{17322} & \gt{17063} &	\gt{105334} & \gt{227731} & \gt{141037} & \gt{23} & \gt{Wishbone enhanced VGA/LCD controller} \\
\midrule
\bt{des3\_area\cite{opencores}} & \bt{303} & \bt{64} & \bt{4971} & \bt{10006} & \bt{4686} & \bt{30} & \bt{DES3 encrypt/decrypt}\\
\bt{aes\cite{opencores}} & \bt{683} & \bt{529} & \bt{28925} & \bt{58379} & \bt{20494} & \bt{27} &  \bt{AES (LUT-based)}\\
\bt{sha256\cite{mitll-cep}} &	\bt{1943} &	\bt{1042} &	\bt{15816} & \bt{32674}	& \bt{18459} &	\bt{76} & \bt{SHA256 hash} 	\\
\bt{aes\_xcrypt\cite{aesxcrypt}} &	\bt{1975} &	\bt{1805} &	\bt{45840} & \bt{93485}	& \bt{36180} &	\bt{43} & \bt{AES-128/192/256}\\	%\url{https://github.com/crypt-xie/XCryptCore} \\
\bt{aes\_secworks\cite{aessecworks}}  &	\bt{3087} &	\bt{2604} &	\bt{40778} & \bt{84160}	&\bt{45391} &	\bt{42} & \bt{AES-128  (simple)} \\	%\url{https://github.com/secworks/aes} \\
\midrule
\mat{fir\cite{mitll-cep}} & \mat{410} & \mat{351} & \mat{4558} & \mat{9467} & \mat{5696} & \mat{47} & \mat{FIR filter}\\
\mat{iir\cite{mitll-cep}} & \mat{494} & \mat{441} & \mat{6978} & \mat{14397} & \mat{8596} & \mat{73} & \mat{IIR filter} \\
\mat{jpeg\cite{opencores}} & \mat{4962} & \mat{4789} &	\mat{114771} &	\mat{234331} & \mat{146080} &	\mat{40} & \mat{JPEG encoder}\\
\mat{idft\cite{mitll-cep}} & \mat{37603} &	\mat{37419} & \mat{241552} & \mat{520523} & \mat{317210} &	\mat{43} & \mat{Inverse DFT}  \\
\mat{dft\cite{mitll-cep}} &	\mat{37597} & \mat{37417} &	\mat{245046} &	\mat{527509} & \mat{322206} &	\mat{43} & \mat{DFT design}\\

\midrule
\vt{tv80\cite{opencores}} &	\vt{636}	& \vt{361} & \vt{11328} & \vt{23017}	& \vt{11653} &	\vt{54} & \vt{TV80 8-Bit Microprocessor}  \\
\vt{tiny\_rocket\cite{ajayi2019openroad}} &	\vt{4561} &	\vt{4181} &	\vt{52315} & \vt{108811}	& \vt{67410} & \vt{80} & \vt{32-bit tiny riscv core}\\
\vt{fpu\cite{balkind2016openpiton}} &	\vt{632} &	\vt{409} &	\vt{29623} & \vt{59655}	& \vt{37142} &	\vt{819} & \vt{OpenSparc T1 floating point unit} \\
% \midrule 
\bst{picosoc\cite{balkind2016openpiton}} &	\bst{11302} &	\bst{10797} &	\bst{82945} & \bst{176687}	& \bst{107637} &	\bst{43} & \bst{SoC with PicoRV32 riscv} \\
\fgt{dynamic\_node\cite{ajayi2019openroad}}  & \fgt{2708} & \fgt{2575} & \fgt{18094} & \fgt{38763} & \fgt{23377} & \fgt{33}  & \fgt{OpenPiton NoC architecture}\\

	%\url{https://github.com/black-parrot/black-parrot} \\

\bottomrule
\end{tabular}
\caption{Open source IP characteristics (unoptimized).  Primary Inputs (PI), Primary outputs (PO), `Nodes (N), Edges (E), `Inverted edges (I), Netlist Depth (D). Color code: \rt{Communication/Bus protocol}, \gt{Controller}, \bt{Crypto}, \mat{DSP}, \vt{Processor}, \bst{Processor$+$control}, \fgt{Control$+$Communication}\label{tab:structuralcharacteristics}}

%}
\end{table}

% \subsection{OpenABC-D Dataset Generation for EDA \ac{ML} Pipeline}
\subsection{Preprocessing for \ac{ML}}
\label{subsec:datagen-mlpipeline}
This stage involves preparing the circuit data for use with any \ac{ML} framework. % for performing a variety of \ac{ML} tasks. 
We use 
%a graph-based \ac{ML} framework: 
pytorch-geometric APIs to create data samples using AIG graphs, synthesis recipes, and area/delay data from synthesis runs. The dataset is available using a customized dataloader for easy handling for preprocessing, labeling, and transformation. 
We also provide a script that helps partition the dataset (e.g., into train/test) based on user specified learning tasks. 
% The dataset supports an easy splitting infrastructure based on criteria and tasks, helping 
% The framework has a rich \ac{ML}-ready processed dataset that can help \ac{ML} for 
Our aim is to help the EDA community to focus on their domain problems without investing time on data generation and labeling.

\section{OpenABC-D Characteristics}
\label{sec:data_character}

% \begin{figure}[t]
% \centering
% %\includegraphics[width=0.8\textwidth]{figures/datapointDescription.png}
% \includegraphics[width=\textwidth]{figures/datapointDesc.png}
% \caption{Data sample description of OpenABC-D generated dataset}
% \label{fig:datapointdesc}
% \end{figure}

% \usepackage{multirow}
% \usepackage{multirow}

To produce the OpenABC-dataset, we use 29 open source IPs with a wide range of functions. 
In the absence of a preexisting dataset like ImageNet that represents many classes of IPs, we hand-curated IPs of different functionalities from MIT LL labs CEP~\cite{mitll-cep}, OpenCores~\cite{opencores}, and IWLS~\cite{albrecht2005iwls}. %The primary reason for considering such benchmarks: 
%1)
These benchmarks are more complex and functionally diverse compared to ISCAS~\cite{iscas85,iscas89} and EPFL~\cite{amaru2015epfl} benchmarks that are used in prior work and represent large, industrial-sized IP. 
In the context of EDA, functionally diverse IPs should have diverse AIG structures (e.g., tree-like, balanced, and skewed) that mimic the distribution of real hardware designs. 
\autoref{tab:structuralcharacteristics} summarizes the structural and functional characteristics of the data after synthesis (without optimizations).
These designs are functionally diverse -- bus communication protocols, computing processors, digital signal processing cores, cryptographic accelerators and system controllers. 

\begin{figure}[h]
    \centering
        \hspace*{-0.2in}
    \subfloat[\label{fig:h1}Top $1\%$]{\includegraphics[width=0.42\textwidth, valign=c]{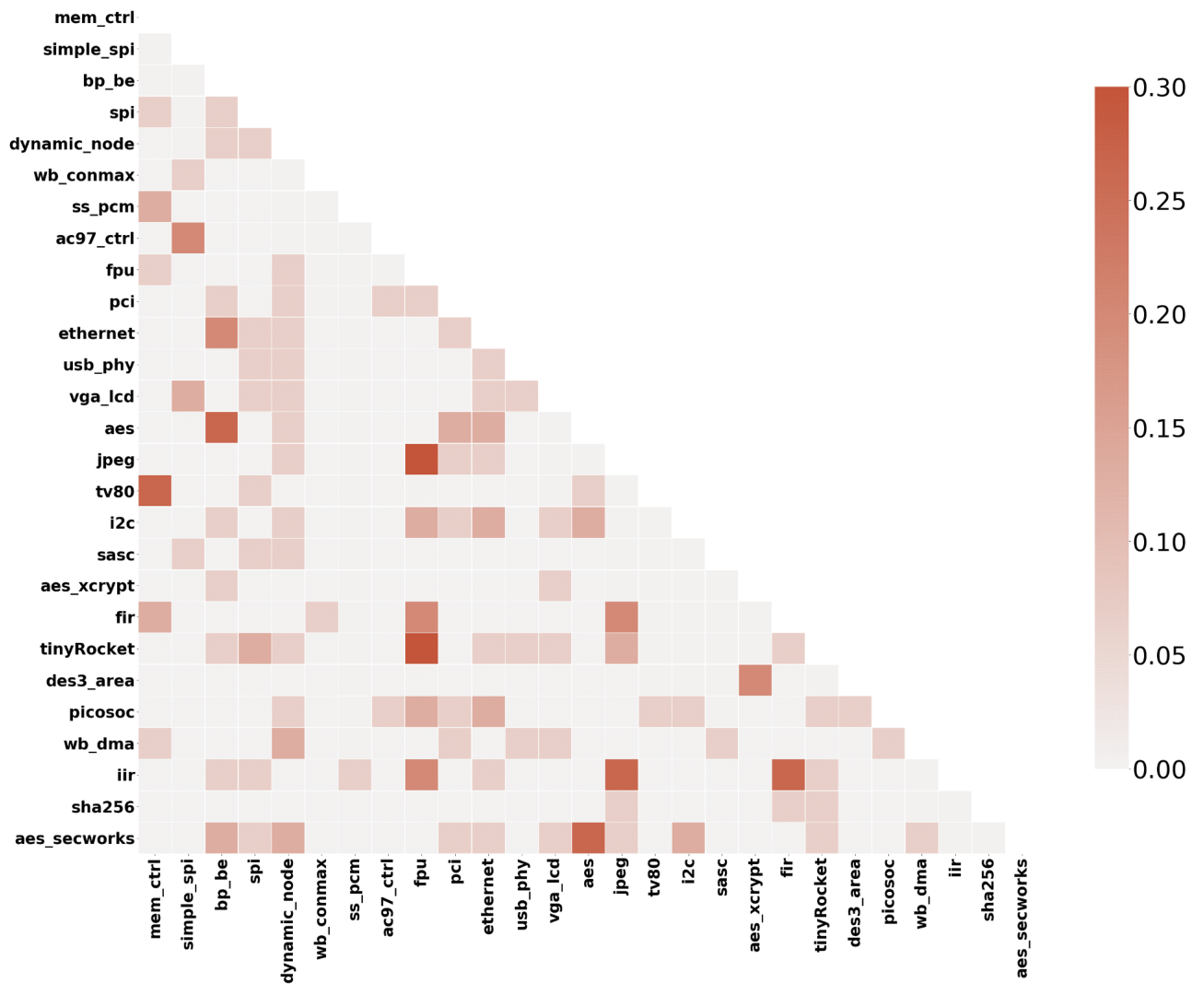}}
    \hspace*{-0.27in}
    \subfloat[\label{fig:h2}Top $5\%$]{\includegraphics[width=0.42\textwidth, valign=c]{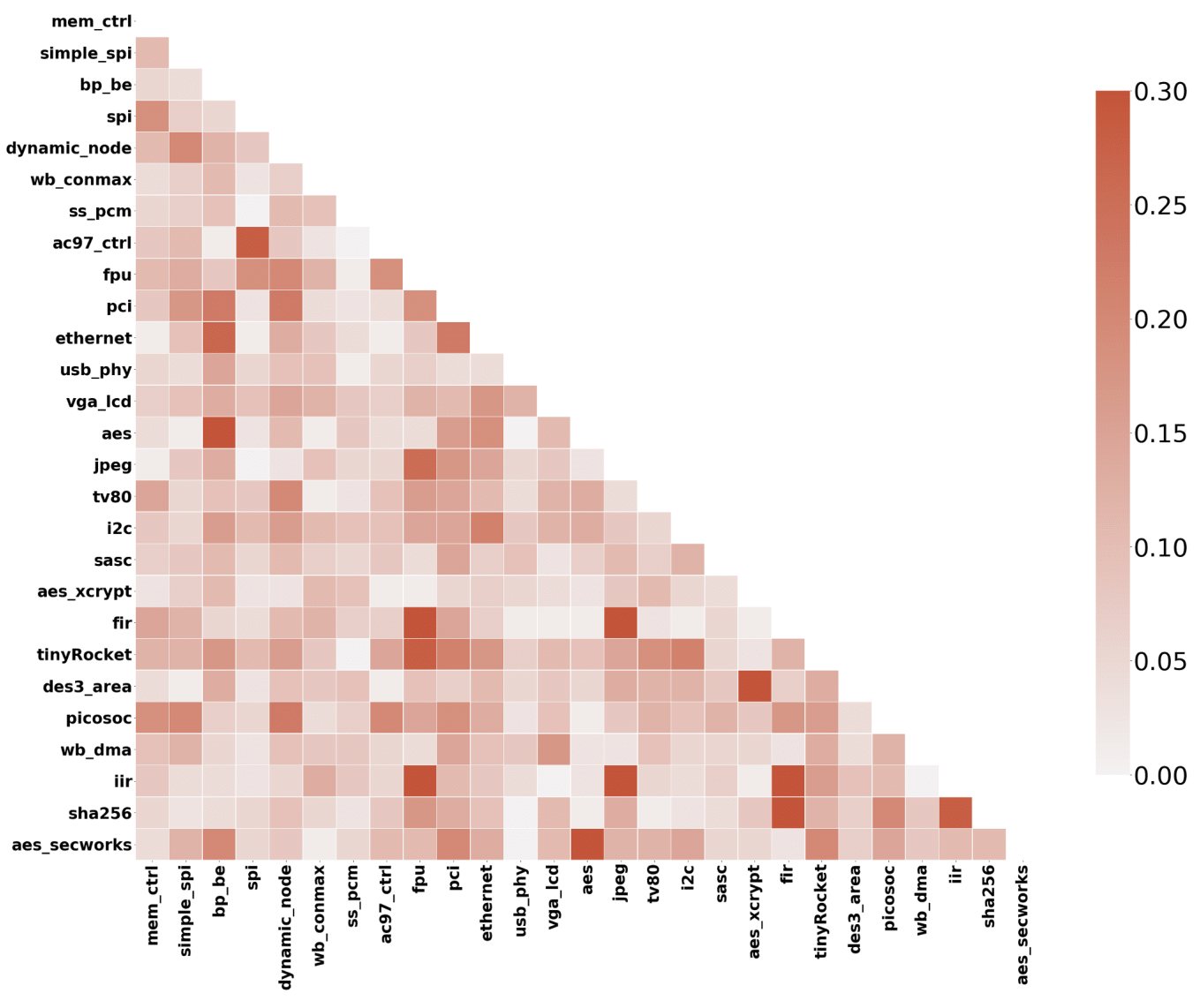}} 
       \hspace*{-0.27in}
    \subfloat[\label{fig:h3}Top $10\%$]{\includegraphics[width=0.42\textwidth, valign=c]{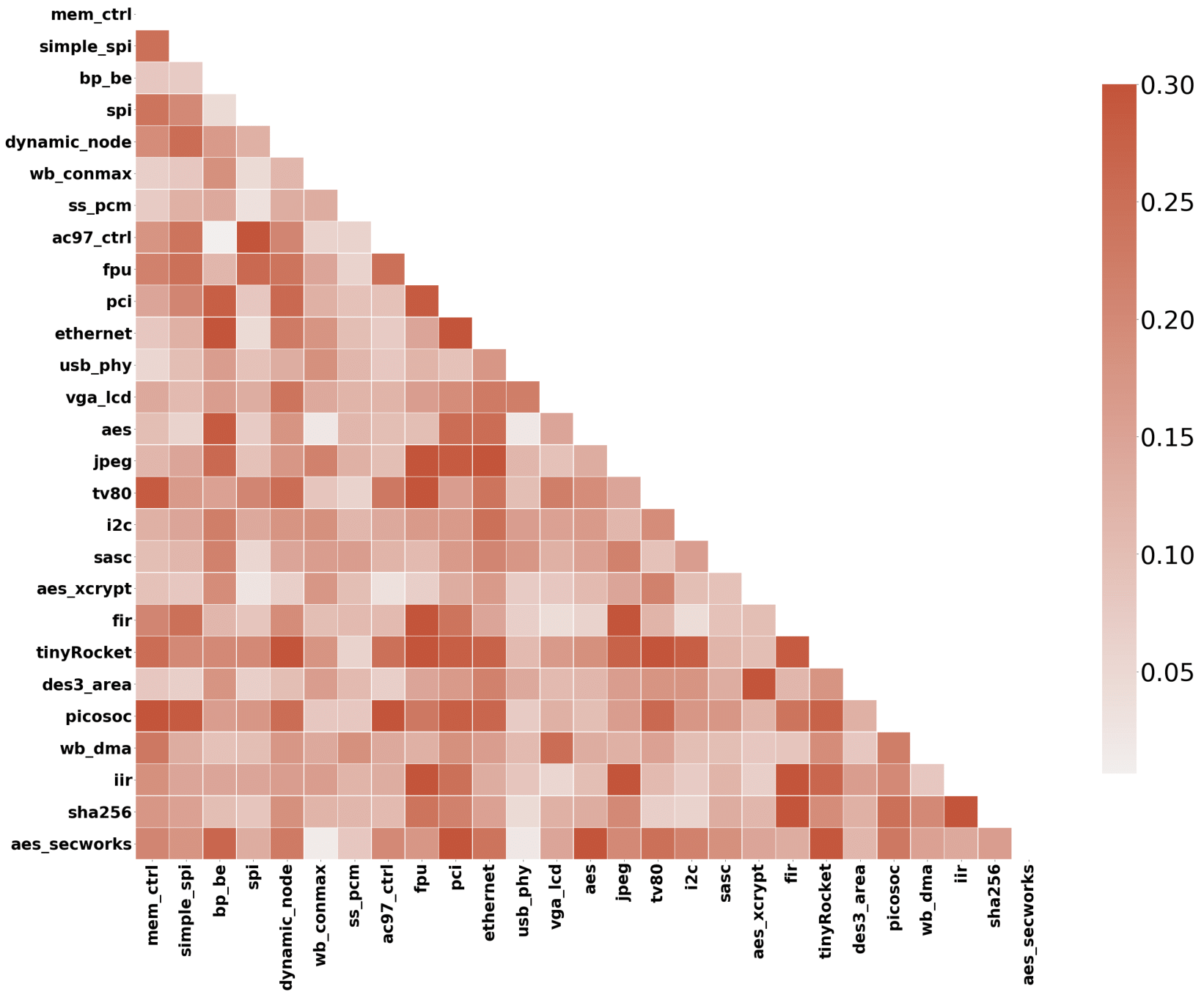}}
  \caption{Correlation plots amongst top $k\%$ synthesis recipes various IPs. Darker colors indicate higher similarity between the top synthesis recipes for the pair of IPs.}
    \label{fig:synthesis-tops}
\end{figure} 

\begin{figure}[h]
    \centering
    \subfloat[\label{fig:ac97_ctrl}]{\includegraphics[width=0.24\columnwidth, valign=c]{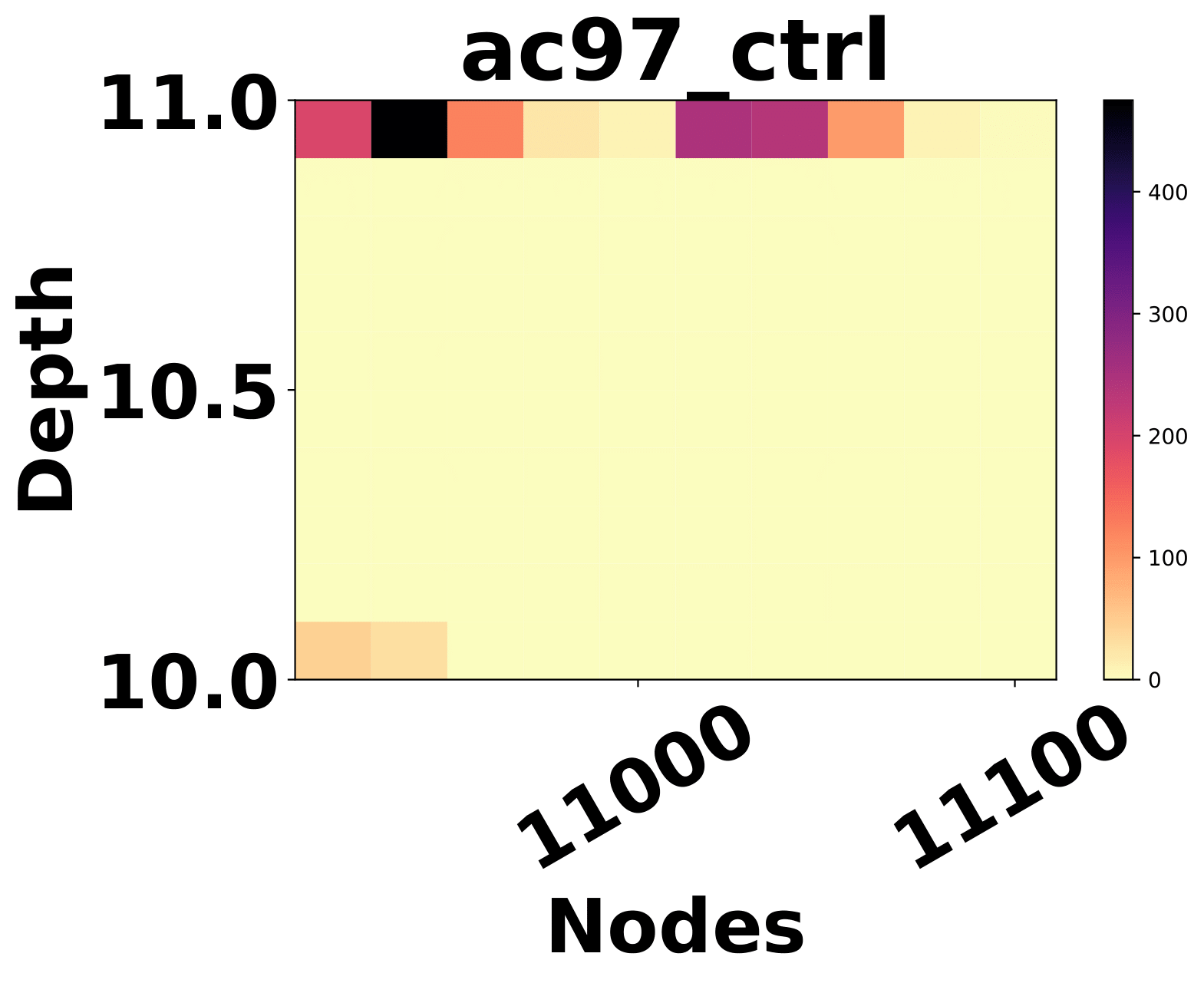}}
    \subfloat[\label{fig:aes_secworks}]{\includegraphics[width=0.2\columnwidth, valign=c]{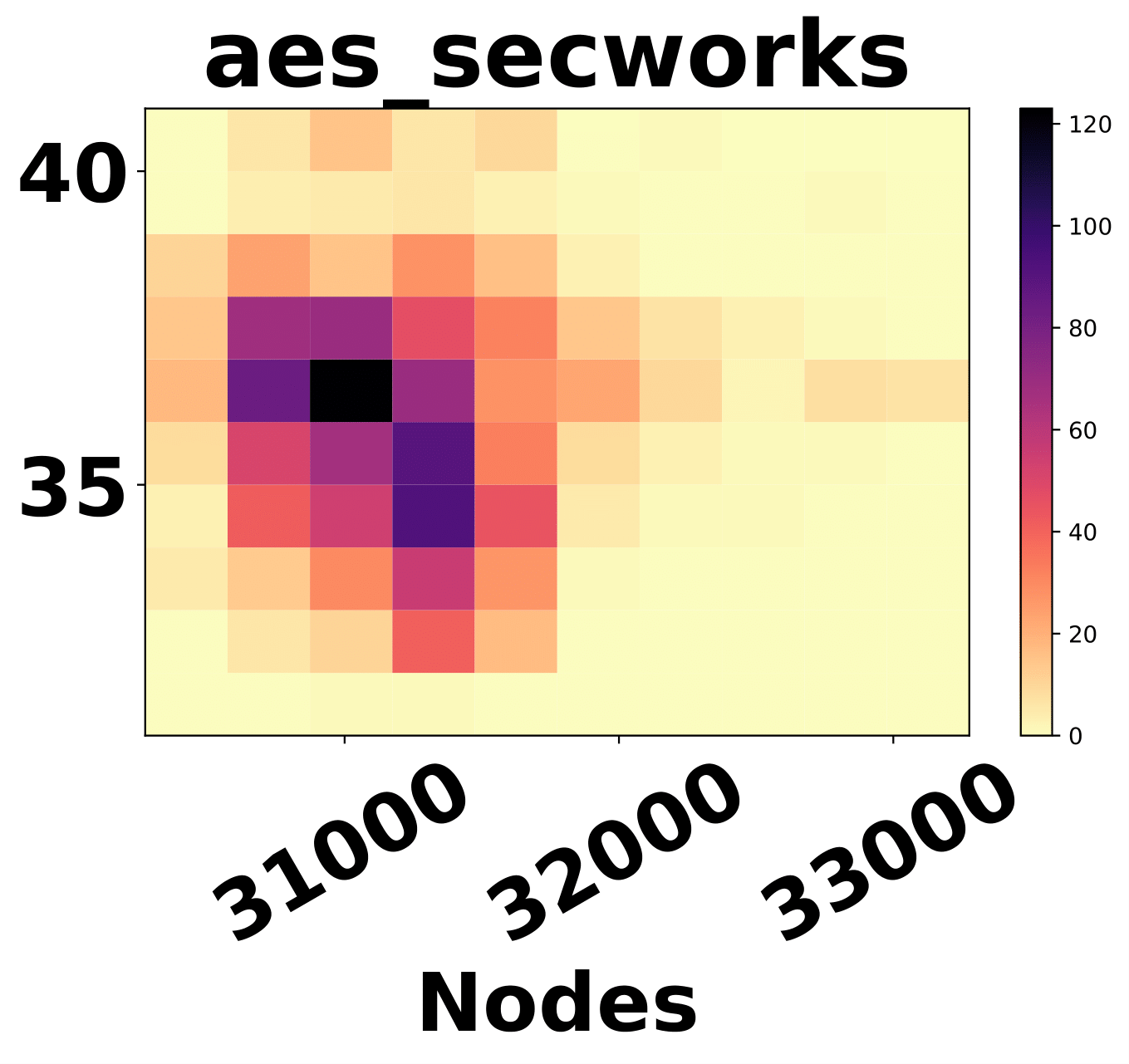}}
    \subfloat[\label{fig:aes_xcrypt}]{\includegraphics[width=0.2\columnwidth, valign=c]{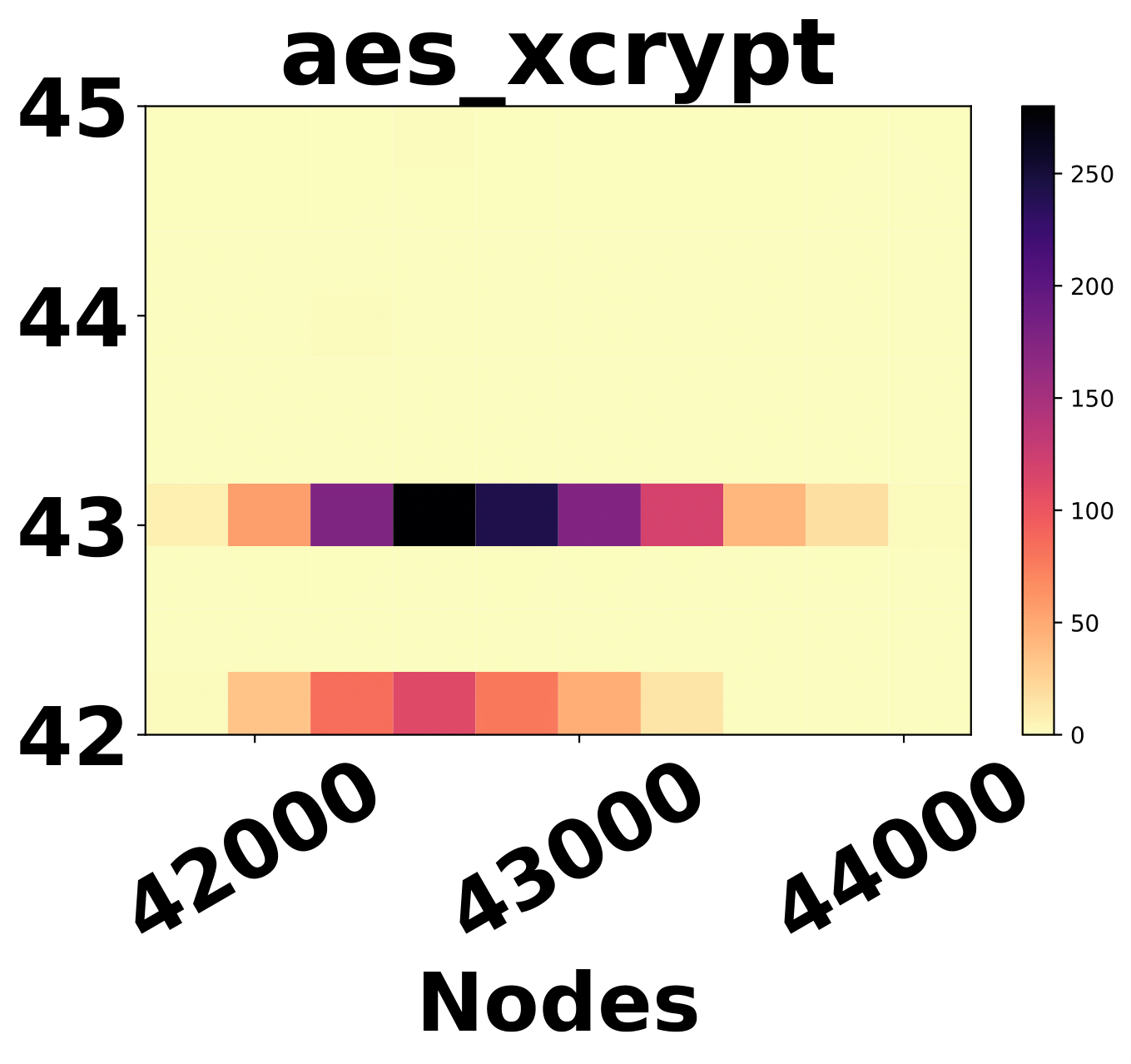}}
    \subfloat[\label{fig:aes}]{\includegraphics[width=0.2\columnwidth, valign=c]{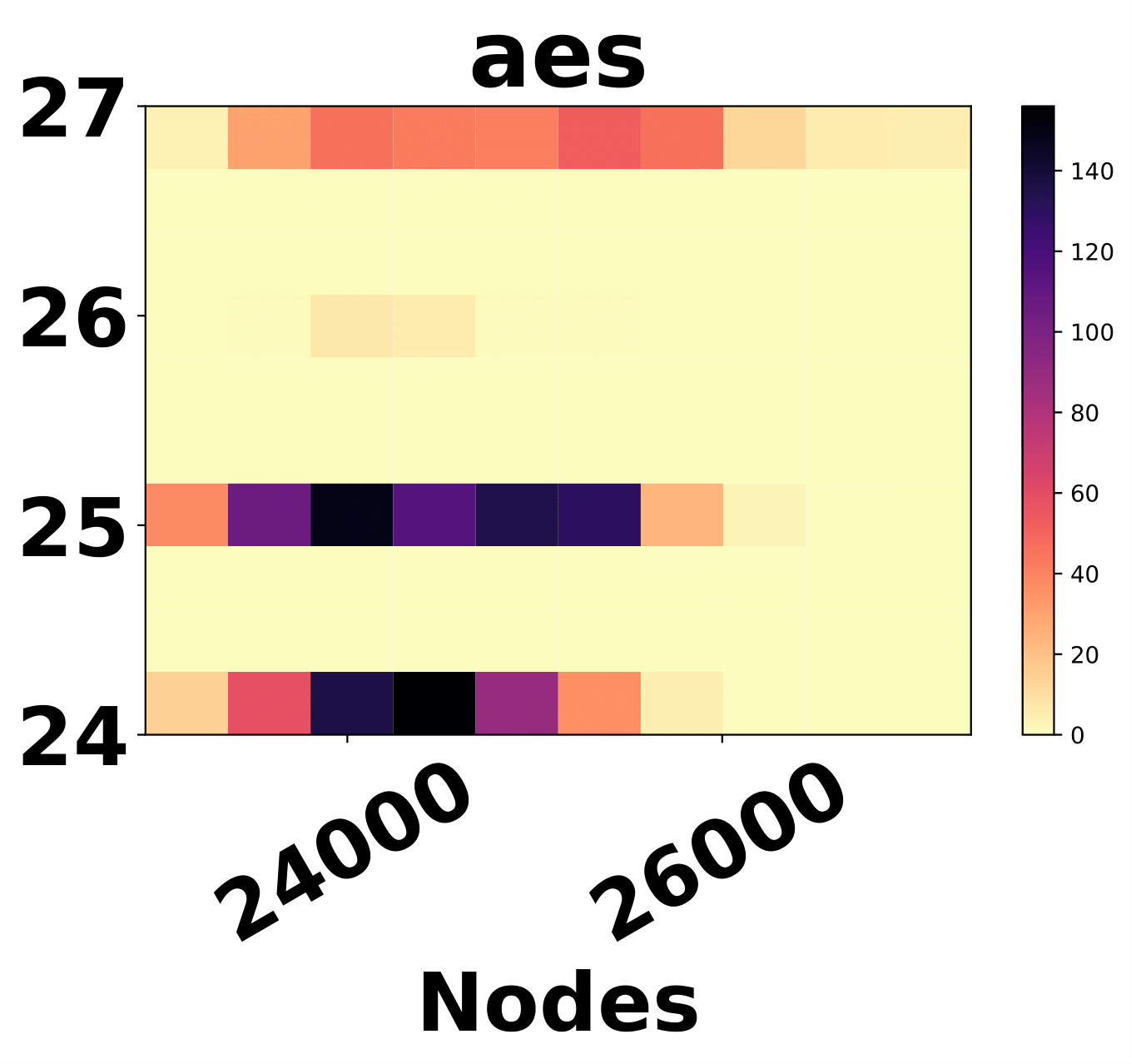}}
    \subfloat[\label{fig:bp_be}]{\includegraphics[width=0.23\columnwidth, valign=c]{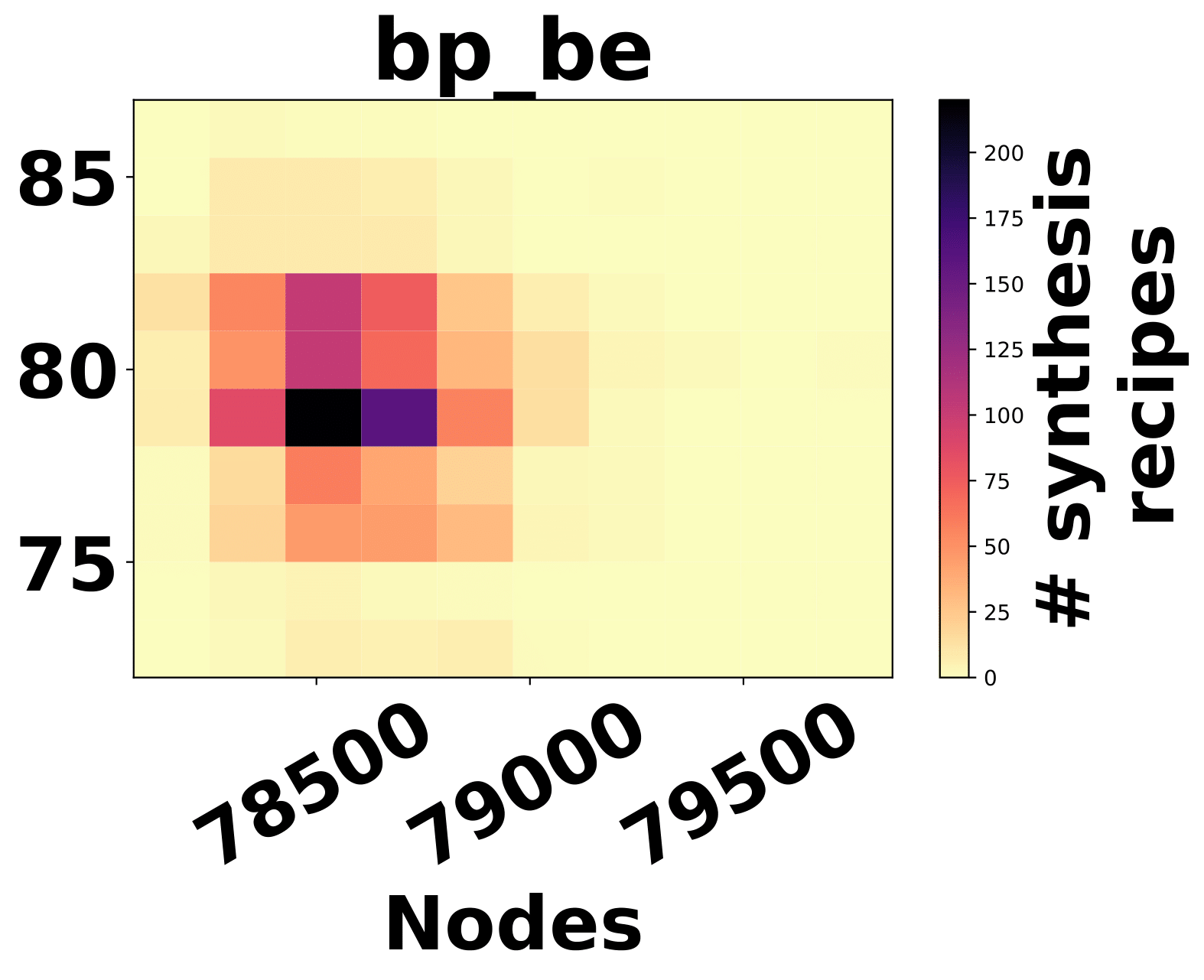}} \\ %\vspace*{-0.2in}
    \subfloat[\label{fig:des3_area}]{\includegraphics[width=0.23\columnwidth, valign=c]{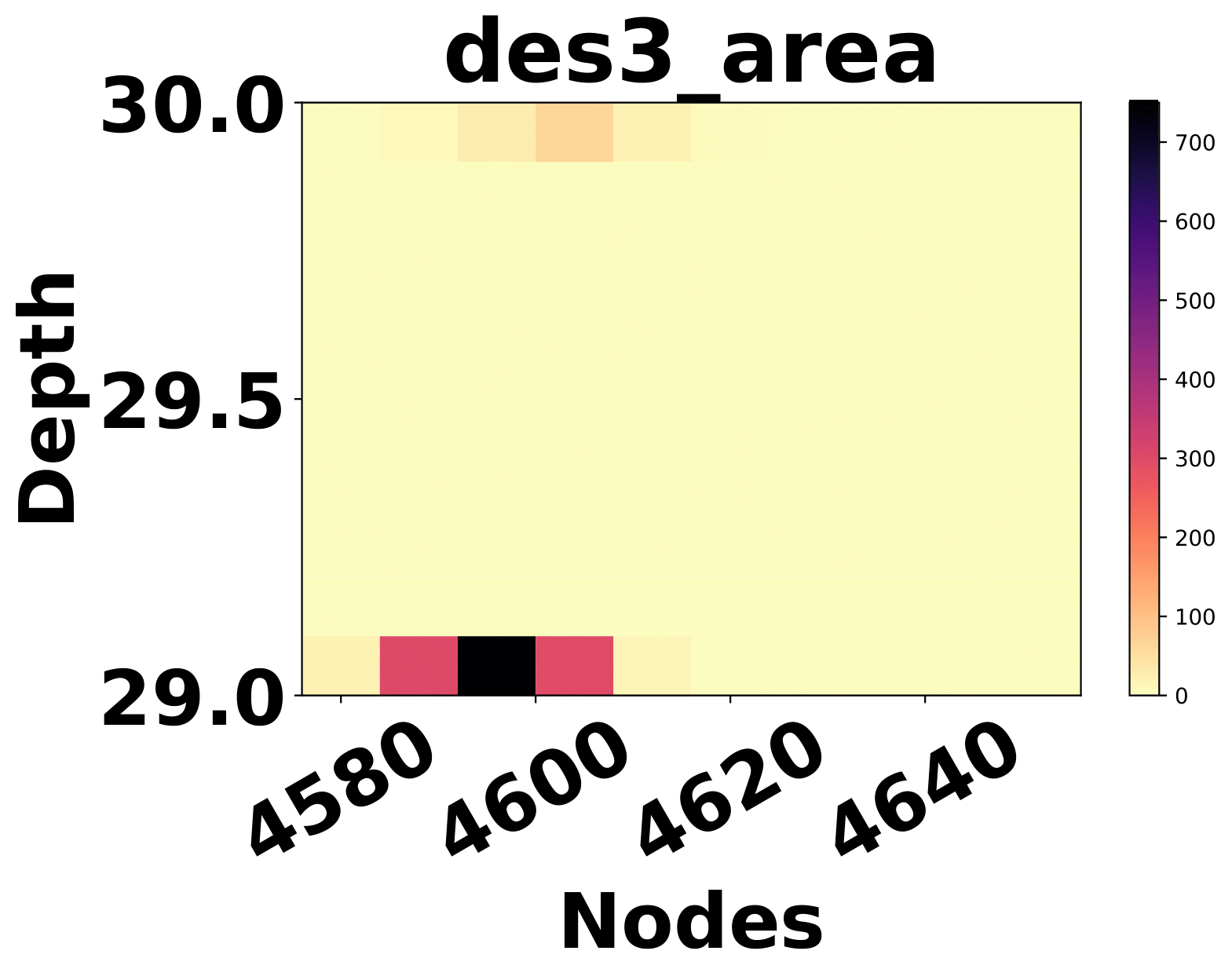}}
    \subfloat[\label{fig:dynamic_node}]{\includegraphics[width=0.19\columnwidth, valign=c]{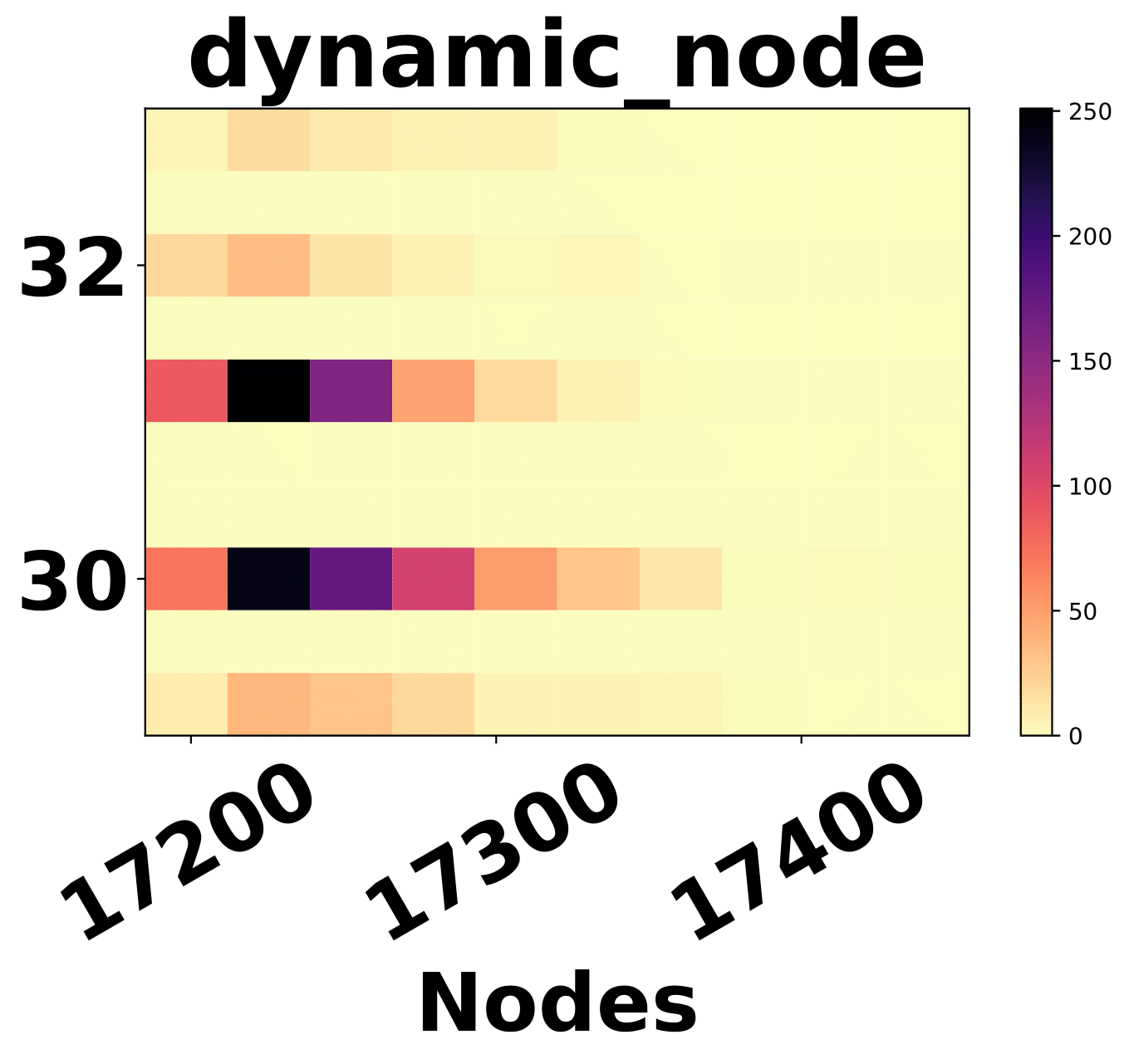}}
    \subfloat[\label{fig:ethernet}]{\includegraphics[width=0.19\columnwidth, valign=c]{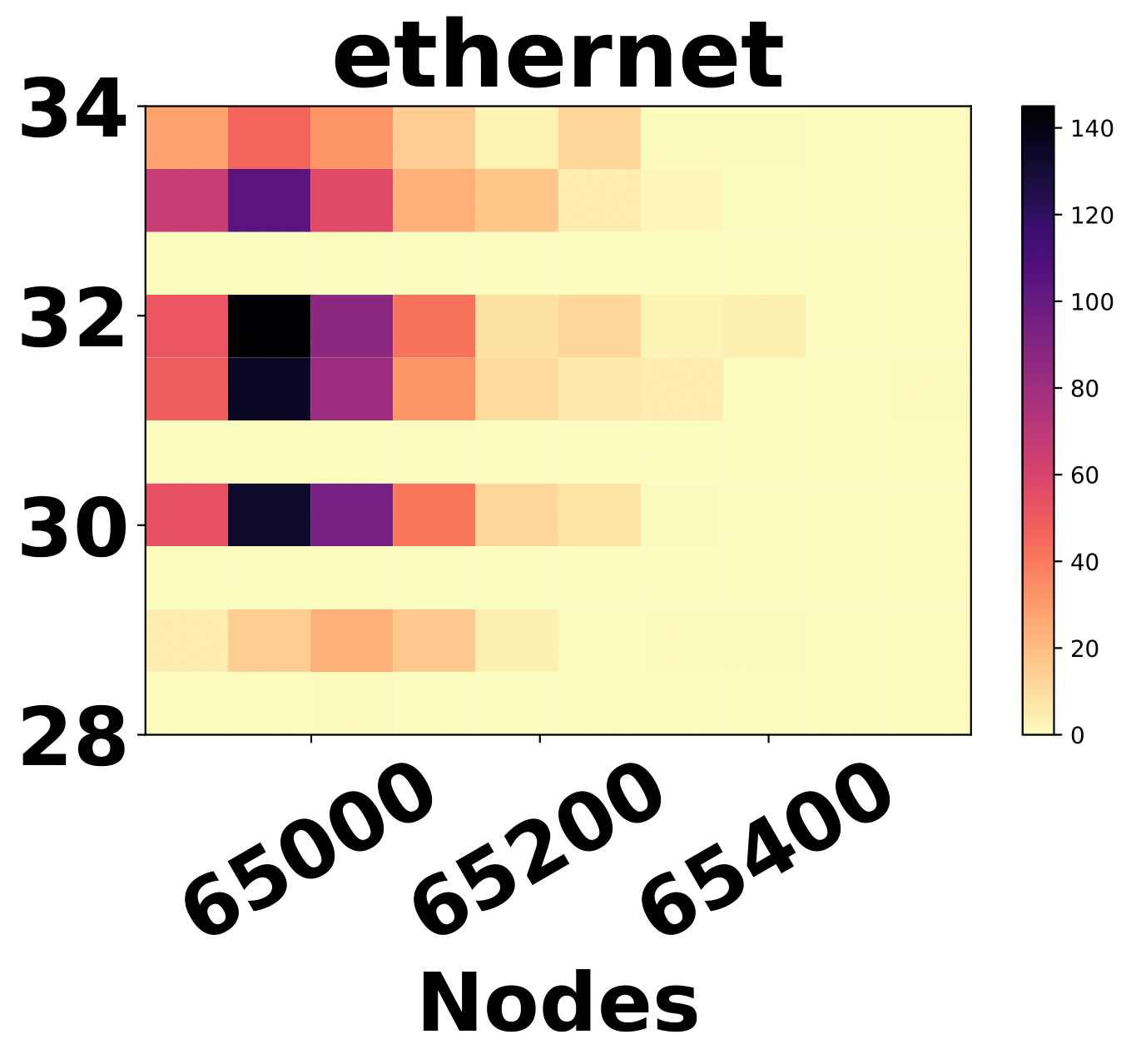}}
    \subfloat[\label{fig:fir}]{\includegraphics[width=0.21\columnwidth, valign=c]{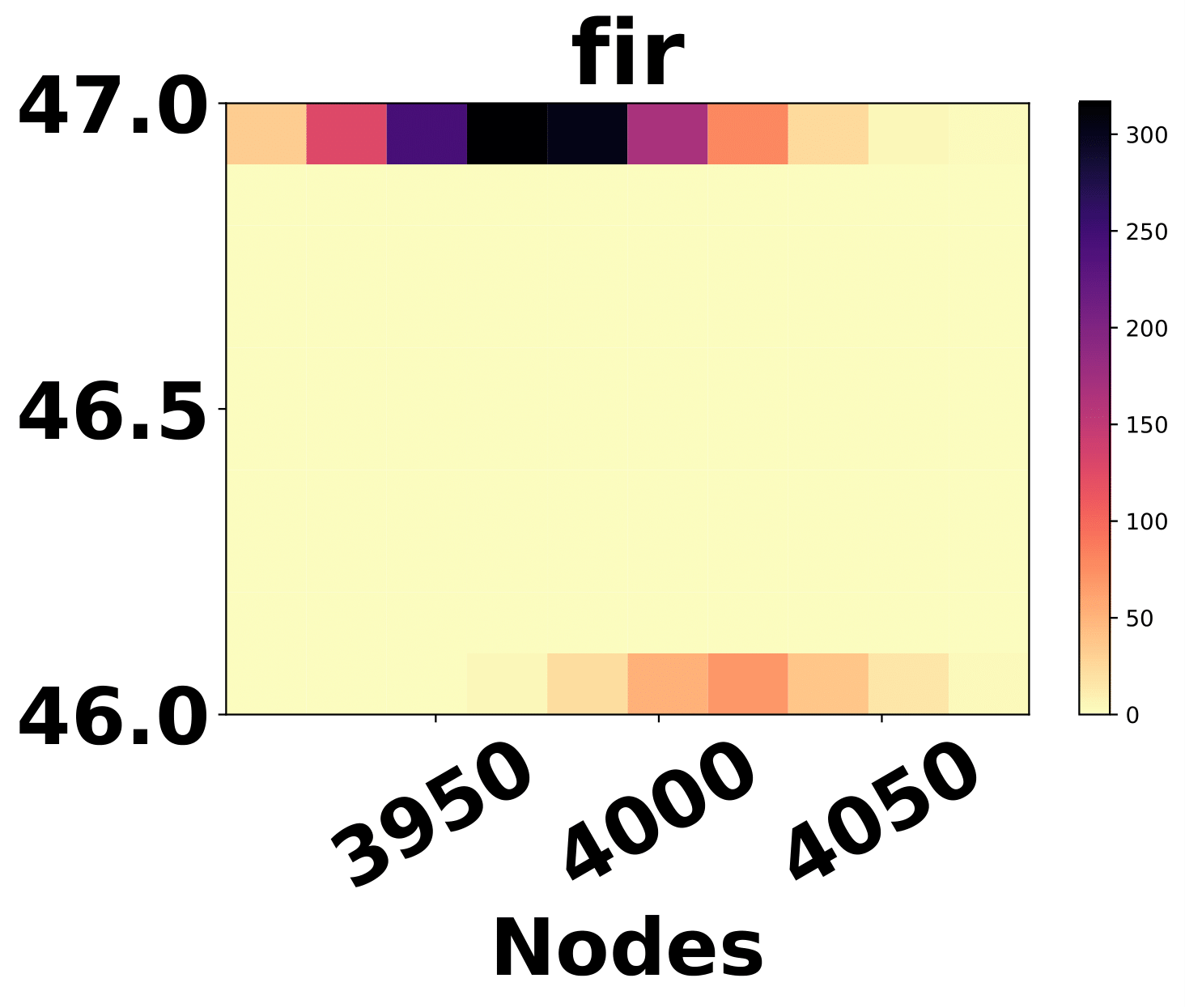}}
    \subfloat[\label{fig:i2c}]{\includegraphics[width=0.23\columnwidth, valign=c]{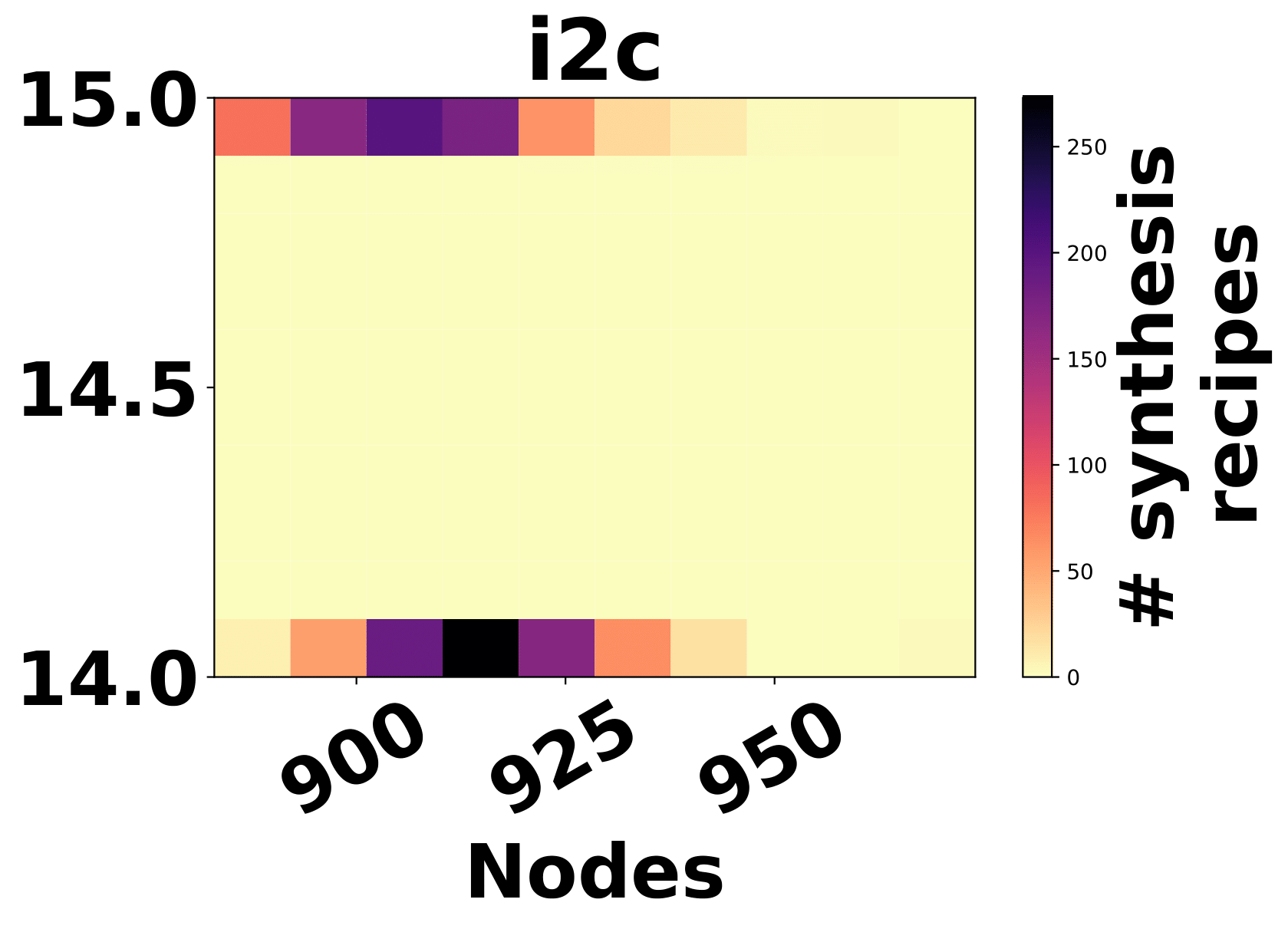}} \\
    %\vspace*{-0.2in}
    \subfloat[\label{fig:iir}]{\includegraphics[width=0.21\columnwidth, valign=c]{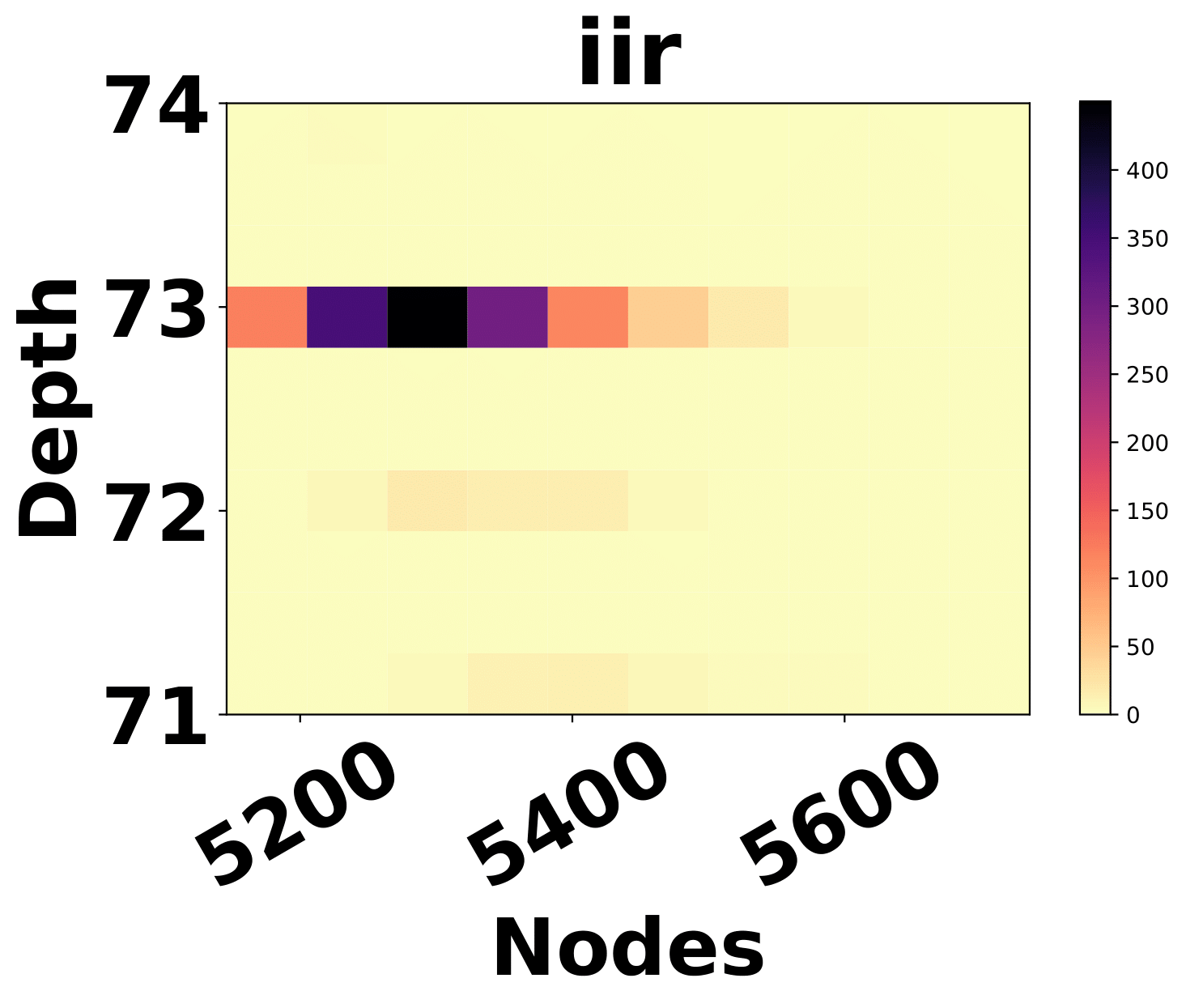}}
    \subfloat[\label{fig:jpeg}]{\includegraphics[width=0.2\columnwidth, valign=c]{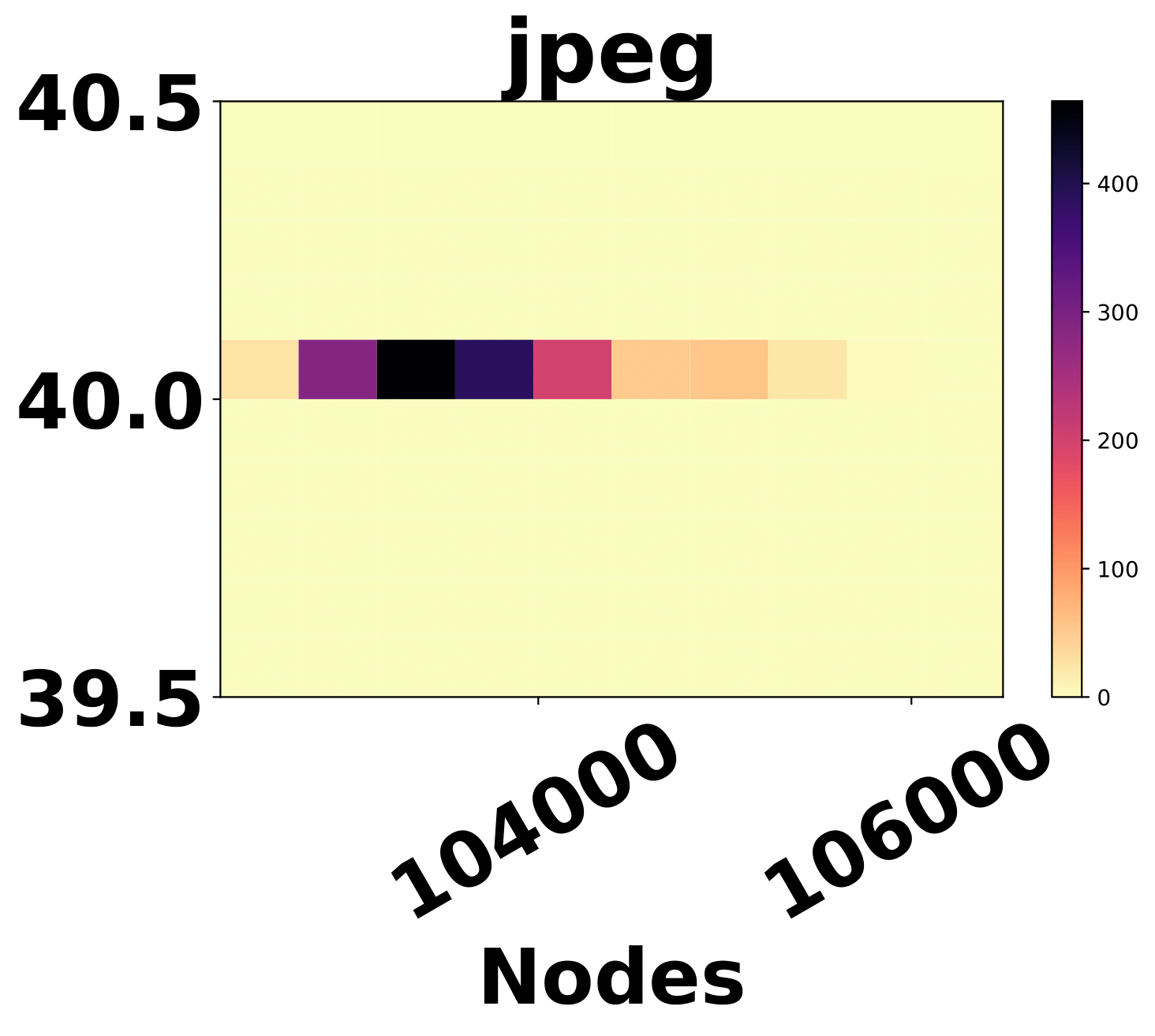}}
    \subfloat[\label{fig:mem-ctrl}]{\includegraphics[width=0.2\columnwidth, valign=c]{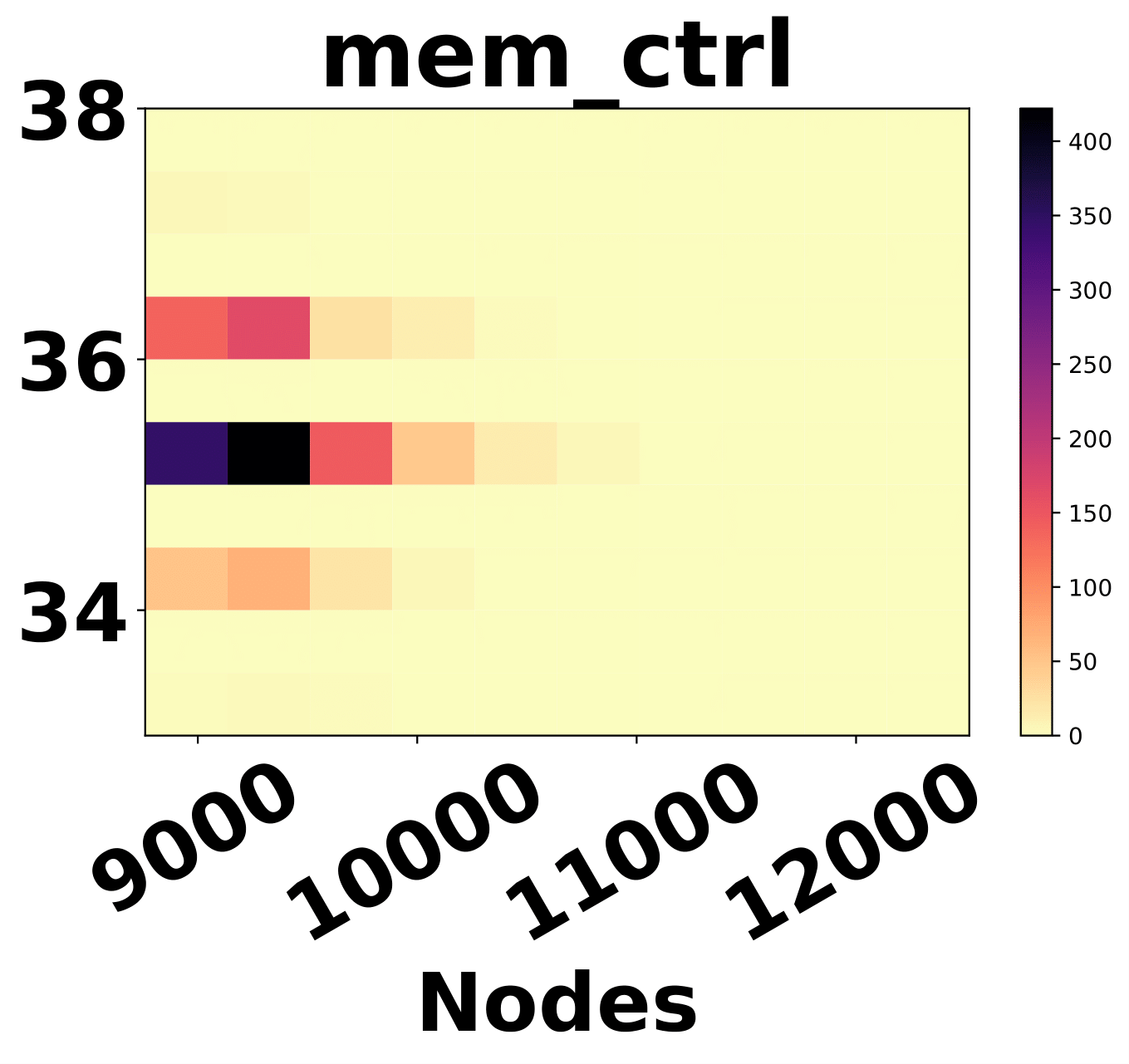}}
    \subfloat[\label{fig:pci}]{\includegraphics[width=0.2\columnwidth, valign=c]{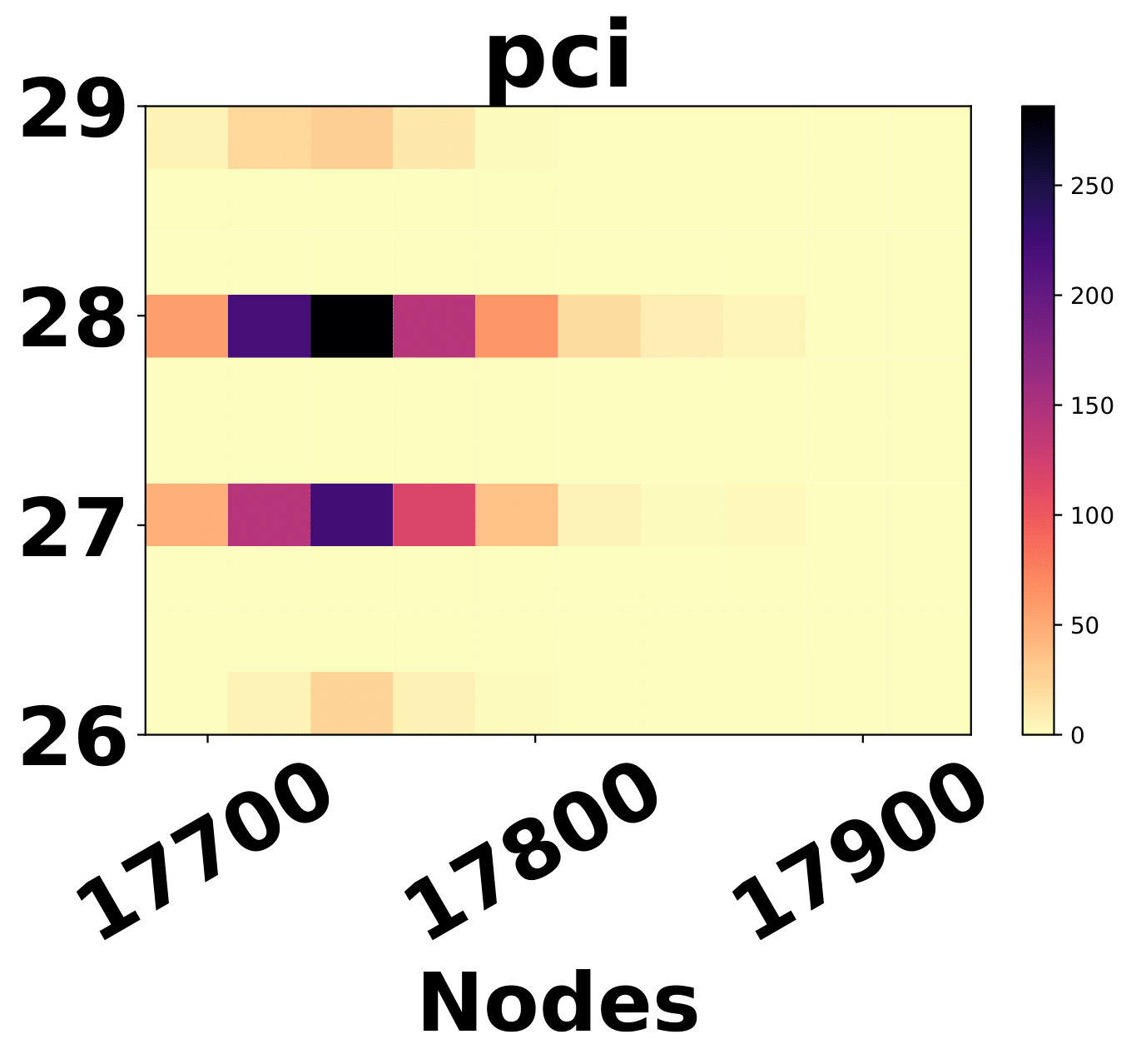}}
    \subfloat[\label{fig:picosoc}]{\includegraphics[width=0.23\columnwidth, valign=c]{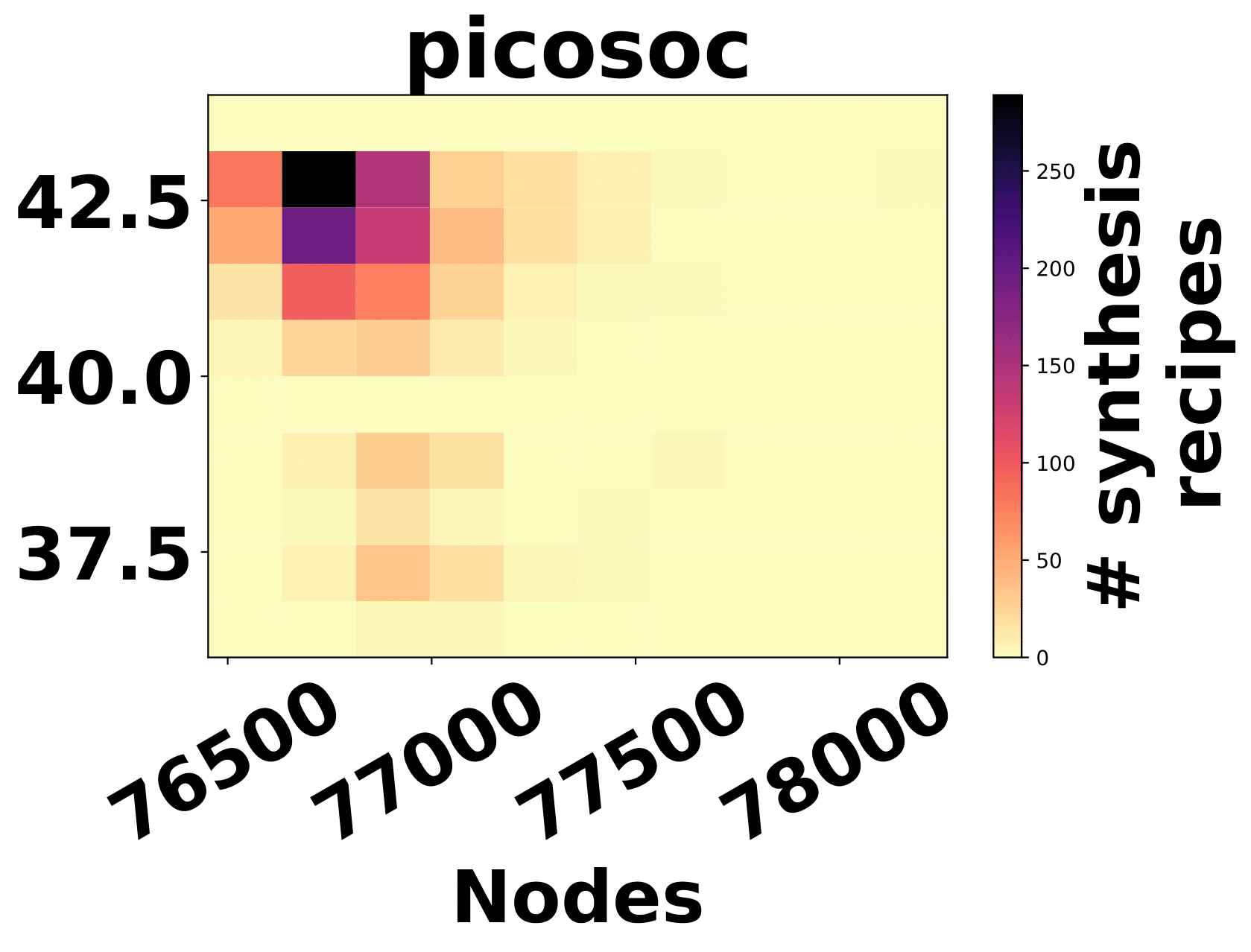}} \\   
    %\vspace*{-0.2in}
    \subfloat[\label{fig:sasc}]{\includegraphics[width=0.21\columnwidth, valign=c]{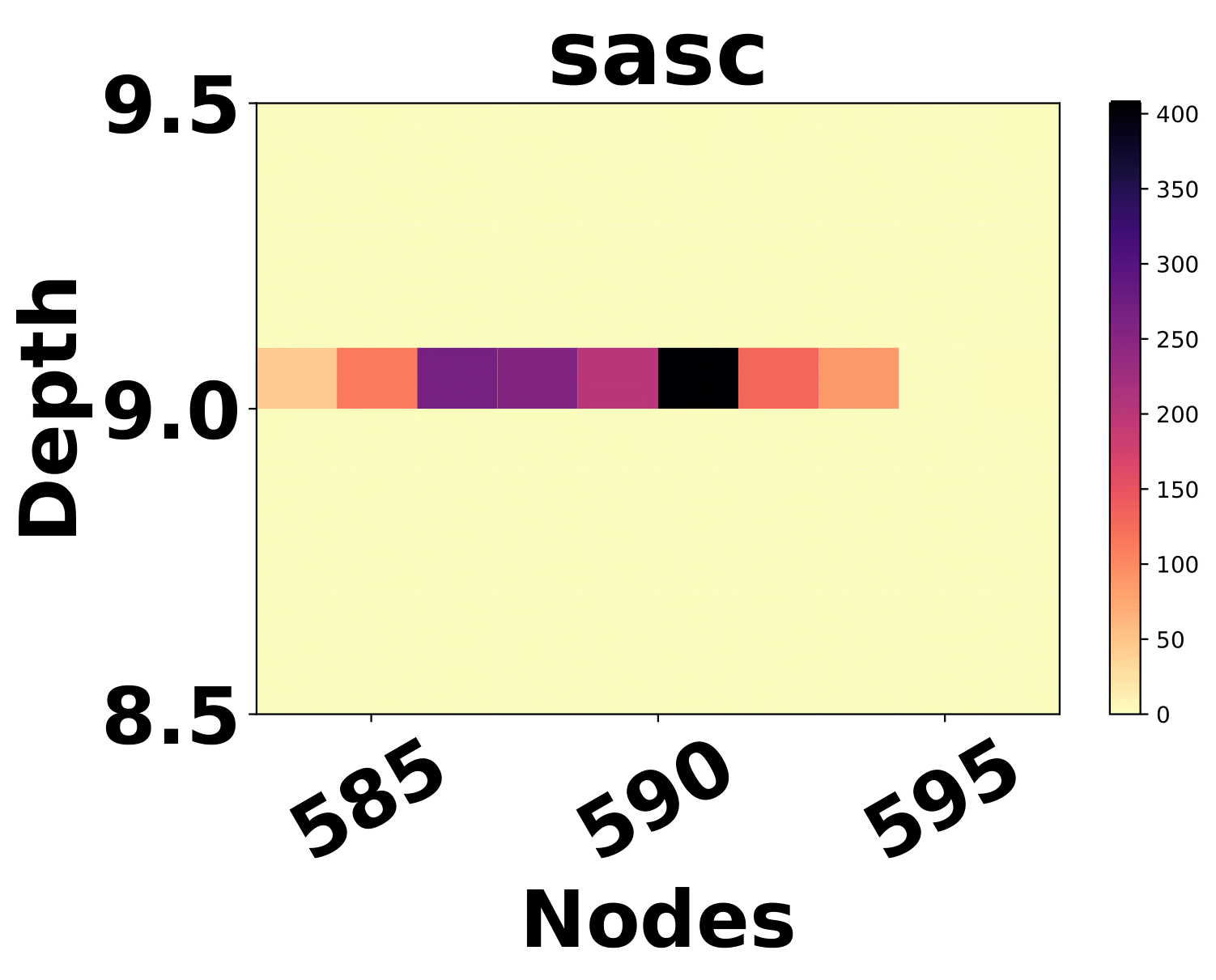}}
    \subfloat[\label{fig:sha256}]{\includegraphics[width=0.2\columnwidth, valign=c]{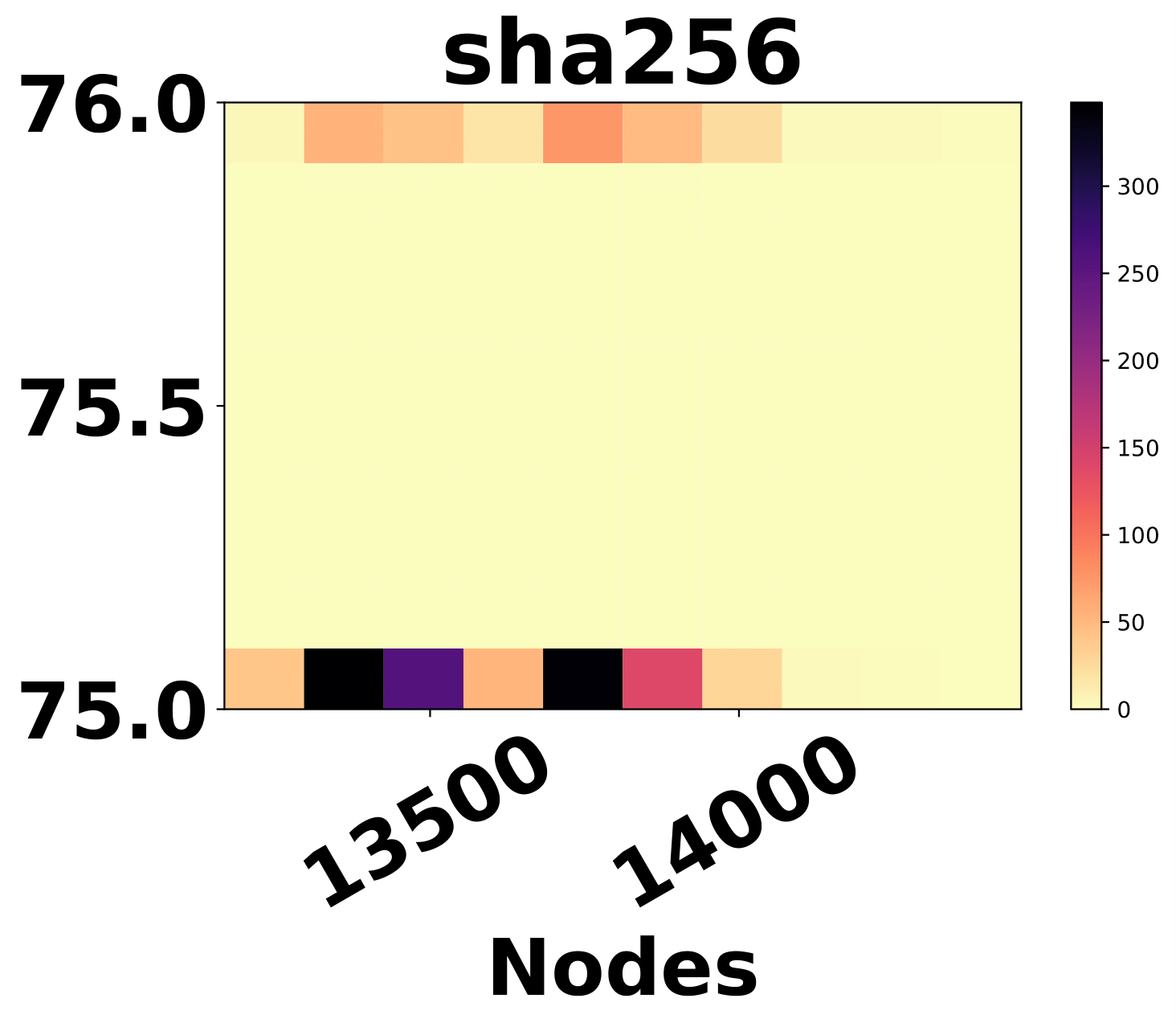}}
    \subfloat[\label{fig:simple-spi}]{\includegraphics[width=0.2\columnwidth, valign=c]{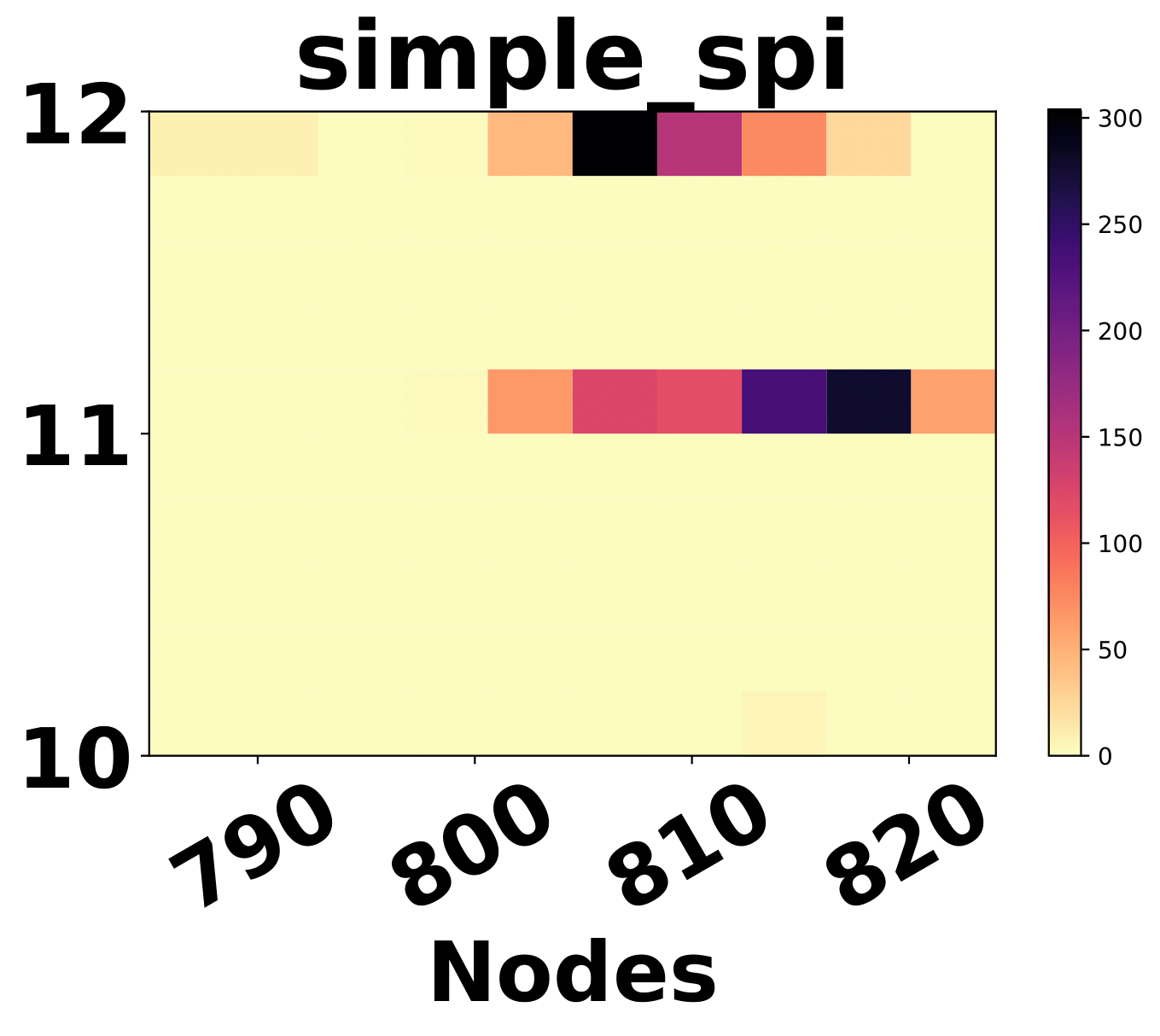}}
    \subfloat[\label{fig:spi}]{\includegraphics[width=0.2\columnwidth, valign=c]{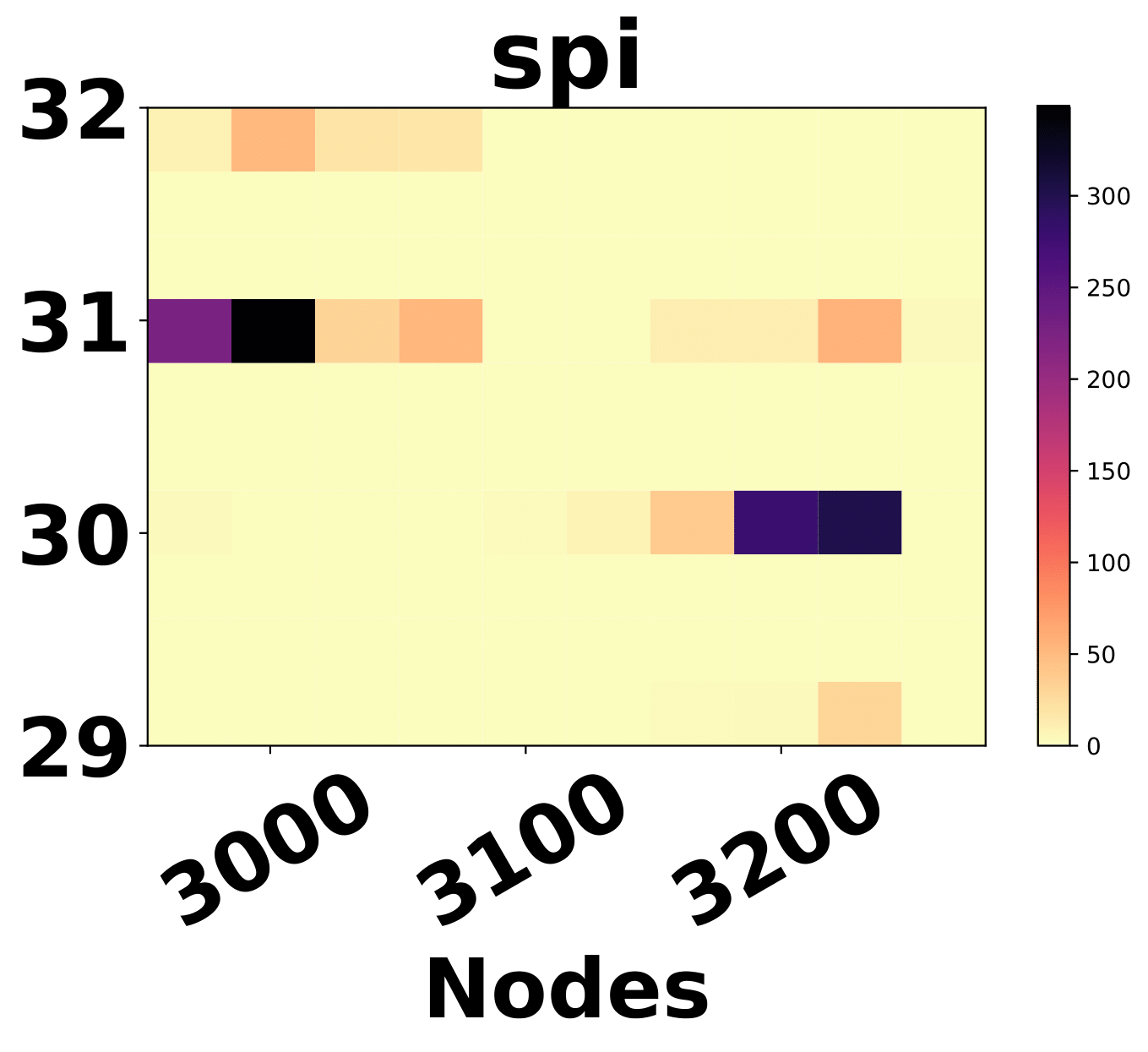}}
    \subfloat[\label{fig:ss-pcm}]{\includegraphics[width=0.23\columnwidth, valign=c]{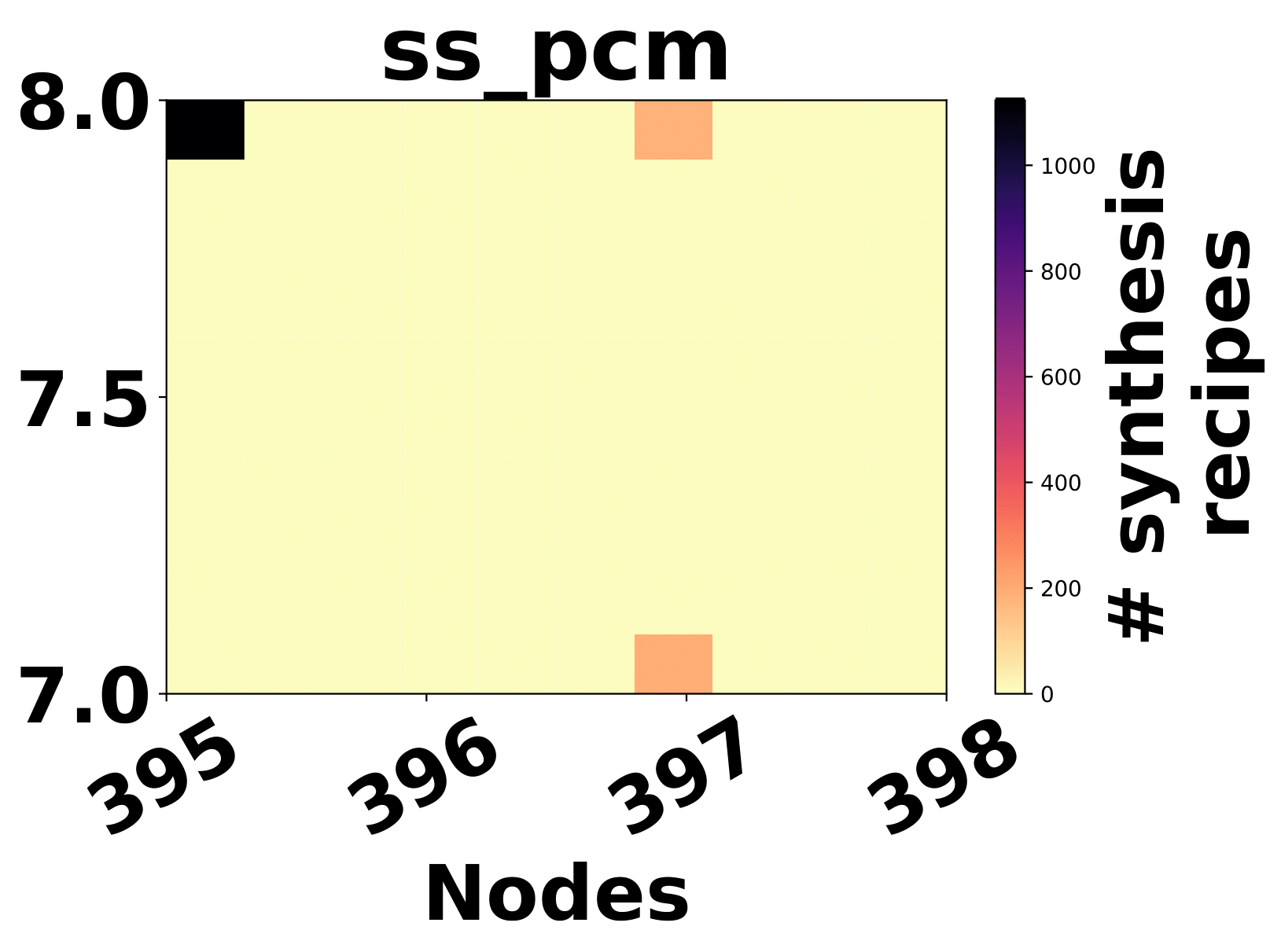}} \\
    %\vspace*{-0.2in}
    \subfloat[\label{fig:tinyrocket}]{\includegraphics[width=0.21\columnwidth, valign=c]{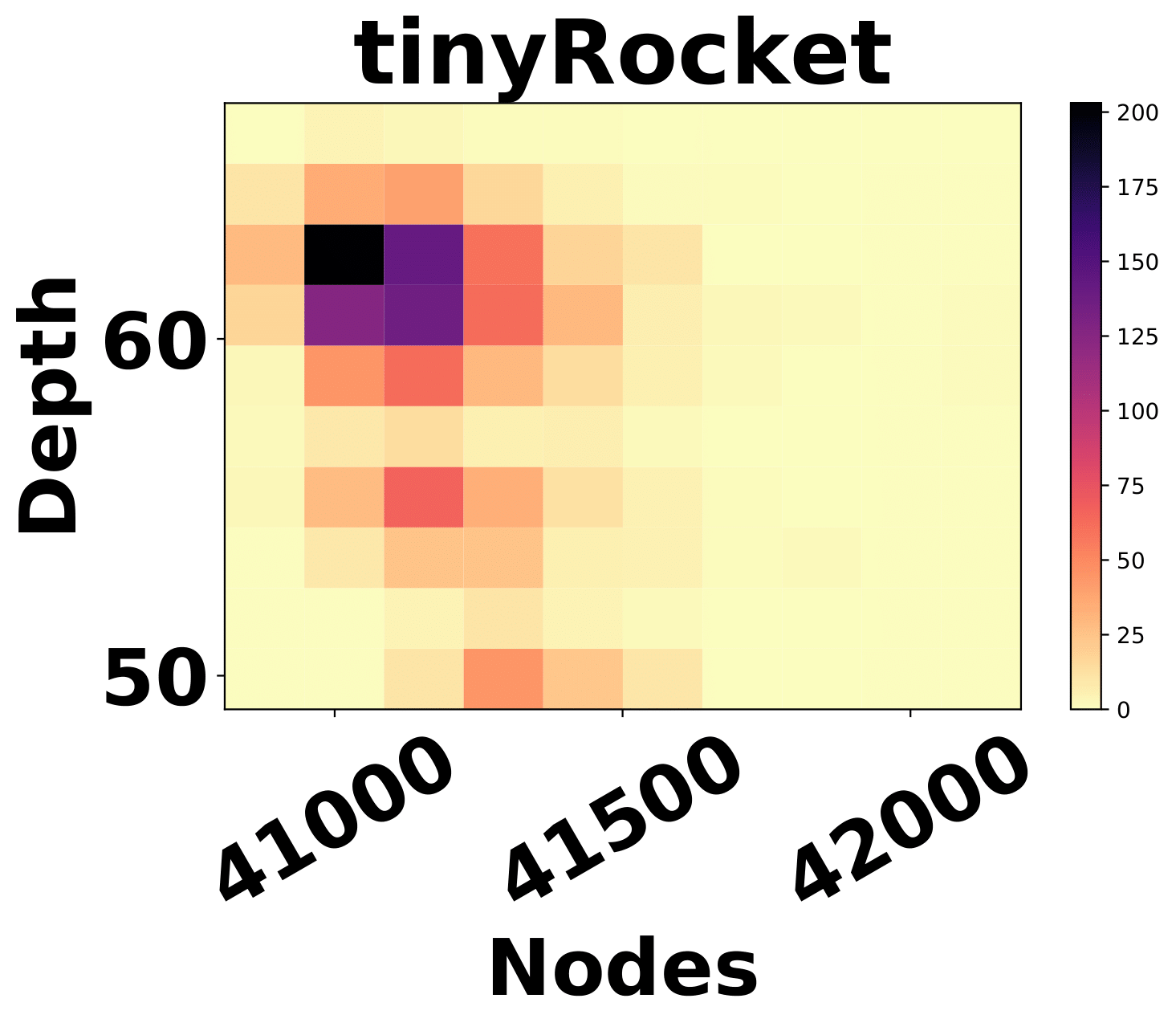}}
    \subfloat[\label{fig:usb-phy}]{\includegraphics[width=0.21\columnwidth, valign=c]{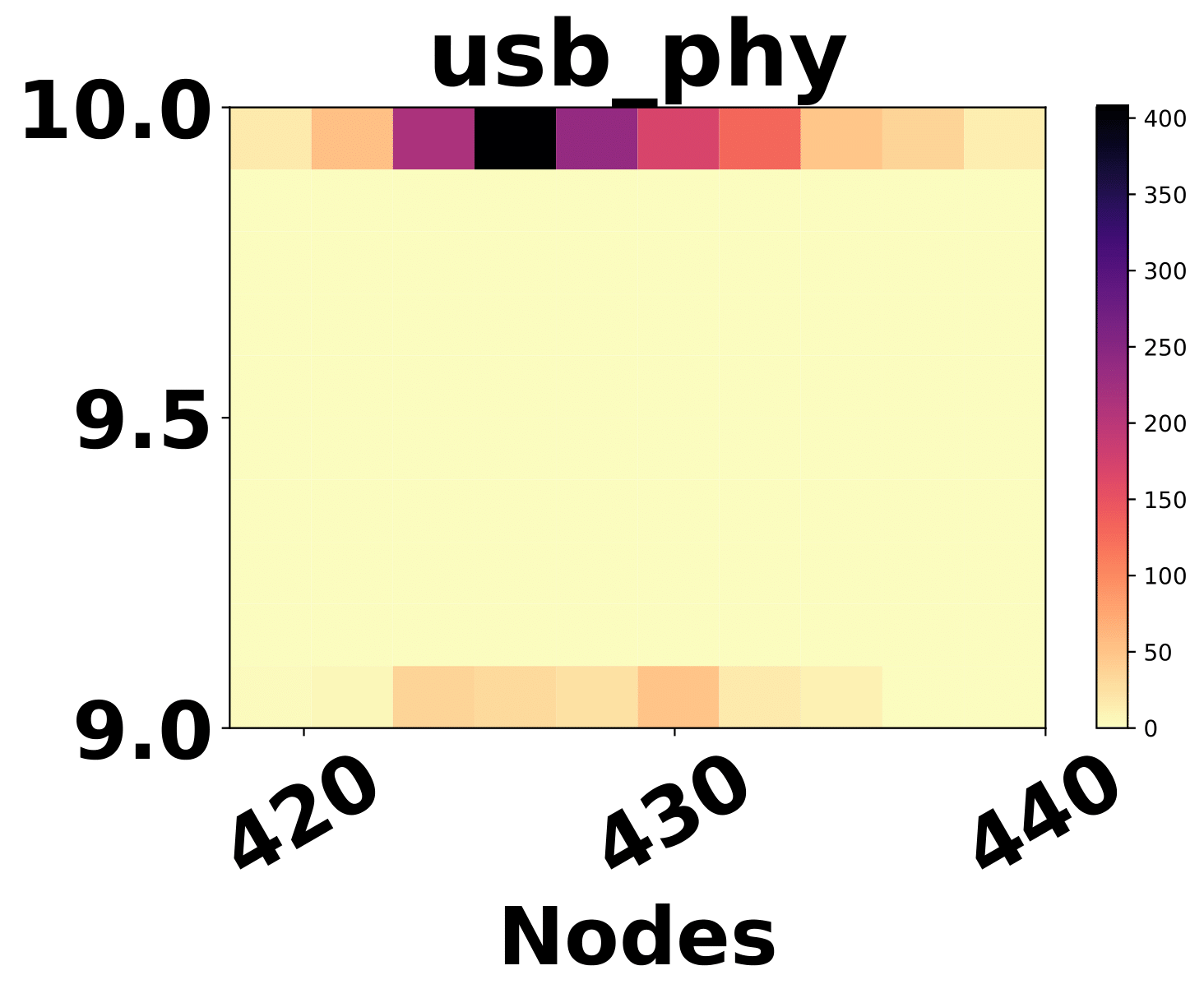}}
    \subfloat[\label{fig:vga-lcd}]{\includegraphics[width=0.2\columnwidth, valign=c]{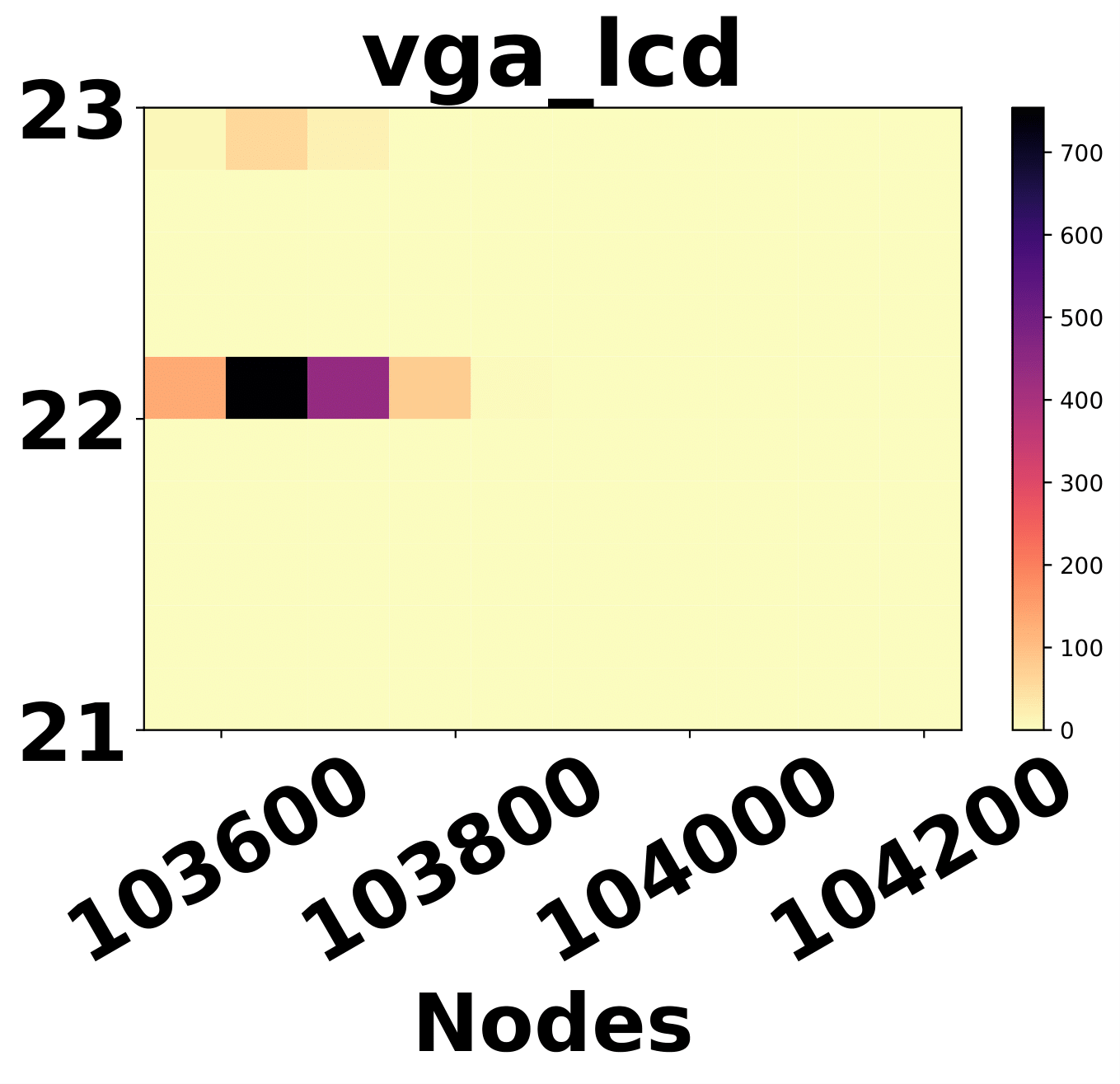}}
    \subfloat[\label{fig:wb-conmax}]{\includegraphics[width=0.2\columnwidth, valign=c]{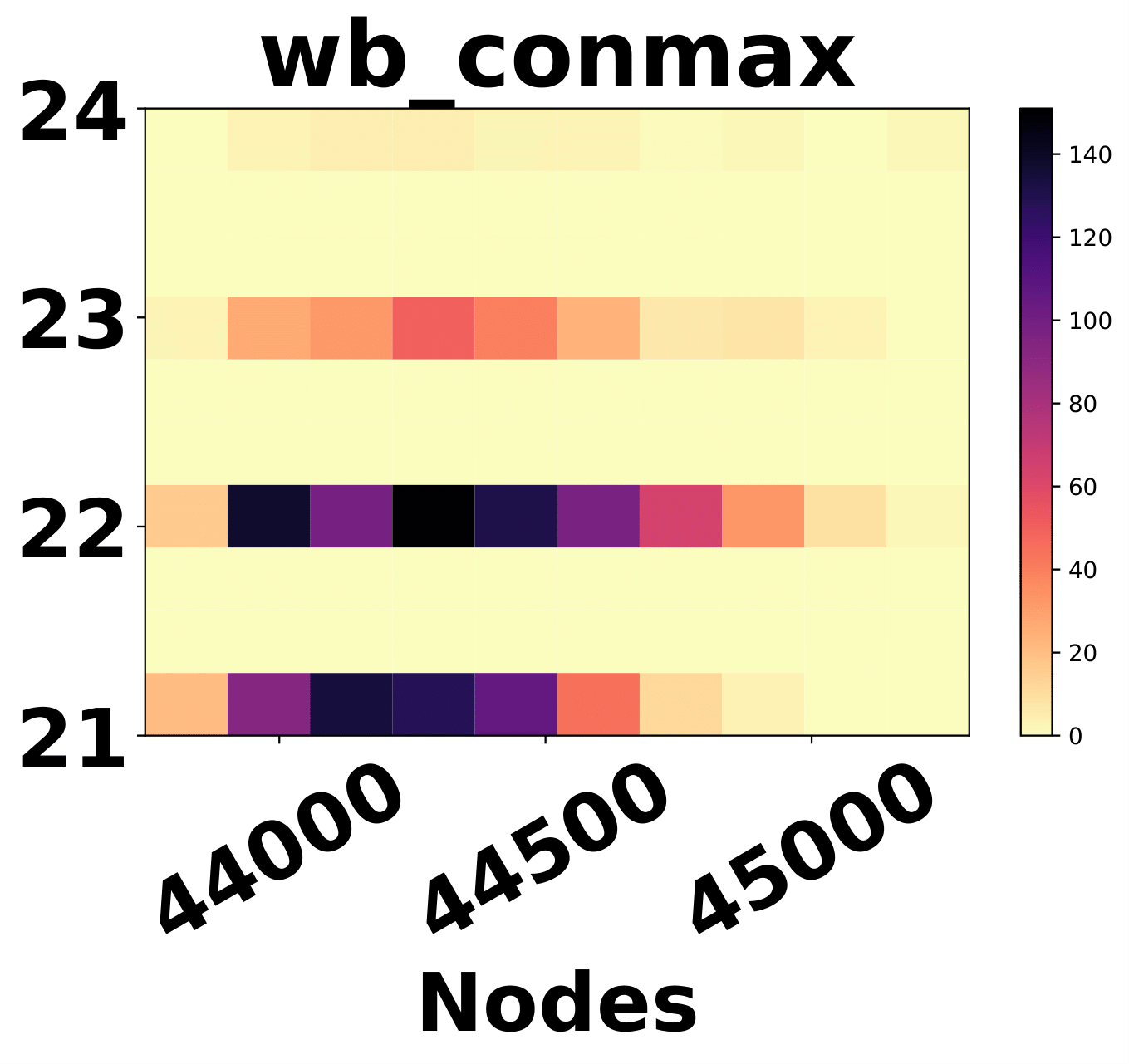}}
    \subfloat[\label{fig:wb-dma}]{\includegraphics[width=0.23\columnwidth, valign=c]{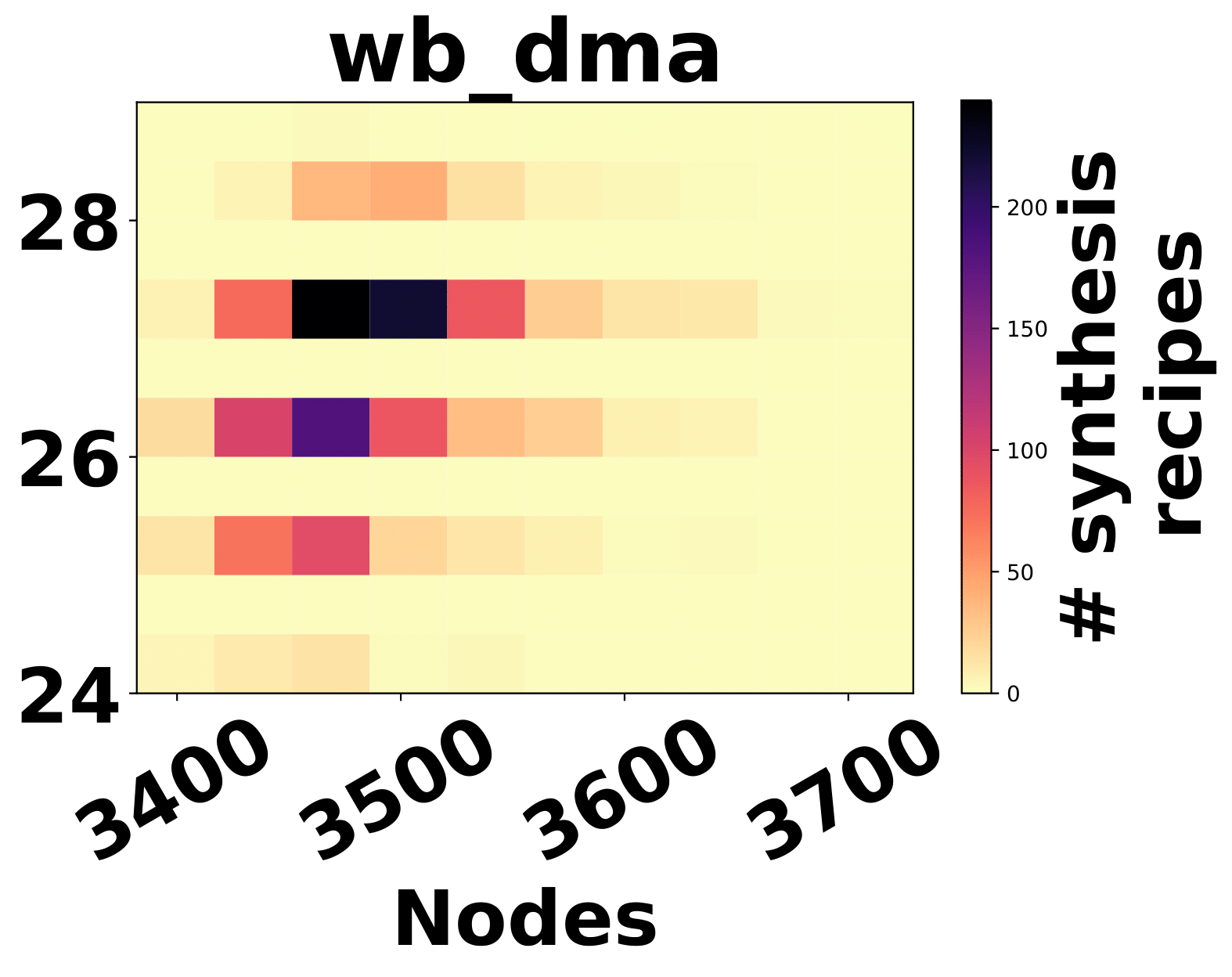}}
    \caption{Heatmaps of the synthesis quality (logic depth vs. \# of nodes) after after using 1500 synthesis recipes. Darker colors represent more recipes achieving that synthesis result.}
    \label{fig:synthesisrecipesHeatmap}
\end{figure}

% \ac{ML} models trained on such designs may yield valuable inference on real-life circuits. 
% 
% The OpenABC-D framework considers 29 open source hardware IPs. 
We run this data through $1500$ different synthesis recipes, each having 20 transformation steps, saving the AIG after each transformation step (i.e., producing 20 AIGs per recipe, per IP). 
The synthesis recipes were prepared by randomly sampling from the set of synthesis transformations, assuming a uniform distribution. 
We ran synthesis using server-grade Intel processors for $200,000+$ computation hours to generate the labeled data. 
We analyzed the top $k$ synthesis recipes by varying $k=15$, $75$ and $150$ ($1\%$ to $10\%$). We found that the similarity of the top recipes is less than 30\% (\autoref{fig:synthesis-tops}).
The synthesis recipes are diverse and relate to the graph structures and sequence of transformations in the recipes. 
% As discussed in \S\ref{subsec:datagen-mlpipeline}, each synthesis recipe run yields 20 new AIGs starting with the combinational AIG of the IP. 
We describe characteristics of each data sample and naming conventions. \\
%In our dataset creation framework, we consider $K$ unique synthesis flow each of length $L$, where $K=1500$ and $L=20$.
\textbf{Encoding the AIGs}: As described in \S\ref{subsec:datagen-graphprocessing}, we convert the \texttt{BENCH} to \texttt{GRAPHML}. We store the graph as an adjacency matrix preserving the node and edge features. We use two node-level features and numerically encode them: (1) node type and (2) number of predecessor inverter edges. For edges, we used binary encoding: 0 for original and 1 for inverted signal. \\
\textbf{Encoding the Synthesis Recipe}: We encode each synthesis transformation that ABC supports and create a vector for all $1500$ synthesis recipes. We tag and identify each synthesis recipe with a \textit{synthesis ID} as a $20$-dimensional vector. \\
\noindent\textbf{Preparing Data Samples}: We create each data sample by combining the AIG encoding and synthesis flow encoding and add label information about the sample. We mark each data sample as \texttt{designIP\_synthesisID\_stepID.pt}. \autoref{fig:datapointdesc} presents the information of each sample. The sample's name indicates the IP, the applied synthesis recipe, and the state of AIG (initial/intermediate/final), e.g., \texttt{aes\_syn149\_step15.pt} represents the intermediate AIG obtained after applying the first $15$ synthesis transformations of recipe ID $149$. 
%Therefore, step $0$ and $20$ indicate original and final optimized AIG, respectively. 
The labels include: \# of primary inputs and outputs; information about the AIG including \# of nodes, inverted edges and depth; IP function; \# of nodes in the AIG after applying the recipe; area and delay post-technology mapping. The quality of the result for every IP is different for each recipes (as illustrated in \autoref{fig:synthesisrecipesHeatmap}). 

\begin{table}[t]
\resizebox{\columnwidth}{!}{
\begin{tabular}{ccccccc}
\toprule
\multicolumn{7}{c}{\rt{\bf{ designIP\_synthesisID\_stepID.pt}}} \\
\midrule
\gt{\bf{ Graph}}   & \multicolumn{2}{c}{\vt{\bf{ Structural features}}} & \multicolumn{2}{c}{\mat{\bf{ Synthesis recipe}}} & \multicolumn{2}{c}{\bt{\bf{ Labels}}}  \\
\gt{Connectivity} & \vt{Node}  & \vt{Edge}  & \mat{Recipe ID} & \mat{Step ID}  & \bt{Individual}  & \bt{Final}  \\ \midrule
\begin{tabular}[c]{@{}c@{}}Adjacency matrix\\ of graph\end{tabular} & \begin{tabular}[c]{@{}c@{}}Type (PI/PO/AND),\\ \# Incoming inverters\\ (0/1/2)\end{tabular} & \begin{tabular}[c]{@{}c@{}}Type\\ (Buffer: 0,\\ Inverter: 1)\end{tabular} & \begin{tabular}[c]{@{}c@{}}ID (0-1499), \\ Recipe array\\ (length 20)\end{tabular} & \begin{tabular}[c]{@{}c@{}}Index of\\ intermediate\\ step of recipe\end{tabular} & \begin{tabular}[c]{@{}c@{}}\#PIs, \#POs,\#nodes,\\ \#inverters, \#edges,\\ depth,IP name\end{tabular} & \begin{tabular}[c]{@{}c@{}}\#nodes, area,\\ delay of \\ final AIG\end{tabular} \\ \bottomrule 
\end{tabular}
}
\caption{Data sample description of OpenABC-D generated dataset}
\label{fig:datapointdesc}
\end{table}

\section{Benchmarking Learning on OpenABC-D}
The OpenABC-D netlist dataset can be used as a benchmarking dataset for \ac{ML}-based EDA tasks.  
% OpenABC-D can be used in at least two ways:
An example of an \ac{ML} task for logic synthesis is supervised learning for predicting the quality of the synthesis result (e.g., \% of nodes optimized, longest path in the design). We now demonstrate the use of OpenABC-D on this task.  % after applying a synthesis recipe (i.e., the impact of a sequence of transformations given the current state of the AIG). %to obtain better results. 
% and (2) unsupervised learning of graph embeddings of the AIG using node and edge features of the graph. We demonstrate these two example tasks using OpenABC-D. %like \# of nodes and graph depth. %, followed by prediction of area and delay resulting from technology mapping steps given a synthesis recipe.

\subsection{Example Task: Predicting the Quality of a Synthesis Recipe}
\label{subsec:prototypicaltask1}
%\noindent{\bf Predicting the Quality of a Synthesis Recipe}
%(\textcolor{red}{Describe the heatmap plots and takeaway message; Another table of top k\% score post prediction})
Determining the ``best'' synthesis recipe is a challenge in logic synthesis. Logic minimization involves finding a transformation sequence that leads to an optimized AIG. Our empirical observations (\autoref{fig:synthesisrecipesHeatmap}) on the impact of synthesis recipes on different IPs show that there is no single synthesis recipe that works well on all IPs. %and this is in line with results in \cite{cunxi_dac}. 
% Experts guide the synthesis recipes to run on IPs to obtain the QoR data.
As synthesis runs are computationally costly, predicting QoR of a given synthesis recipe on a given IP can help designers find better recipes in a short time. 
For example, consider the normalized number of AIG nodes remaining after applying a synthesis recipe as the QoR.  One task variant is described next (more variants discussed in Appendix). \\
% Variants of this problem admit different data split policies. We outline three cases. \\
%Now the problem is one of predicting the QoR of a selected synthesis recipe when run on a hardware IP. In this task, we predict the number of node of AIG going to be generated post applying a synthesis flow on a given AIG. 
\noindent{\bf Variant 1: Predict QoR of unseen synthesis recipes}
Given an IP and synthesis recipe, can we predict the quality of the synthesis result? 
We train a model using all IP netlists and the AIG outputs of 1000 synthesis recipes used on those netlists (training dataset of $29\times1000 =29000$ samples). 
To evaluate the model, we test if it predicts \# of nodes in the AIG after synthesis with each of the 500 remaining unseen recipes. 
% and test-data consists of outcome of rest of the 500 synthesis recipes. The model is trained to predict the performance of unseen 500 recipes. 
This mimics the scenario when existing expert guided synthesis recipes have been tried out on IP blocks and QoR prediction is required for new synthesis recipes (to pick the best since synthesis of all options is time consuming). \\

\begin{table}[t]
\centering
\resizebox{\textwidth}{!}{%
\begin{tabular}{@{}cccccccccccc@{}}
\toprule
\multirow{2}{*}{Net} & \multicolumn{4}{c}{AIG Embedding} & \multicolumn{4}{c}{Recipe Encoding} & \multicolumn{3}{c}{FC Layers} \\ \cmidrule(lr){2-5}\cmidrule(lr){6-9}\cmidrule(lr){10-12} 
 & I & L1 & L2 & Pool & I & \# filters & kernels & stride & \# layers & architecture & dr \\ \midrule
Net1 & 4 & 128 & 128 & Max+Mean & 60 & 3 & 6,9,12 & 3 & 3 & 310-128-128-1 & 0 \\
Net2 & 4 & 64 & 64 & Max+Mean & 60 & 4 & 12,15,18,21 & 3 & 4 & 190-512-512-512-1 & 0 \\
Net3 & 4 & 64 & 64 & Max+Mean & 60 & 4 & 21,24,27,30 & 3 & 4 & 178-512-512-512-1 & 0,2 \\ \bottomrule
\end{tabular}%
}
\caption{Hyperparameters for the QoR Prediction Models. I: Input dimension, dr: dropout ratio. L1, L2: dimension of GCN layers}
\label{tab:network-architecture-params}
\end{table}

\begin{figure}[t]
\centering
%\includegraphics[width=0.7\textwidth]{figures/NeuripsGCN.png}
%\caption{Graph convolution network based IP classification/area prediction}
\subfloat[\label{fig:synthnet}Baseline model]{\includegraphics[width=0.4\columnwidth, valign=c]{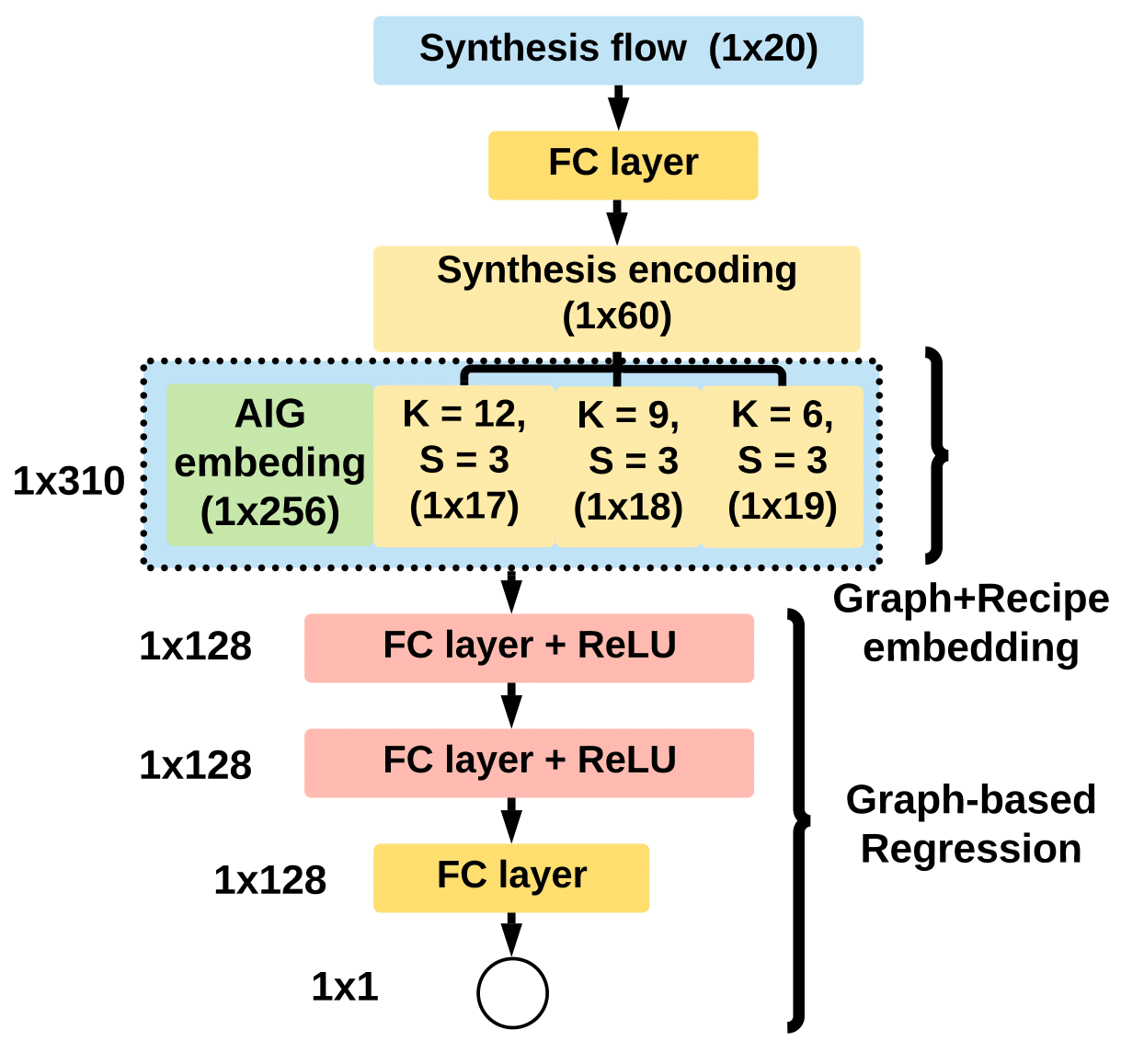}}
\subfloat[\label{fig:classnet}AIG embedding network]{\includegraphics[width=0.5\columnwidth, valign=c]{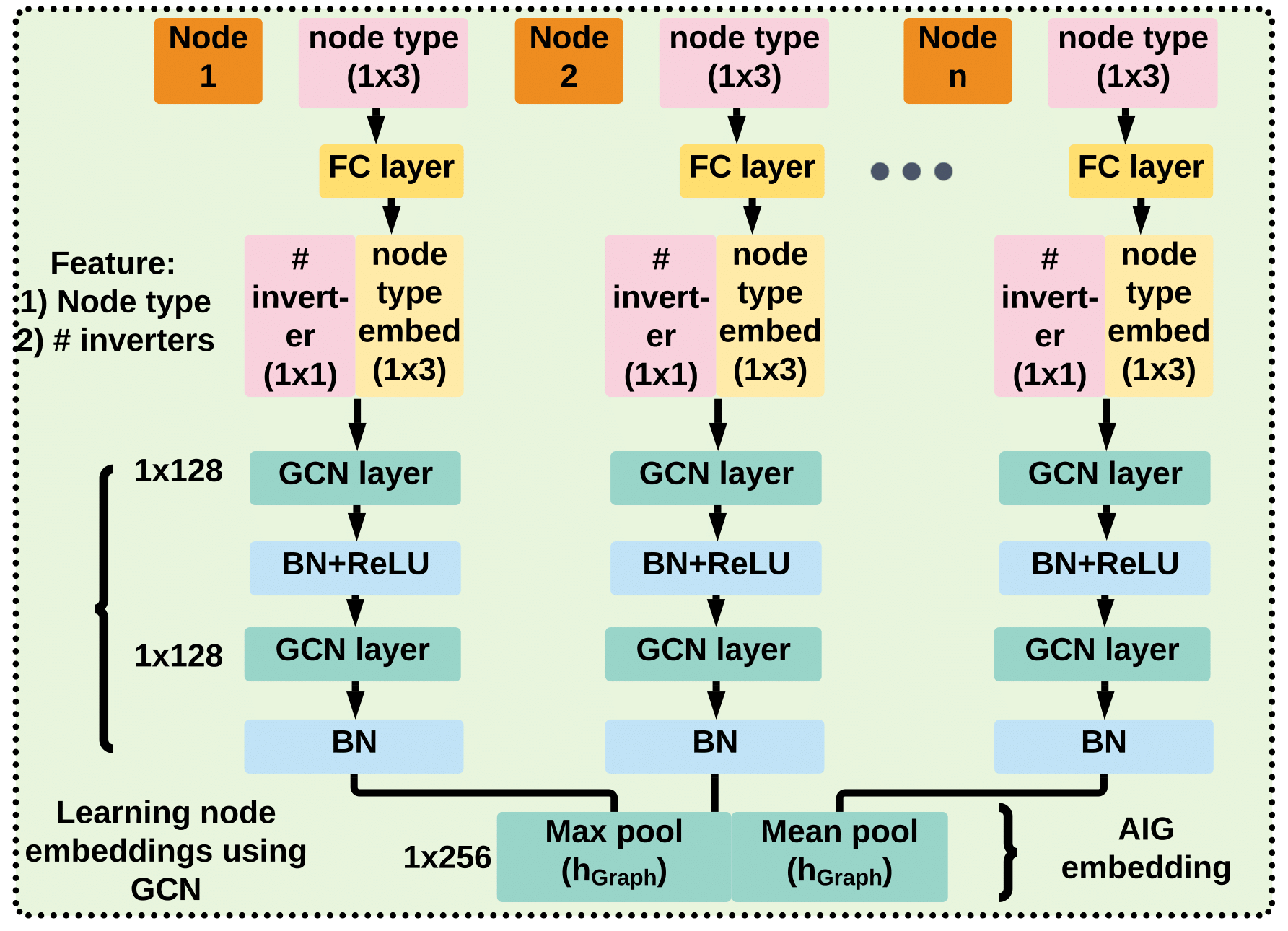}}
\caption{Graph convolution network for synthesis recipe QoR prediction} %and (b) IP classification}
\label{fig:networkarchitecture1}
\end{figure}

\noindent{\bf Model architecture and hyperparameters}: For all task variants, %train-test split strategies, 
we train the model to do graph-level predictions using a synthesis recipe encoding. 
To benchmark the performance of graph convolution networks (GCN), we considered a simple architecture (\autoref{fig:synthnet}). Here, the AIG is input to a two-layer GCN. The GCN learns node-level embeddings. Graph-level embedding is generated by a readout across all nodes in the graph. In our case, the readout is a global max pooling and average pooling. For synthesis recipe encoding, we pass the numerically encoded synthesis recipe through a linear layer and follow it with a set of filters of 1D convolution layer. The kernel size and stride length are tunable hyper-parameters. We concatenate graph-level and synthesis recipe embeddings and pass it through a set of fully connected layers to perform regression. 
In~\autoref{tab:network-architecture-params}, we show three configurations of the model (hyperparameter settings). % like kernel size and \# neurons in graph convolution and fully connected layers. 
For all settings, we used batchsize=$64$ and initial learning rate=$0.001$. 
We trained all the networks for 80 epochs. We used Adam optimizer for our experiments.  \\
% \usepackage{multirow}
\noindent{\bf Results}: We consider mean squared error (MSE) metric to evaluate model effectiveness:  
% \noindent{\bf Variant 1}: 
Net1: $0.648\pm0.05$, Net2: $0.815\pm0.02$, Net3: $0.579\pm0.02$. 
We show scatter plots of inferred vs. actual values of normalized number of nodes in the optimized netlist in Appendix D.
We observe that GCNs plus synthesis encodings consistently showed good results on most benchmarks as seen in ~\autoref{fig:net1_set1},~\ref{fig:net2_set1} and \ref{fig:net3_set1}). 
The scatter plots follow the trend of $y=x$ showing that QoR prediction of an unknown synthesis recipe is good. 
The kernel filters learn and capture properties amongst the synthesis recipe subsequences trained on which are responsible for effective performance. 
However, there is a slightly noticeable difference in scatter plots for IPs: \texttt{fir}, \texttt{iir} and \texttt{mem\_ctrl}. 
Inference of net3 is better on these IPs than net1 and net2. 
Both net1 and net2 performs bad on \texttt{fir} and \texttt{iir}; however net1 performs better than net2 on \texttt{mem\_ctrl}. 
% The test MSE scores justify the ordering of performance of the networks on the task of creating a baseline for comparison. 
\\

\section{Discussion and further insights}
\label{sec:discussion}
\noindent{\textbf{Limitations}}
Using OpenABC-D dataset, we demonstrated the use of a simple GCN model for a typical task. 
% Setting this baseline performance with a standardized dataset is an important task to set the pace for \ac{ML} research in EDA. 
In any \ac{ML}-guided \ac{EDA} pipeline, it is important that the model is trained on dataset generated from samples of hardware IPs representing the true distribution. 
Our effort of creating OpenABC-D with data generated from hardware IP of different functionalities is the first step towards this. 
We note, however, that there is room for OpenABC-D to grow. 
While OpenABC-D is more comprehensive and tailored towards use with \ac{ML} approaches compared to existing datasets, there remains a scarcity of industrial-scale open source hardware IPs. 
Therefore, it is important for an \ac{EDA} engineer to be cautious about using the inference results and perform an out-of-distribution check for any unseen hardware IP.

\begin{figure}[t]
\centering
%\includegraphics[width=0.7\textwidth]{figures/NeuripsGCN.png}
%\caption{Graph convolution network based IP classification/area prediction}
\subfloat[\label{fig:trainEmbed}Train samples (1000/class)]{\includegraphics[width=0.45\columnwidth, valign=c]{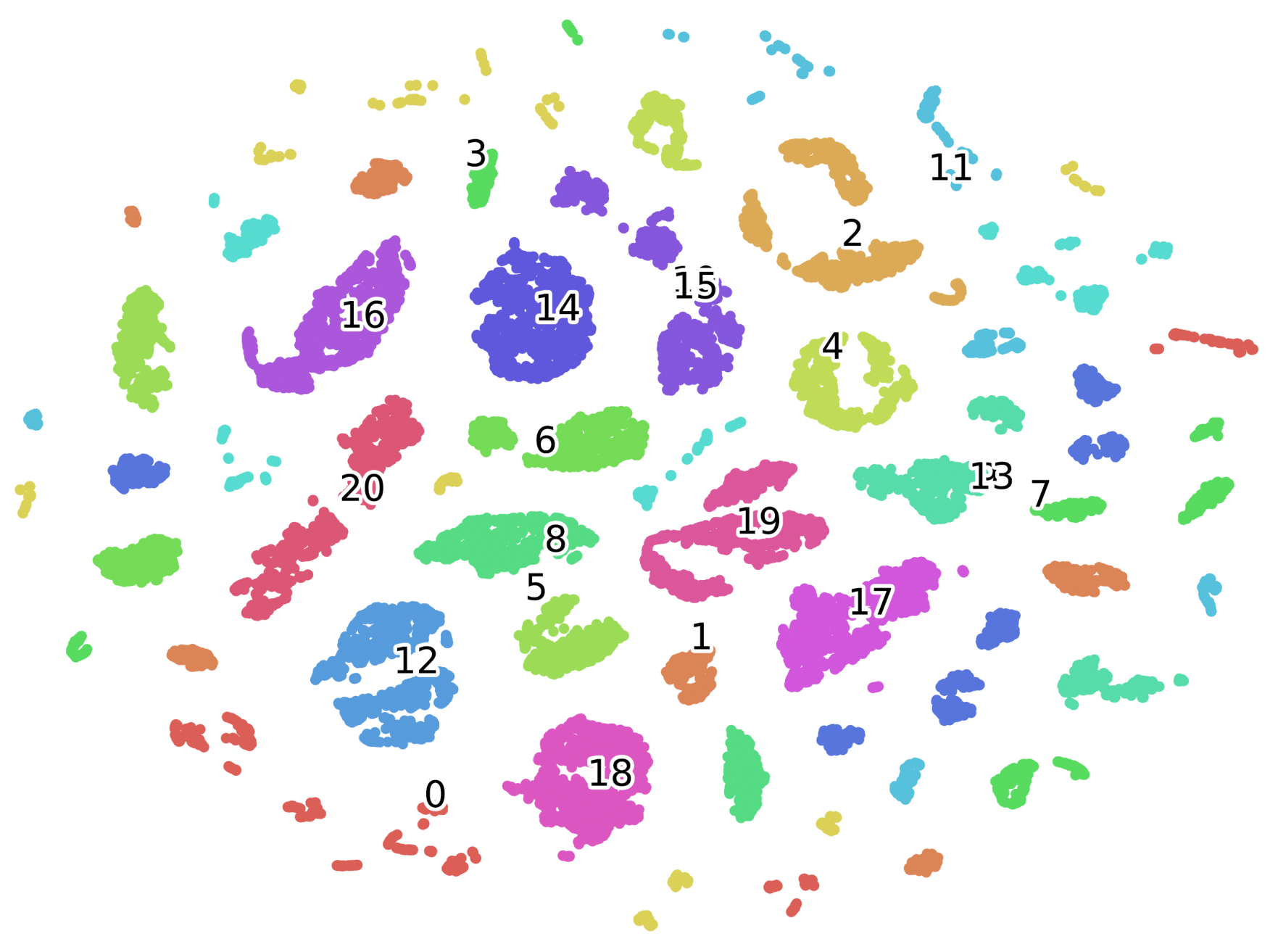}}
\subfloat[\label{fig:testEmbed}Test samples (500/class)]{\includegraphics[width=0.45\columnwidth, valign=c]{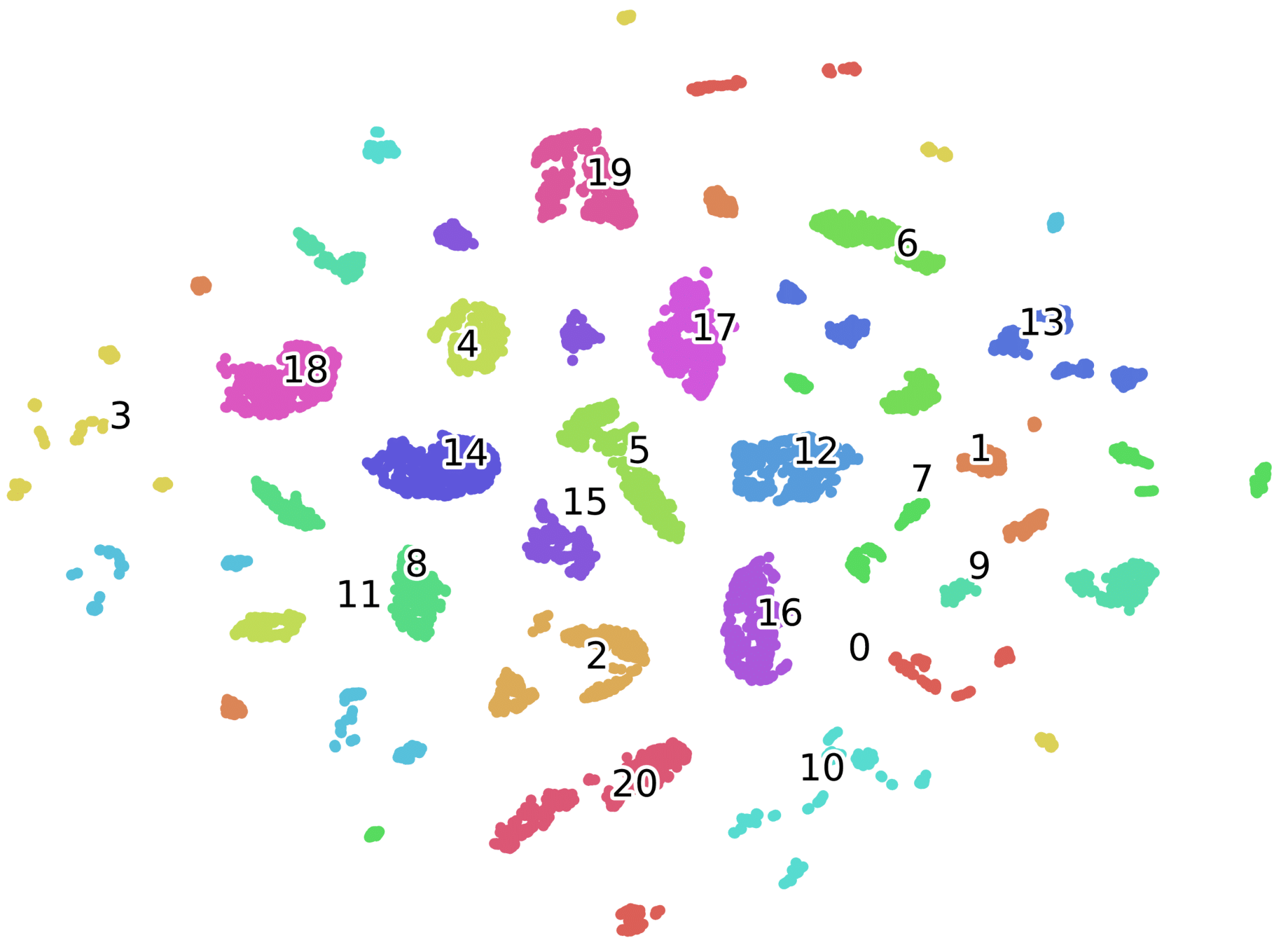}}
\caption{t-SNE plots. Labels - ac97\_ctrl: 0, sasc: 1, wb\_conmax: 2, ss\_pcm: 3, tinyRocket: 4, i2c: 5, mem\_ctrl: 6, des3\_area: 7, aes\_secworks: 8, simple\_spi: 9, pci: 10, dynamic\_node: 11, usb\_phy: 12, wb\_dma: 13, iir: 14, sha256: 15, aes: 16, fpu: 17, fir: 18, tv80: 19, spi: 20.}
\label{fig:aiggcnembeddings}
\end{figure}

\noindent{\textbf{What can a GCN learn from this data?}} \label{subsec:prototypicaltask2}
For further insights, we examined the embeddings learned by a GCN trained on our proposed data. 
The netlist embedding is the 
%is important for producing a 
lower-dimensional representation of complex IPs for tasks like area and delay prediction after technology mapping onto a cell library. 
% A GCN trained on a variety of AIG embeddings can be fine-tuned to various EDA problems that use graph structured data. 
OpenABC-D has labels like number of nodes, function of IP, and depth of DAG. They can be used to learn robust embeddings from the netlist graph. %As another example \ac{ML} task, we want to classify a netlist's functionality from its netlist structure and characteristics. \\
We trained a two-layer GCN to identify the functionality of an IP based on its netlist structure (batch size$=64$ with Adam optimizer, learning rate$=10^{-3}$, decay$=10^{-2}$, and categorical cross-entropy as loss function) and generated t-SNE plots to visualize the learned embeddings in \autoref{fig:aiggcnembeddings}. 
We obtained 98.05\% accuracy for IP classification after 35 epochs of training. Given an unknown IP, it might be possible to see if it has similarities with the known IPs, and if not, suggest when a model should be retrained with new data. 
In addition, the insights from \ac{GCN} embeddings suggest that the models can learn rich structural and functional information from \acp{AIG}. This shows that \ac{GCN} models can be pre-trained using simple self-supervised tasks (like predicting depth of \ac{AIG}) and can later be used for fine-tuning/transfer learning approaches for tasks like predicting delay of circuit.

% \noindent{\bf Predict IP Function:} An ML model for this task seeks to classify AIGs based on the IP type and analyze graph embeddings obtained from it. We performed a train-test split of 70\% AIGs of each IP as training data and the rest as inference data. The classification task uses categorical cross-entropy loss function and metric for evaluation accuracy. 
% \noindent{\bf Model Architecture:} We use a two-layer GCN with global mean pooling as readout and two fully connected layers in~\autoref{fig:classnet}. We use batch size$=64$ with Adam optimizer, learning rate$=10^{-3}$ and decay$=10^{-2}$. We use categorical cross-entropy as loss function and accuracy as evaluation metric.  % to evaluate the model. \\
% \noindent{\bf Result:} We obtained 98.05\% accuracy for IP classification after 35 epochs of training showing that GCNs learn function of AIGs using graph structures (node type and \# incoming inverted edges to nodes). The t-SNE plots of graph embeddings in \autoref{fig:aiggcnembeddings} show that GCNs can encode AIG information for the task it is trained for.

\section{Conclusion}

\ac{ML}-guided IC design needs standard datasets and baseline models to nurture open, reproducible research. OpenABC-D dataset and benchmarking models will help the ML for EDA community towards an open, standardized evaluation pipeline (like ImageNET~\cite{krizhevsky2012imagenet} for images, GLUE~\cite{wang2018glue} for language, LibriSpeech~\cite{panayotov2015librispeech} for speech). Creating such a large-scale domain-specific dataset involves substantial effort in generating labelled data, pre-processing it, and making it available in standard formats that are ingestible by standard \ac{ML} evaluation pipelines. Creating OpenABC-D dataset required 200,000+ hours of computational resources (see ~\autoref{tab:computationResource}). We discussed an important prototypical and fundamental task in logic synthesis: predicting the QoR quality of a synthesis recipe for a given IP and benchmarked simple \ac{GCN} models across a diverse set of hardware IPs. %OpenABC-D can be used by an IC designer. %An \ac{ML} engineer can apply standard ML techniques to evaluate effectiveness of approaches. 
 %This work is  the first initiative to open source dataset, framework and models to promote reproducibility and standardize \ac{ML} research in the EDA  community.

\small

\bibliographystyle{unsrt}
\bibliography{biblio}

\appendix
\newpage
\section{Checklist}

\begin{enumerate}

\item For all authors...
\begin{enumerate}
  \item Do the main claims made in the abstract and introduction accurately reflect the paper's contributions and scope?
    \answerYes{}
  \item Did you describe the limitations of your work?
    \answerYes{} (See ~\autoref{sec:discussion})
  \item Did you discuss any potential negative societal impacts of your work?
    \answerNA{}
  \item Have you read the ethics review guidelines and ensured that your paper conforms to them?
    \answerNA{}
\end{enumerate}

\item If you are including theoretical results...
\begin{enumerate}
  \item Did you state the full set of assumptions of all theoretical results?
    \answerNA{}
	\item Did you include complete proofs of all theoretical results?
    \answerNA{}
\end{enumerate}

\item If you ran experiments...
\begin{enumerate}
  \item Did you include the code, data, and instructions needed to reproduce the main experimental results (either in the supplemental material or as a URL)?
    \answerYes{}
  \item Did you specify all the training details (e.g., data splits, hyperparameters, how they were chosen)?
    \answerYes{See~\autoref{subsec:prototypicaltask1}, \ref{subsec:prototypicaltask2}}
	\item Did you report error bars (e.g., with respect to the random seed after running experiments multiple times)?
    \answerYes{}
	\item Did you include the total amount of compute and the type of resources used (e.g., type of GPUs, internal cluster, or cloud provider)?
    \answerYes{}(see~\autoref{tab:computationResource})
\end{enumerate}

\item If you are using existing assets (e.g., code, data, models) or curating/releasing new assets...
\begin{enumerate}
  \item If your work uses existing assets, did you cite the creators?
    \answerNA{}
  \item Did you mention the license of the assets?
    \answerYes{}
  \item Did you include any new assets either in the supplemental material or as a URL?
    \answerNA{}
  \item Did you discuss whether and how consent was obtained from people whose data you're using/curating?
    \answerNA{}
  \item Did you discuss whether the data you are using/curating contains personally identifiable information or offensive content?
    \answerNA{}
\end{enumerate}

\item If you used crowdsourcing or conducted research with human subjects...
\begin{enumerate}
  \item Did you include the full text of instructions given to participants and screenshots, if applicable?
    \answerNA{}
  \item Did you describe any potential participant risks, with links to Institutional Review Board (IRB) approvals, if applicable?
    \answerNA{}
  \item Did you include the estimated hourly wage paid to participants and the total amount spent on participant compensation?
    \answerNA{}
\end{enumerate}

\end{enumerate}

\newpage

\section{Benchmarks used in prior state-of-art}
\label{sec:priorSotabenchmarks}
\begin{table*}[h]
\centering
\resizebox{\textwidth}{!}{
\begin{tabular}{@{}cccc@{}}
\toprule
\toprule
\multirow{2}{*}{\begin{tabular}[c]{@{}c@{}}{\textbf{Prior}} \\ \textbf{{work}}\end{tabular}} & \multirow{2}{*}{\begin{tabular}[c]{@{}c@{}}{\textbf{Task}}\\ {\textbf{Application}}\end{tabular}} &  \multirow{2}{*}{\begin{tabular}[c]{@{}c@{}}{\textbf{Netlists used}}\\ {\textbf{Source, (\rt{\# benchmarks}})}\end{tabular}} & \multirow{2}{*}{\begin{tabular}[c]{@{}c@{}}{\textbf{\#nodes in}}\\ {\textbf{largest benchmark}}\end{tabular}} \\
 & & & \\ \midrule \midrule
\multirow{2}{*}{Yu \textit{et al.}~\cite{cunxi_dac}} & \multirow{2}{*}{Classification of synthesis flows} & \multirow{2}{*}{64bit ALU \& AES-128 (\rt{2})} & \multirow{2}{*}{44045} \\
\multirow{2}{*}{Nato \textit{et al.}~\cite{lsoracle}} & \multirow{2}{*}{Optimizer selection for minimization} &  \multirow{2}{*}{ISCAS89~\cite{iscas89} \& OpenPiton\cite{balkind2016openpiton} (\rt{5})} & \multirow{2}{*}{124565} \\
\multirow{2}{*}{Haaswijik \textit{et al.}~\cite{firstWorkDL_synth}} & \multirow{2}{*}{Optimal synthesis recipe}  & \multirow{2}{*}{DSD funcs. \& MCNC\cite{iscas85} (\rt{5})} & \multirow{2}{*}{$\le$2000} \\
\multirow{2}{*}{Hosny \textit{et al.}~\cite{drills}} & \multirow{2}{*}{Optimal synthesis recipe}  & \multirow{2}{*}{arithmetic EPFL\cite{amaru2015epfl} (\rt{10})} & \multirow{2}{*}{176938} \\
\multirow{2}{*}{Yu \textit{et al.}~\cite{cunxi_iccad}} & \multirow{2}{*}{Optimal synthesis recipe} & \multirow{2}{*}{8 DSP funcs. from VTR\cite{vtr8} (\rt{8})} & \multirow{2}{*}{30003} \\ 
\multirow{2}{*}{Zhu \textit{et al.}~\cite{mlcad_abc}} & \multirow{2}{*}{Optimal synthesis recipe} & \multirow{2}{*}{ISCAS85\cite{iscas85} (\rt{10})} & \multirow{2}{*}{2675}
\\

&&& \\ \bottomrule
\bottomrule
\end{tabular}
}
\caption{Prior work on machine learning for Logic synthesis} 
\label{table:priorWork}
\end{table*}

\section{Computational resource usage}

\begin{table}[h]
\small
\resizebox{\columnwidth}{!}{
\begin{tabular}{@{}lllll@{}}
\toprule
\toprule
\textbf{Purpose}                                                                                 & \textbf{Machine configuration}                                                                                                                                        & \textbf{\begin{tabular}[c]{@{}l@{}}\# threads \\ used\end{tabular}} & \textbf{\begin{tabular}[c]{@{}l@{}}\# hours\\ used\end{tabular}}                               & \textbf{\begin{tabular}[c]{@{}l@{}}Computing \\ hours\end{tabular}} \\ \midrule \midrule
\multirow{5}{*}{\begin{tabular}[c]{@{}l@{}}Data\\ collection\end{tabular}}                       & \begin{tabular}[c]{@{}l@{}}4x AMD EPYC 7551 32-Core Processor,\\ RAM: 504GB, Freq.: 2.0GHz\end{tabular}                                                                          & 100                                                                 & \begin{tabular}[c]{@{}l@{}}35 days\\ = 840hrs\end{tabular}                                     & 84,000                                                              \\
\cline{2-5}
                                                                                                 & \begin{tabular}[c]{@{}l@{}}Dual Intel(R) Xeon(R) CPU E5-2650 v3\\ RAM: 502GB, Freq. 1.8GHz\end{tabular}                                                                            & 80                                                                  & \begin{tabular}[c]{@{}l@{}}40 days\\ = 960hrs\end{tabular}                                     & 76,800                                                              \\
                                                                                                 \cline{2-5}
                                                                                                 & \begin{tabular}[c]{@{}l@{}}Intel(R) Xeon(R) CPU E5-2640\\ RAM: 252GB, Freq. 2.5GHz\end{tabular}                                                                                    & 40                                                                  & \begin{tabular}[c]{@{}l@{}}35 days\\ = 840hrs\end{tabular}                                     & 33,600                                                              \\
                                                                                                 \cline{2-5}
                                                                                                 & \begin{tabular}[c]{@{}l@{}}AMD Ryzen Threadripper 2920X \\ 12-Core Processor\\ RAM: 32GB, Freq: 2.32GHz\end{tabular}                                                               & 16                                                                  & \begin{tabular}[c]{@{}l@{}}7 days\\ = 168hrs\end{tabular}                                      & 2688                                                                \\
                                                                                                 \cline{2-5}
                                                                                                 & \begin{tabular}[c]{@{}l@{}}8x Intel(R) Core(TM) i7-6700\\ RAM: 96GB, Freq: 2.10GHz (avg.)\end{tabular}                                                                             & 40                                                                  & \begin{tabular}[c]{@{}l@{}}55 days\\ = 1320hrs\end{tabular}                                    & 52,800                                                              \\ \hline
\multirow{2}{*}{\begin{tabular}[c]{@{}l@{}}Model:\\ \\ Training \\ and\\ inference\end{tabular}} & \begin{tabular}[c]{@{}l@{}}Lambda-quad\\ Intel(R) Core(TM) i9-7920X CPU,\\ RAM: 126GB, Freq: 2.5GHz\\ \\ \\ Batchsize: 4\\ GPU: GTX 1080i, 11GB VRAM\end{tabular}                   & -                                                                   & \begin{tabular}[c]{@{}l@{}}QoR \\ prediction:\\ 48 hrs\\ Classification:\\ 18 hrs\end{tabular} & -                                                                   \\
\cline{2-5}
                                                                                                 & \begin{tabular}[c]{@{}l@{}}Greene (High performance computing)\\ 2x Intel Xeon Platinum 8268\\ RAM: 369GB, Freq. 2.9GHz\\ \\ Batchsize: 64\\ GPU: RTX 8000, 48GB VRAM\end{tabular} & -                                                                   & \begin{tabular}[c]{@{}l@{}}QoR \\ prediction:\\ \~36 hrs\\ Classification:\\ \~12 hrs\end{tabular}  & -                                                                   \\ \bottomrule \bottomrule
\end{tabular}
}
\caption{Computational resource usage breakdown for data generation and benchmarking models}
\label{tab:computationResource}
\end{table}
\section{Additional Results on GCN Models using OpenABC-D (QoR prediction)}

In addition to predicting the QoR given synthesis using an unseen recipe (Variant 1, \S\ref{subsec:prototypicaltask1}), we also considered two further task variants. 

\noindent{\bf Variant 2: Predicting QoR of synthesizing unseen IPs}
We train a model on the QoR from synthesizing a set of smaller IPs and evaluate the model on its ability to predict the QoR of synthesizing the unseen larger IPs, given the IP and a synthesis recipe. 
% The train-test split is done on the IPs. Here, we consider small-size IPs (small number of nodes) for training and large-size IPs for inference. 
This mimics real-world problems where synthesis runs on large IPs take weeks to complete while synthesis runs on small IPs are possible within a short time. 
The motivation is to understand whether models trained with data generated from small-size IPs can be used to meaningfully predict on large-size IPs. We consider 16 smaller IPs during training and 8 larger IPs during inference.

\begin{figure}
    \centering
    \subfloat[\label{fig:ac97_ctrl}]{\includegraphics[width=0.2\columnwidth, valign=c]{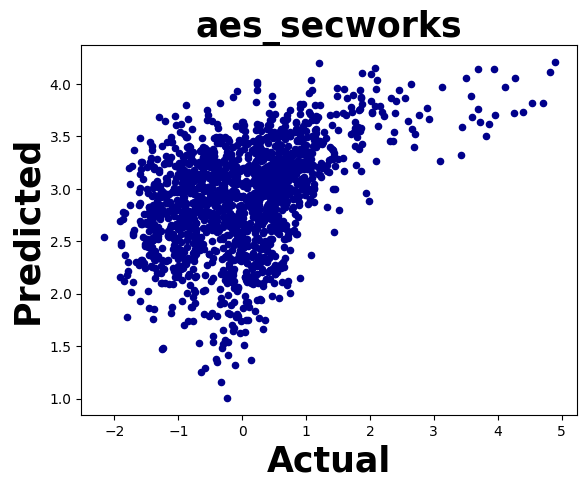}}
    \subfloat[\label{fig:aes_secworks}]{\includegraphics[width=0.2\columnwidth, valign=c]{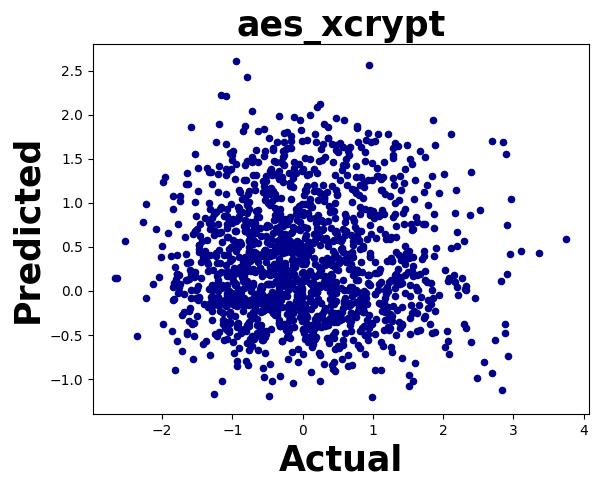}}
    \subfloat[\label{fig:aes_xcrypt}]{\includegraphics[width=0.2\columnwidth, valign=c]{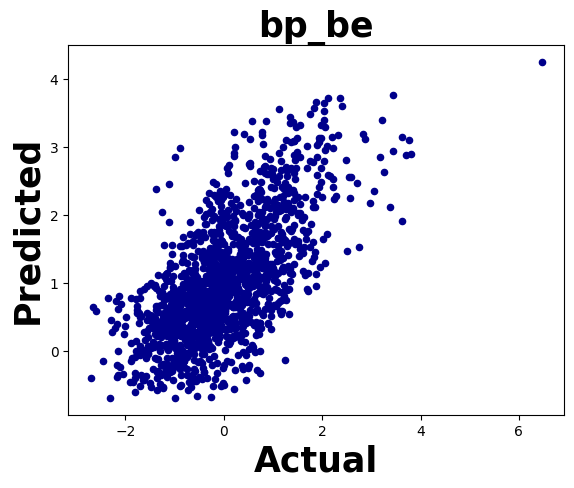}}
    \subfloat[\label{fig:aes}]{\includegraphics[width=0.2\columnwidth, valign=c]{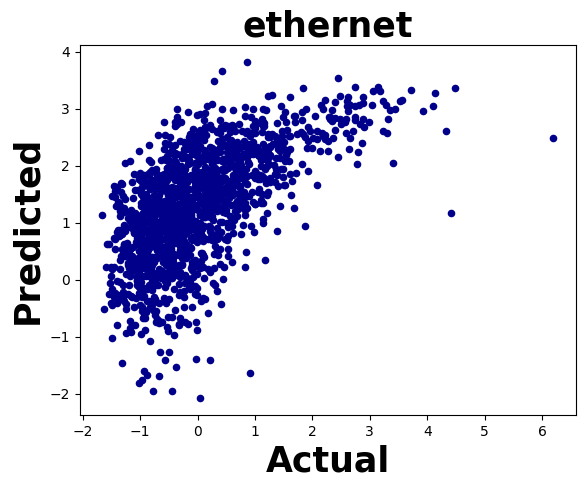}}
 \\ \vspace*{-0.1in}
    \subfloat[\label{fig:ac97_ctrl}]{\includegraphics[width=0.2\columnwidth, valign=c]{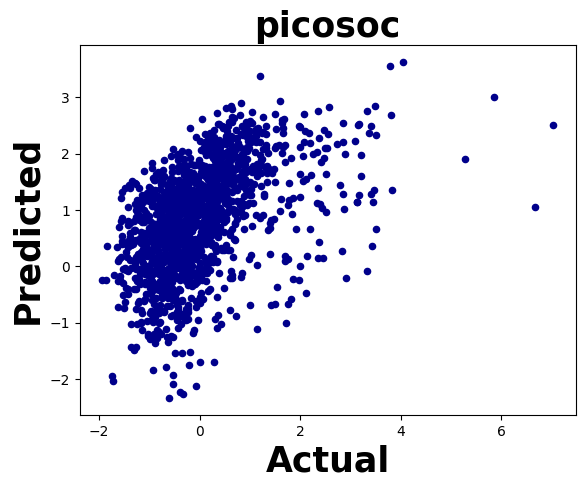}}
    \subfloat[\label{fig:aes_secworks}]{\includegraphics[width=0.2\columnwidth, valign=c]{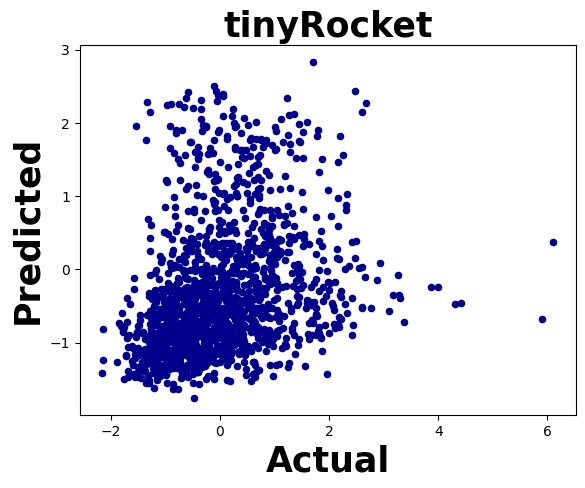}}
        \subfloat[\label{fig:ac97_ctrl}]{\includegraphics[width=0.2\columnwidth, valign=c]{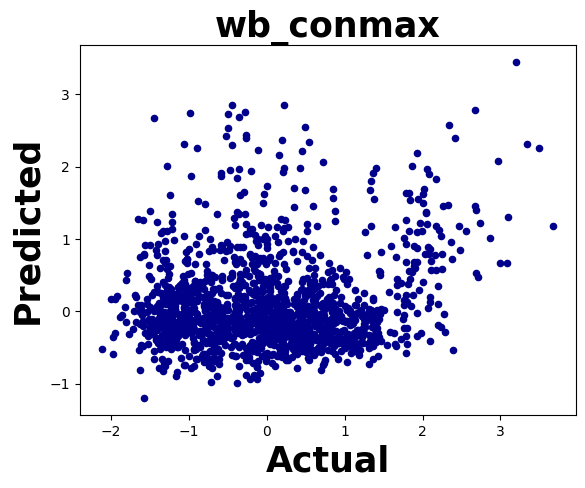}}
    \subfloat[\label{fig:bp_be}]{\includegraphics[width=0.2\columnwidth, valign=c]{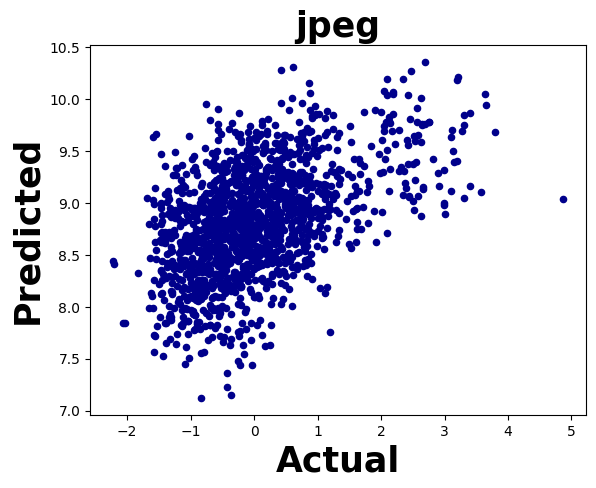}}
   \caption{Net 1 for QoR Task Variant 2 (Unseen IP)}
    \label{fig:net1_set2}
\end{figure}

\begin{figure}
    \centering
    \subfloat[\label{fig:ac97_ctrl}]{\includegraphics[width=0.2\columnwidth, valign=c]{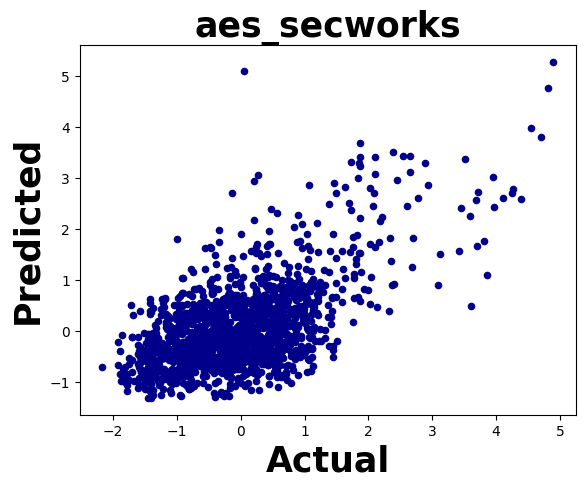}}
    \subfloat[\label{fig:aes_secworks}]{\includegraphics[width=0.2\columnwidth, valign=c]{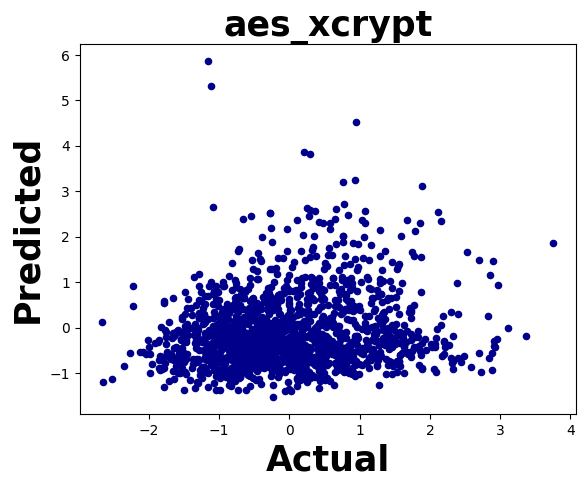}}
    \subfloat[\label{fig:aes_xcrypt}]{\includegraphics[width=0.2\columnwidth, valign=c]{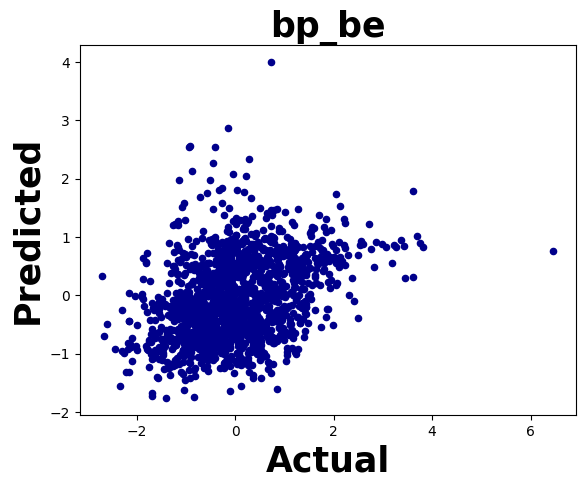}}
    \subfloat[\label{fig:aes}]{\includegraphics[width=0.2\columnwidth, valign=c]{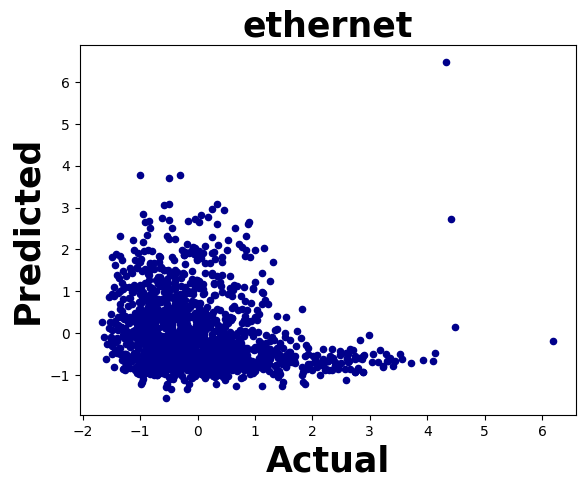}}
\\ \vspace*{-0.1in}
    \subfloat[\label{fig:ac97_ctrl}]{\includegraphics[width=0.2\columnwidth, valign=c]{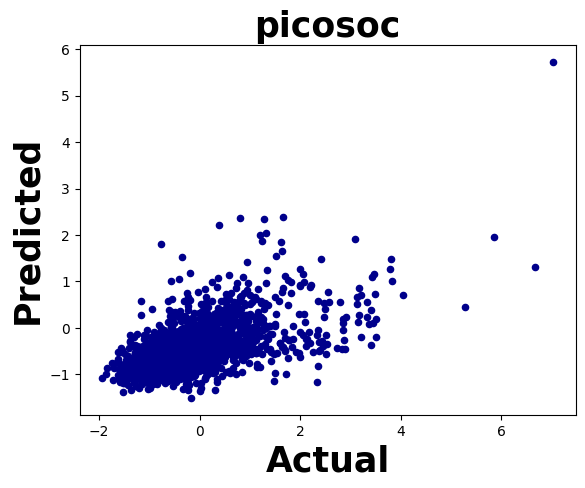}}
    \subfloat[\label{fig:aes_secworks}]{\includegraphics[width=0.2\columnwidth, valign=c]{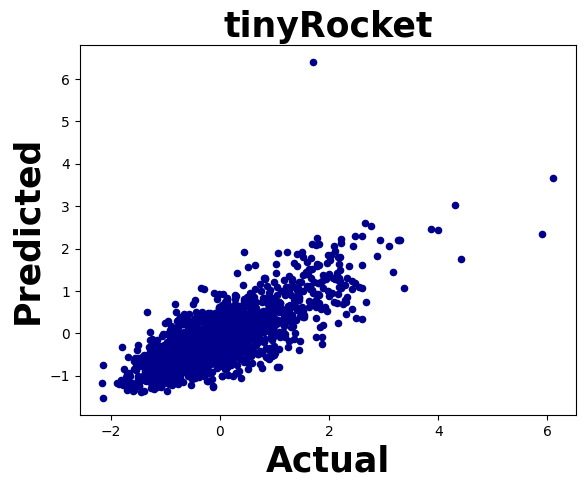}}
        \subfloat[\label{fig:ac97_ctrl}]{\includegraphics[width=0.2\columnwidth, valign=c]{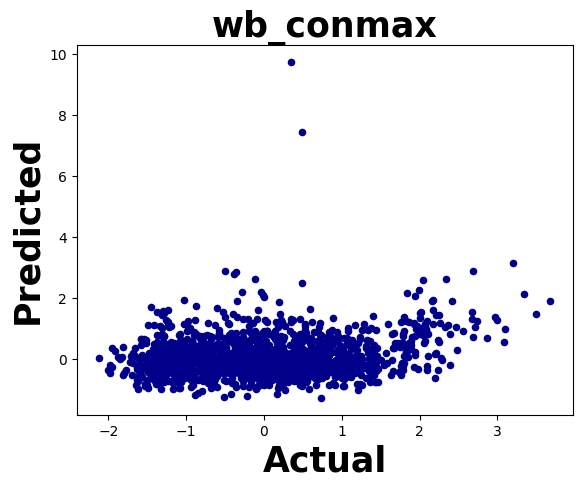}}
    \subfloat[\label{fig:bp_be}]{\includegraphics[width=0.2\columnwidth, valign=c]{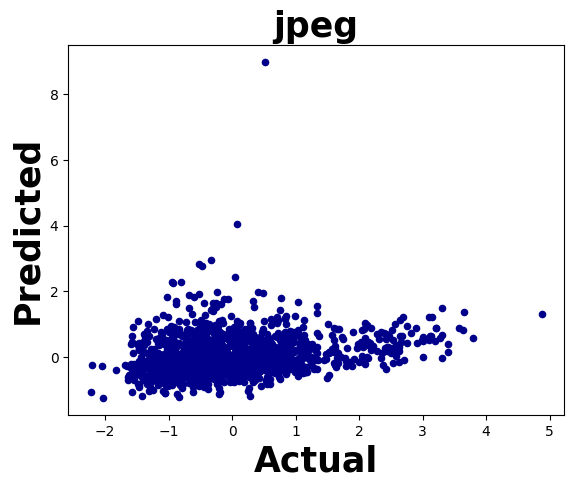}} 
   \caption{Net 2 for QoR Task Variant 2 (Unseen IP)}
    \label{fig:net2_set2}
\end{figure}

\begin{figure}
    \centering
    \subfloat[\label{fig:ac97_ctrl}]{\includegraphics[width=0.2\columnwidth, valign=c]{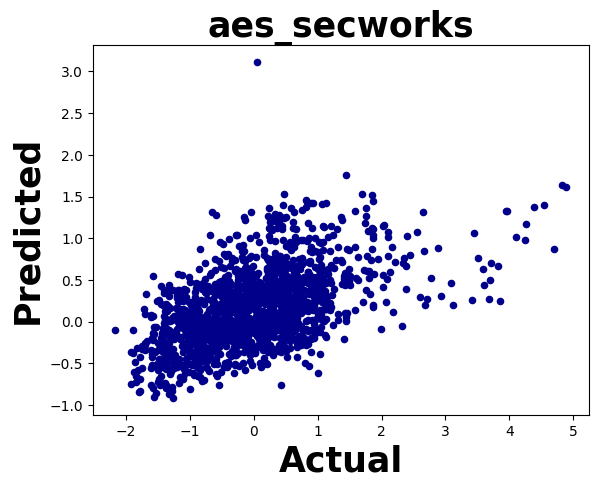}}
    \subfloat[\label{fig:aes_secworks}]{\includegraphics[width=0.2\columnwidth, valign=c]{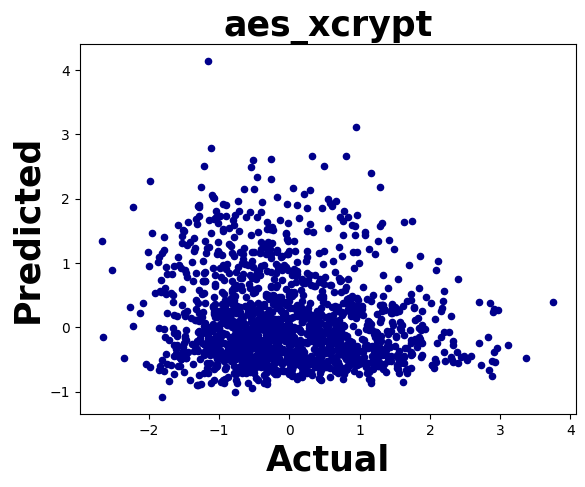}}
    \subfloat[\label{fig:aes_xcrypt}]{\includegraphics[width=0.2\columnwidth, valign=c]{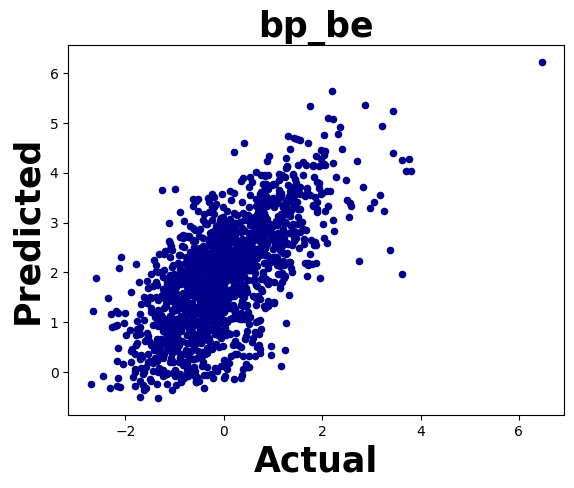}}
    \subfloat[\label{fig:aes}]{\includegraphics[width=0.2\columnwidth, valign=c]{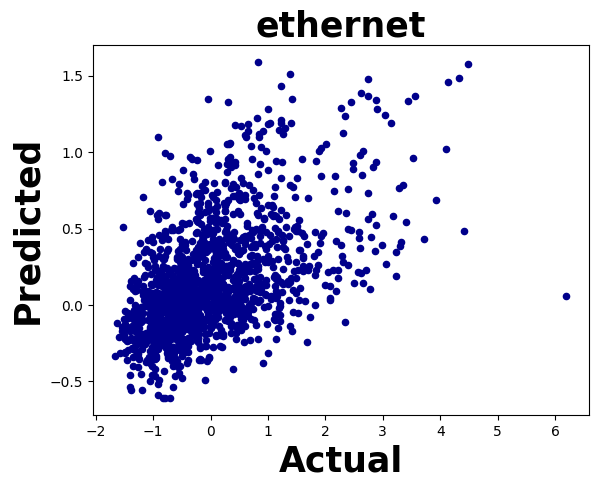}}
 \\ \vspace*{-0.1in}
    \subfloat[\label{fig:ac97_ctrl}]{\includegraphics[width=0.2\columnwidth, valign=c]{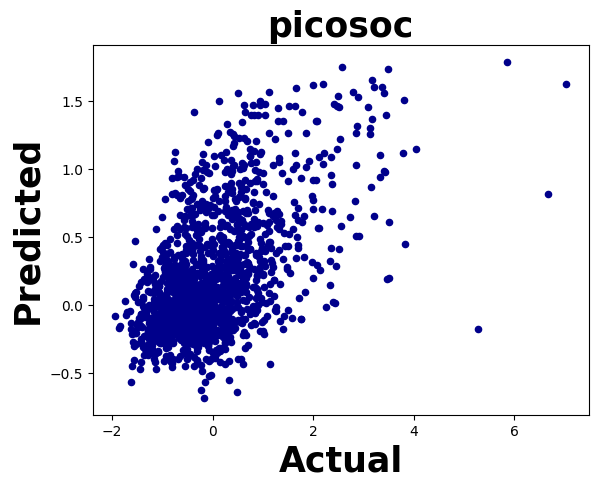}}
    \subfloat[\label{fig:aes_secworks}]{\includegraphics[width=0.2\columnwidth, valign=c]{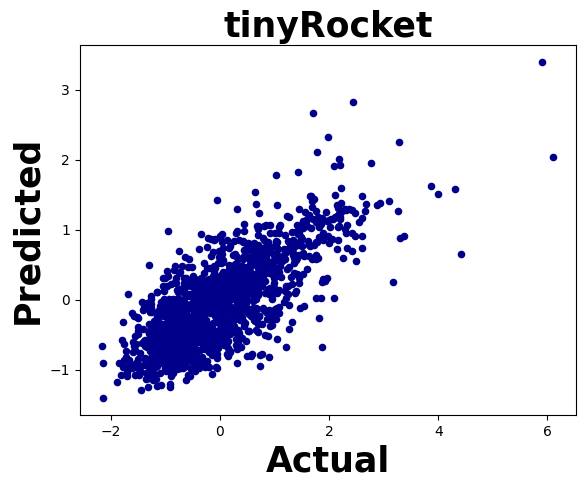}}
        \subfloat[\label{fig:ac97_ctrl}]{\includegraphics[width=0.2\columnwidth, valign=c]{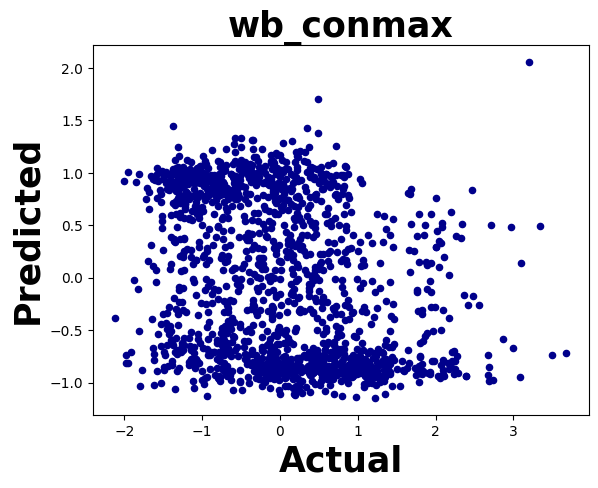}}
    \subfloat[\label{fig:bp_be}]{\includegraphics[width=0.2\columnwidth, valign=c]{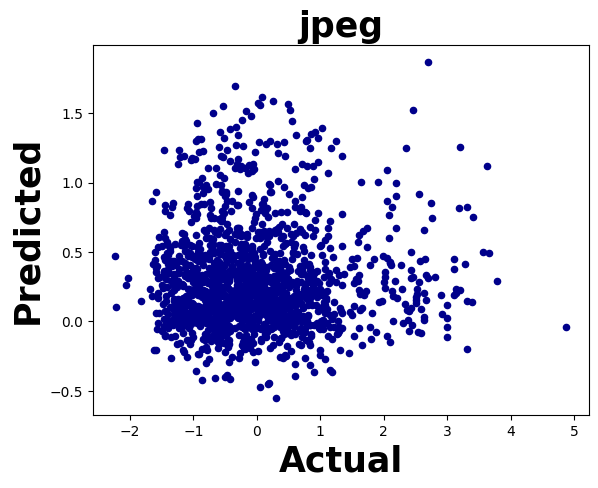}}
   \caption{Net 2 for QoR Task Variant 2 (Unseen IP)}
    \label{fig:net3_set2}
\end{figure}

\noindent{\bf Results}: Using MSE as metric for comparison on test dataset, we presented the results in ~\autoref{tab:qorsynthesisrecipe}.
%MSE -- Net1: $10.59\pm2.78$, Net2: $1.236\pm0.15$, Net3: $1.47\pm0.14$.
In train-test split strategy where IPs are unknown, the performance varied across test IPs for different networks (see~\autoref{fig:net1_set2},\ref{fig:net2_set2} and \ref{fig:net3_set2}). For IPs like \texttt{aes\_xcrypt} and \texttt{wb\_conmax}, the inference results are consistently poor indicating AIG embeddings are different from training data AIG embedding. Inference results on like \texttt{bp\_be}, \texttt{tinyRocket}, and \texttt{picosoc} are close to  QoR values across networks indicating the network has learnt from graph structures of training data IPs. 

% The goal is to infer QoR of synthesis recipes on unseen netlists. \\
\noindent{\bf Variant 3: Predicting QoR on unseen IP-Synthesis Recipe Combination}
%In this train-test split up, we randomly pick 70\% of the entire dataset as training and rest for inference purpose. In this setting, test data consist of IP with a synthesis flow not seen together in training data but as separate entity as part of different IP synthesis flow combination.
% In this train-test split, for each IP we 
We train a model on the QoR achieved from a random pick of 70\% of the synthesis recipes across all IPs and test the model's ability to predict the QoR given  % as training set and rest for inference. 
% The test data consists of IP with a synthesis recipe not seen together during training. 
an unseen IP-recipe pair. 
% They maybe seen during training as a different IP-synthesis recipe combination. 
In this use case, experts develop synthesis recipes for specific IPs and the user is interested in assessing the performance of these recipes on other IPs without running the synthesis recipes on them.

\noindent{\bf Results}: \autoref{tab:qorsynthesisrecipe} shows the results on three baseline networks.
%MSE -- Net1: $0.588\pm0.04$, Net2: $0.538\pm0.01$, Net3: $0.536\pm0.03$.
The trained model has observed IPs and synthesis recipes as standalone entities as part of training data. Inference of networks are expected to be better than previous two data-split strategies and results in \autoref{tab:qorsynthesisrecipe} confirm this. Performance of network 3 is better indicating a large kernel size helps network learn more information about effectiveness of synthesis recipes. This  is similar to results obtained in \cite{cunxi_dac} and therefore sets a baseline for GCN-based networks on such tasks.

\begin{figure}[h]
    \centering
    \subfloat[\label{fig:ac97_ctrl}]{\includegraphics[width=0.2\columnwidth, valign=c]{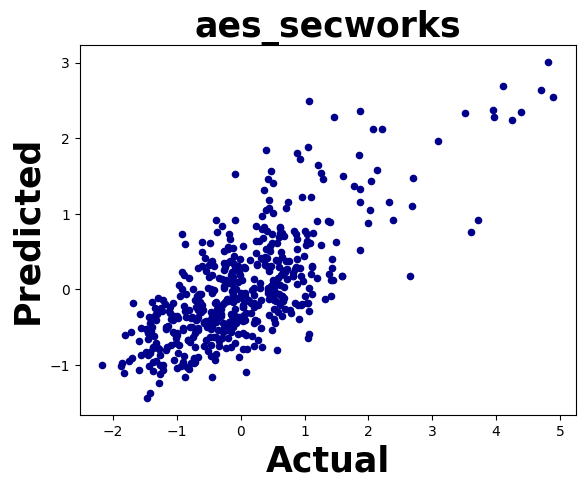}}
    \subfloat[\label{fig:aes_secworks}]{\includegraphics[width=0.2\columnwidth, valign=c]{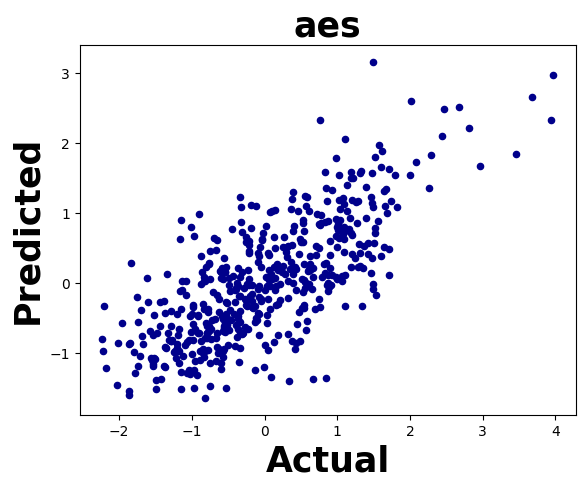}}
    \subfloat[\label{fig:aes_xcrypt}]{\includegraphics[width=0.2\columnwidth, valign=c]{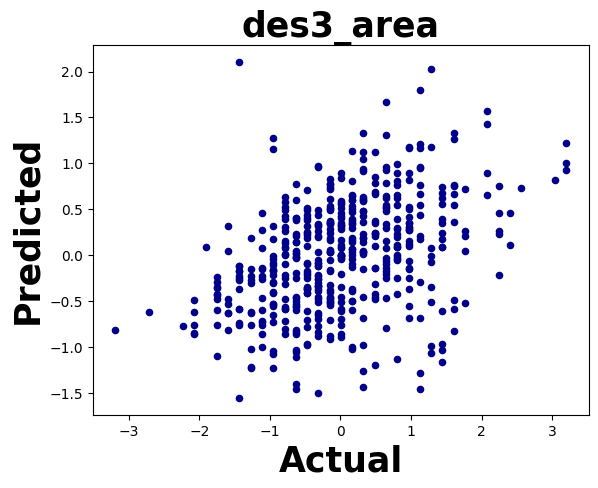}}
    \subfloat[\label{fig:aes}]{\includegraphics[width=0.2\columnwidth, valign=c]{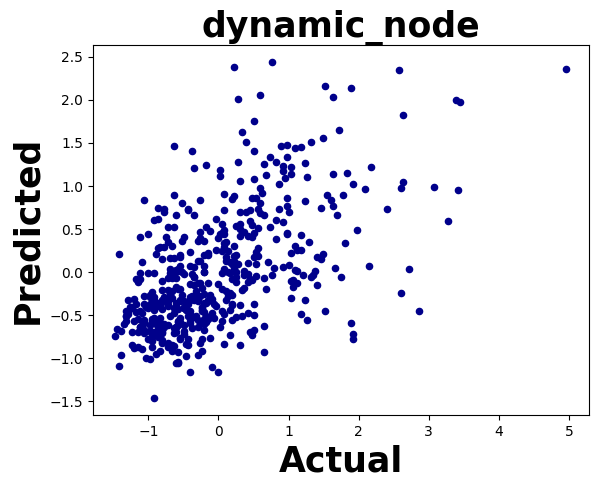}}
    \subfloat[\label{fig:bp_be}]{\includegraphics[width=0.2\columnwidth, valign=c]{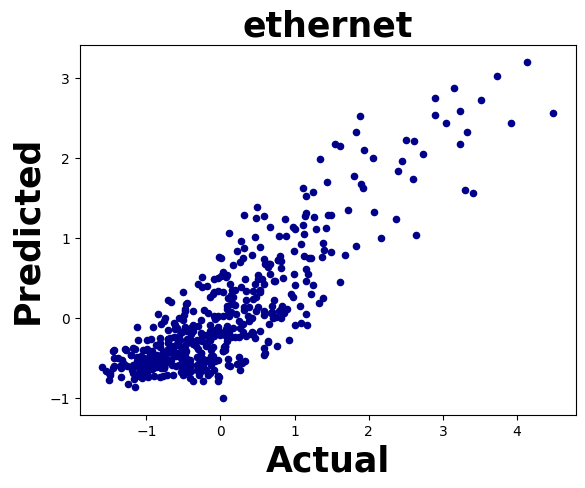}} \\ \vspace*{-0.1in}
    \subfloat[\label{fig:ac97_ctrl}]{\includegraphics[width=0.2\columnwidth, valign=c]{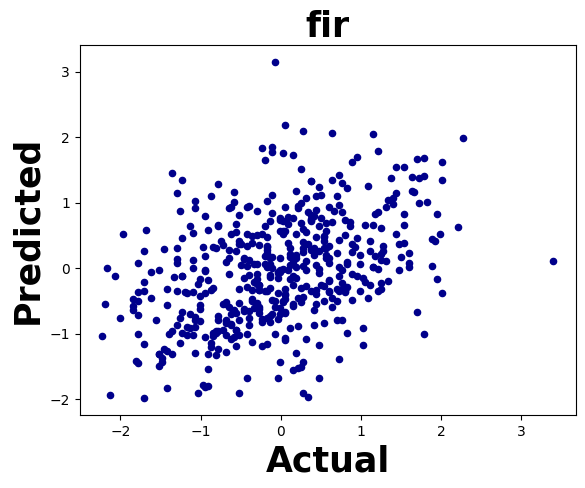}}
    \subfloat[\label{fig:aes_secworks}]{\includegraphics[width=0.2\columnwidth, valign=c]{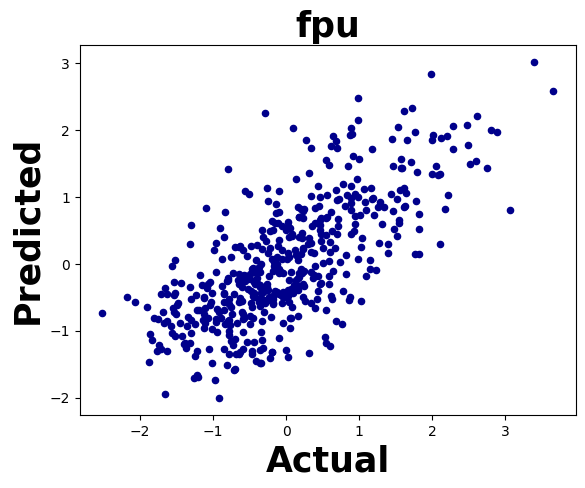}}
    \subfloat[\label{fig:aes_xcrypt}]{\includegraphics[width=0.2\columnwidth, valign=c]{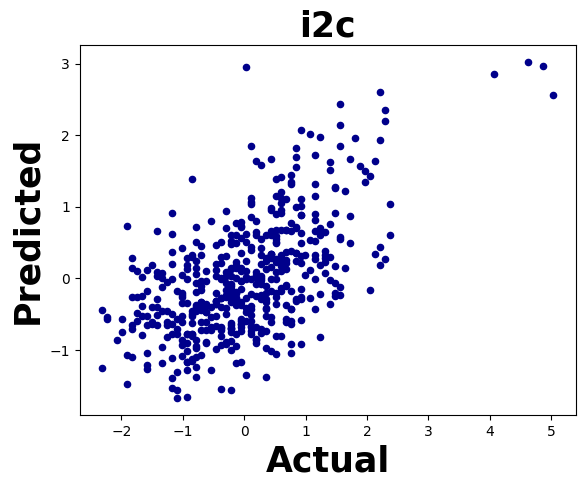}}
    \subfloat[\label{fig:aes}]{\includegraphics[width=0.2\columnwidth, valign=c]{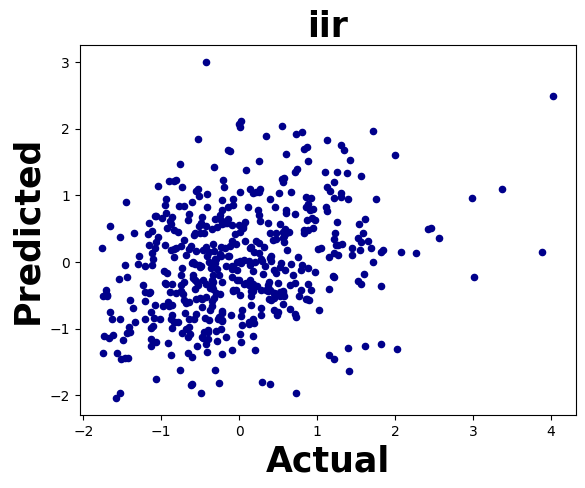}}
    \subfloat[\label{fig:bp_be}]{\includegraphics[width=0.2\columnwidth, valign=c]{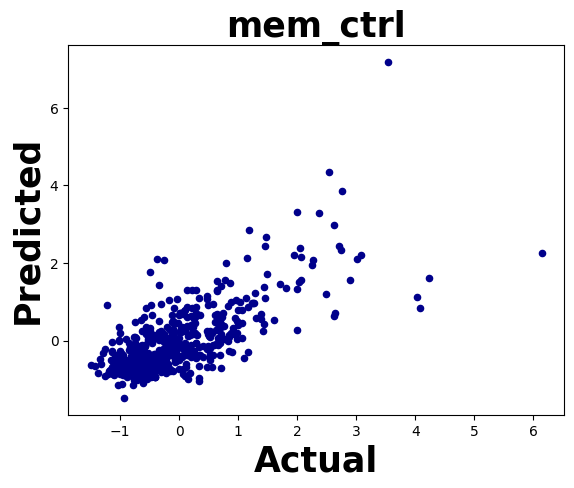}} \\\vspace*{-0.1in}
    \subfloat[\label{fig:ac97_ctrl}]{\includegraphics[width=0.2\columnwidth, valign=c]{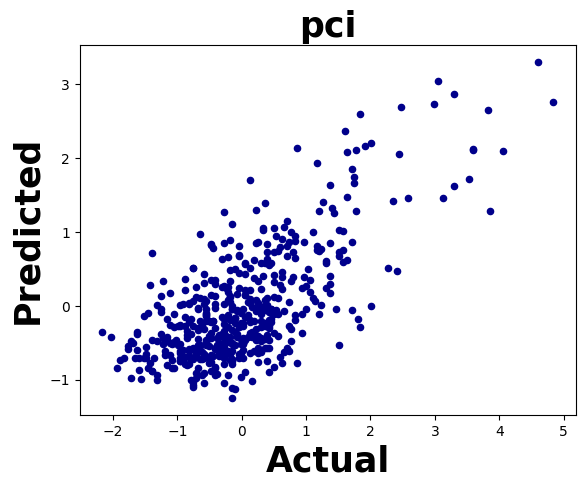}}
    \subfloat[\label{fig:aes_xcrypt}]{\includegraphics[width=0.2\columnwidth, valign=c]{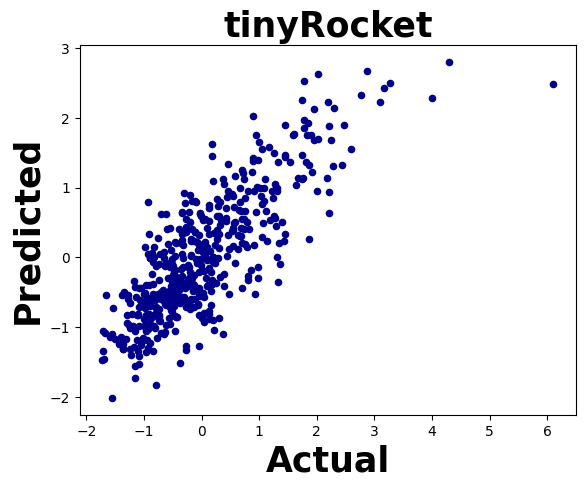}}
    \subfloat[\label{fig:aes}]{\includegraphics[width=0.2\columnwidth, valign=c]{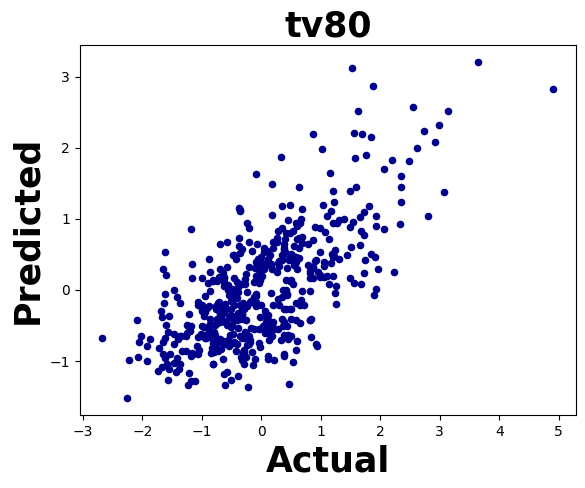}}
    \subfloat[\label{fig:bp_be}]{\includegraphics[width=0.2\columnwidth, valign=c]{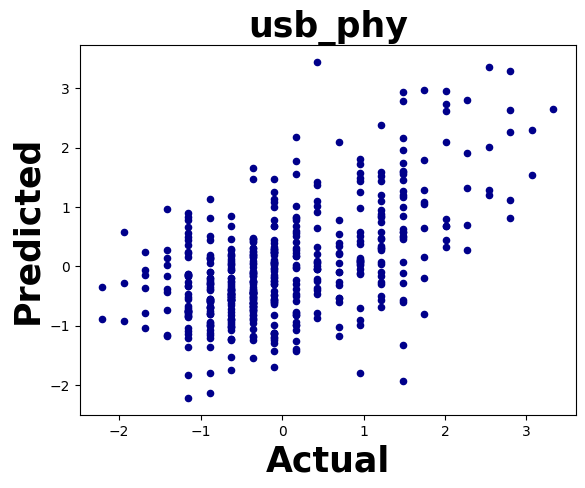}}
        \subfloat[\label{fig:bp_be}]{\includegraphics[width=0.2\columnwidth, valign=c]{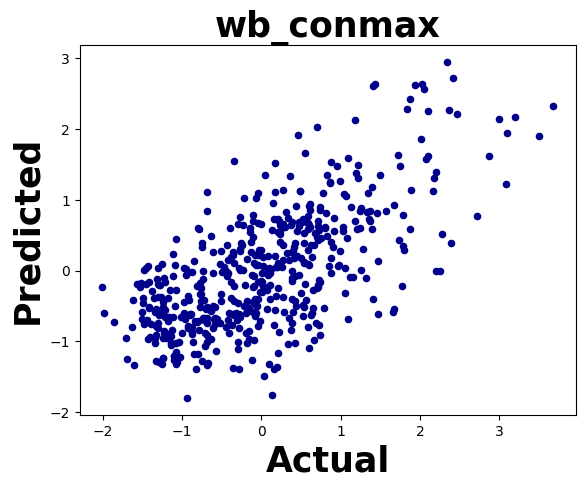}}
   \caption{Net 1 for QoR Task Variant 1 (Unseen Recipe)}
    \label{fig:net1_set1}
\end{figure}

\begin{figure}
    \centering
    \subfloat[\label{fig:ac97_ctrl}]{\includegraphics[width=0.2\columnwidth, valign=c]{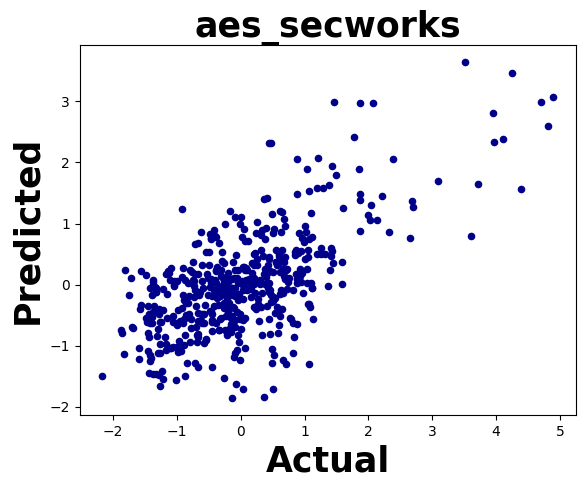}}
    \subfloat[\label{fig:aes_secworks}]{\includegraphics[width=0.2\columnwidth, valign=c]{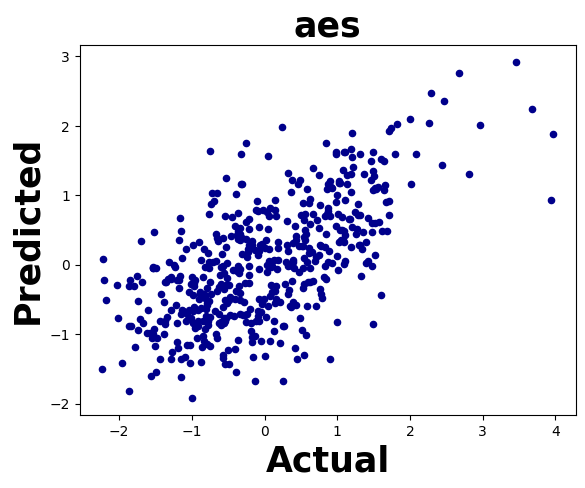}}
    \subfloat[\label{fig:aes_xcrypt}]{\includegraphics[width=0.2\columnwidth, valign=c]{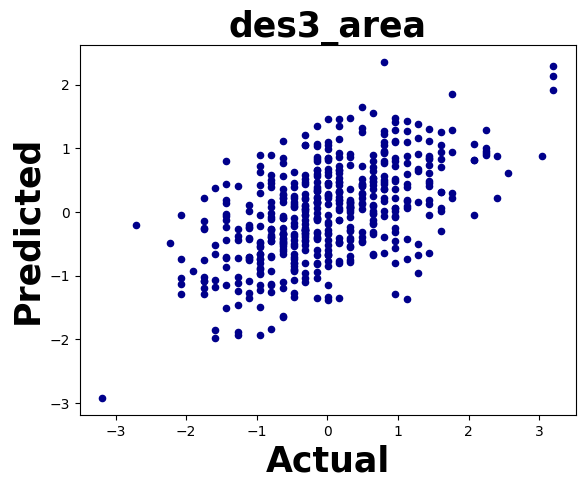}}
    \subfloat[\label{fig:aes}]{\includegraphics[width=0.2\columnwidth, valign=c]{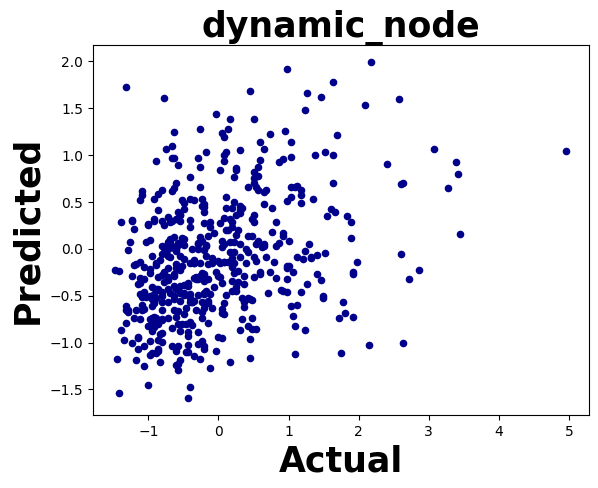}}
    \subfloat[\label{fig:bp_be}]{\includegraphics[width=0.2\columnwidth, valign=c]{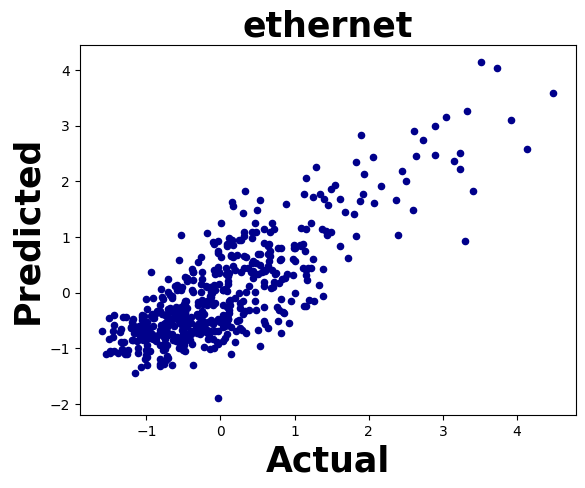}} \\ \vspace*{-0.1in}
    \subfloat[\label{fig:ac97_ctrl}]{\includegraphics[width=0.2\columnwidth, valign=c]{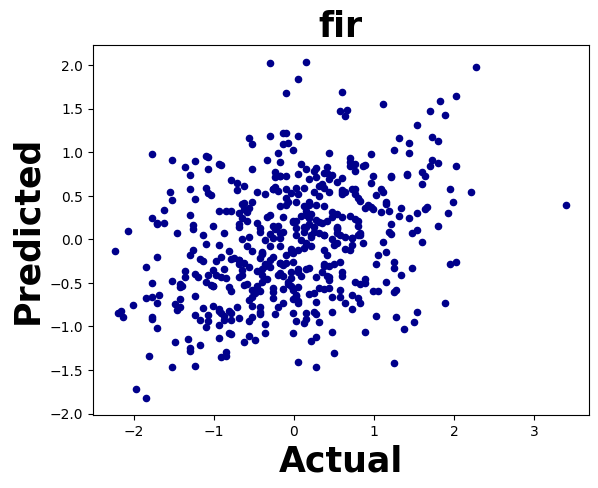}}
    \subfloat[\label{fig:aes_secworks}]{\includegraphics[width=0.2\columnwidth, valign=c]{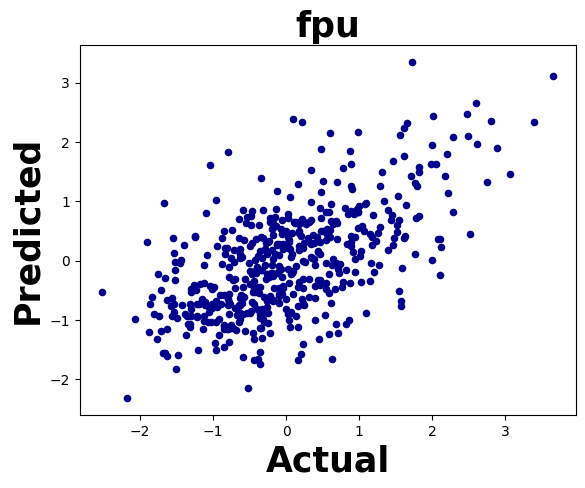}}
    \subfloat[\label{fig:aes_xcrypt}]{\includegraphics[width=0.2\columnwidth, valign=c]{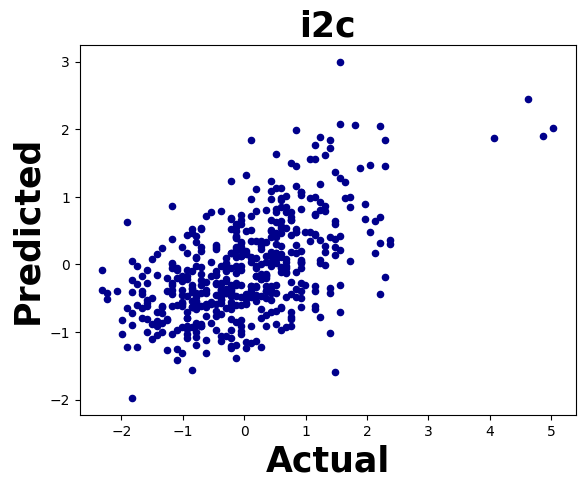}}
    \subfloat[\label{fig:aes}]{\includegraphics[width=0.2\columnwidth, valign=c]{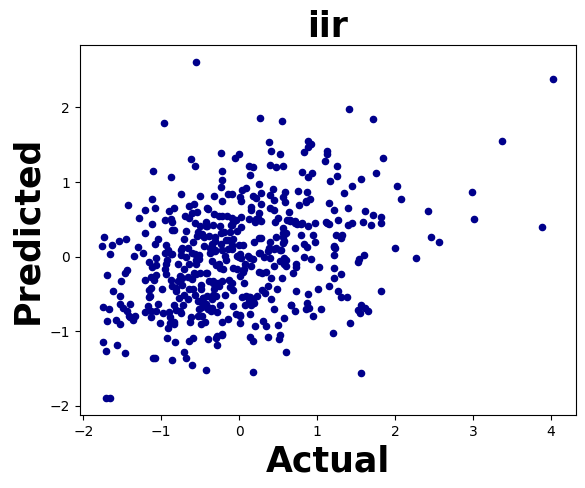}}
    \subfloat[\label{fig:bp_be}]{\includegraphics[width=0.2\columnwidth, valign=c]{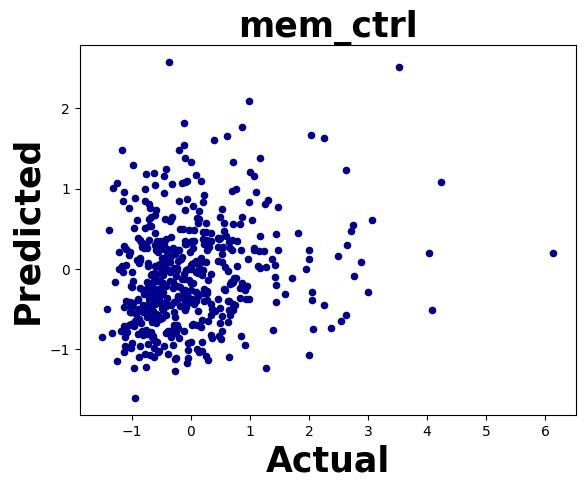}} \\\vspace*{-0.1in}
    \subfloat[\label{fig:ac97_ctrl}]{\includegraphics[width=0.2\columnwidth, valign=c]{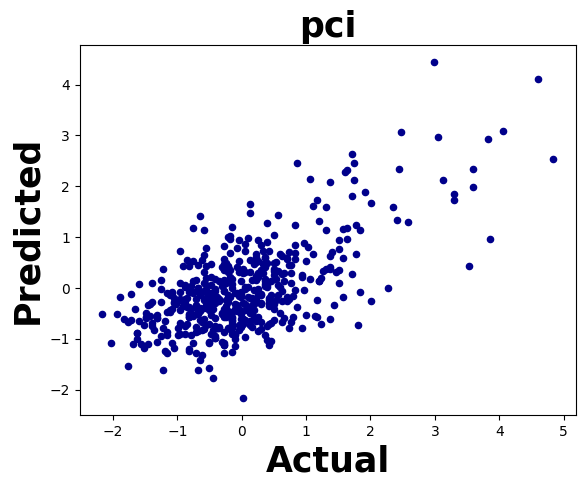}}
    \subfloat[\label{fig:aes_xcrypt}]{\includegraphics[width=0.2\columnwidth, valign=c]{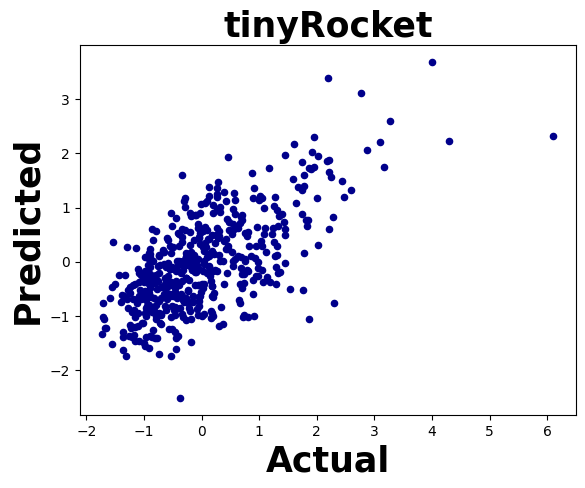}}
    \subfloat[\label{fig:aes}]{\includegraphics[width=0.2\columnwidth, valign=c]{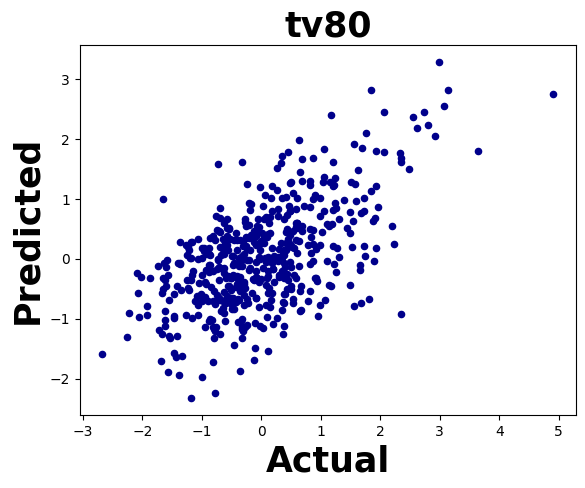}}
    \subfloat[\label{fig:bp_be}]{\includegraphics[width=0.2\columnwidth, valign=c]{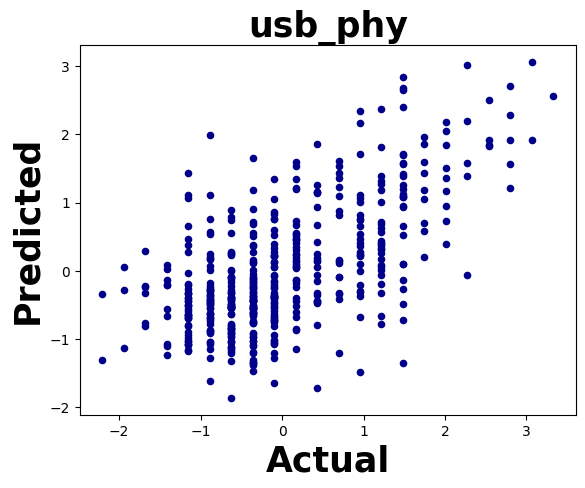}}
        \subfloat[\label{fig:bp_be}]{\includegraphics[width=0.2\columnwidth, valign=c]{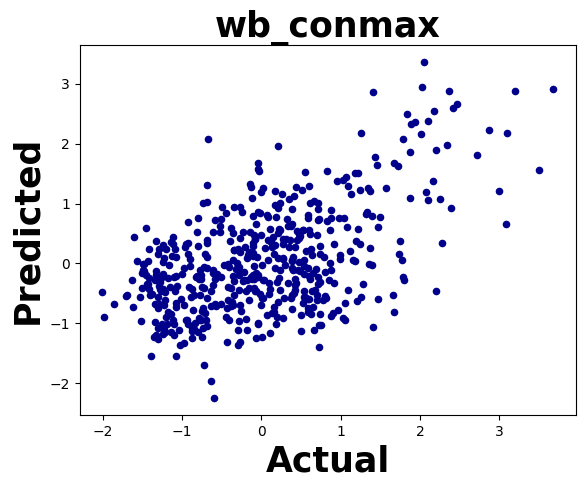}}
   \caption{Net 2 for QoR Task Variant 1 (Unseen Recipe)}
    \label{fig:net2_set1}
\end{figure}

\begin{figure}
    \centering
    \subfloat[\label{fig:ac97_ctrl}]{\includegraphics[width=0.2\columnwidth, valign=c]{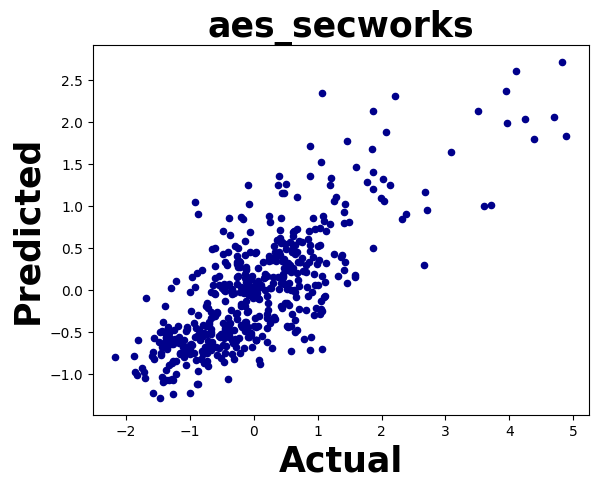}}
    \subfloat[\label{fig:aes_secworks}]{\includegraphics[width=0.2\columnwidth, valign=c]{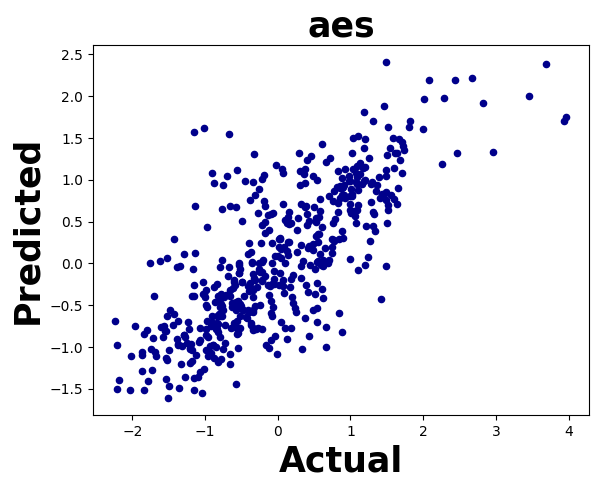}}
    \subfloat[\label{fig:aes_xcrypt}]{\includegraphics[width=0.2\columnwidth, valign=c]{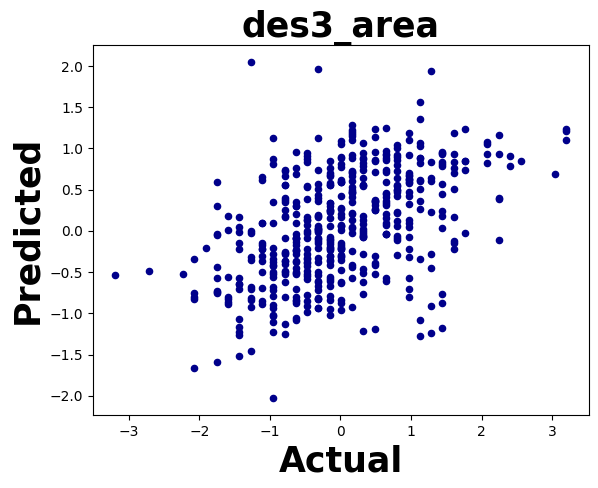}}
    \subfloat[\label{fig:aes}]{\includegraphics[width=0.2\columnwidth, valign=c]{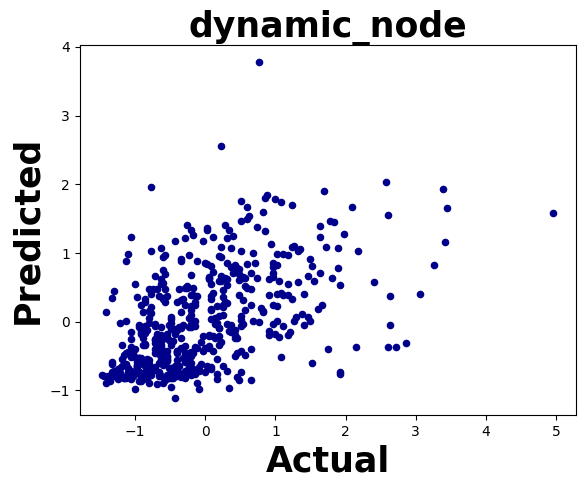}}
    \subfloat[\label{fig:bp_be}]{\includegraphics[width=0.2\columnwidth, valign=c]{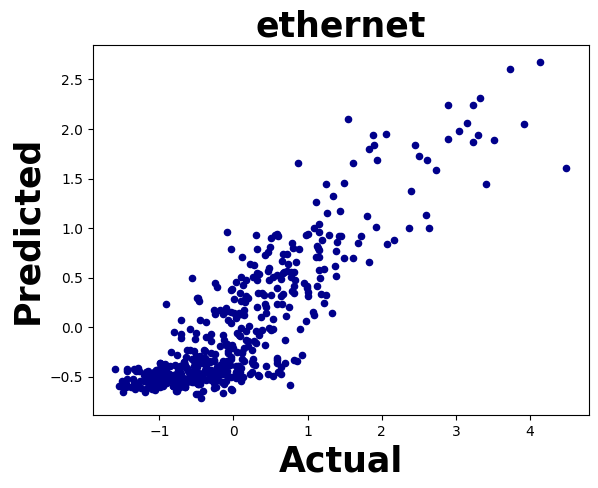}} \\ \vspace*{-0.1in}
    \subfloat[\label{fig:ac97_ctrl}]{\includegraphics[width=0.2\columnwidth, valign=c]{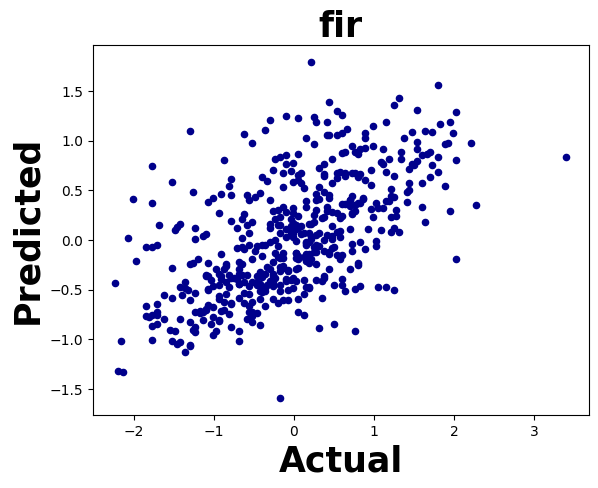}}
    \subfloat[\label{fig:aes_secworks}]{\includegraphics[width=0.2\columnwidth, valign=c]{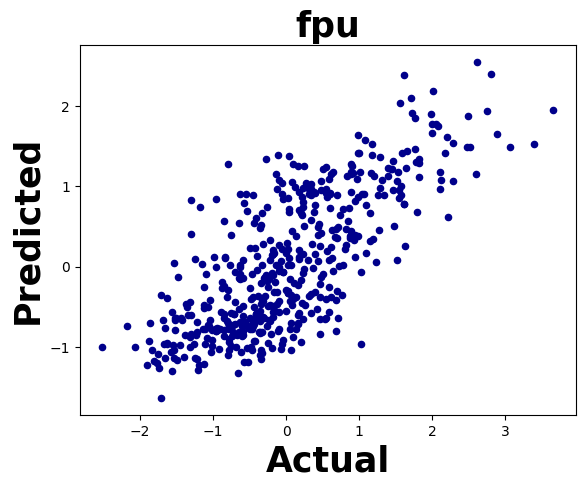}}
    \subfloat[\label{fig:aes_xcrypt}]{\includegraphics[width=0.2\columnwidth, valign=c]{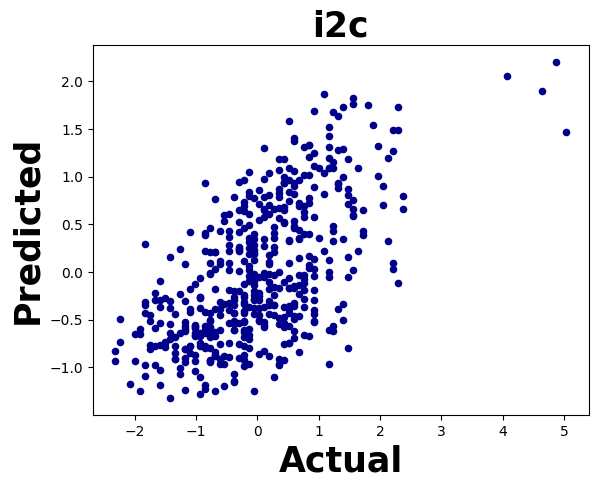}}
    \subfloat[\label{fig:aes}]{\includegraphics[width=0.2\columnwidth, valign=c]{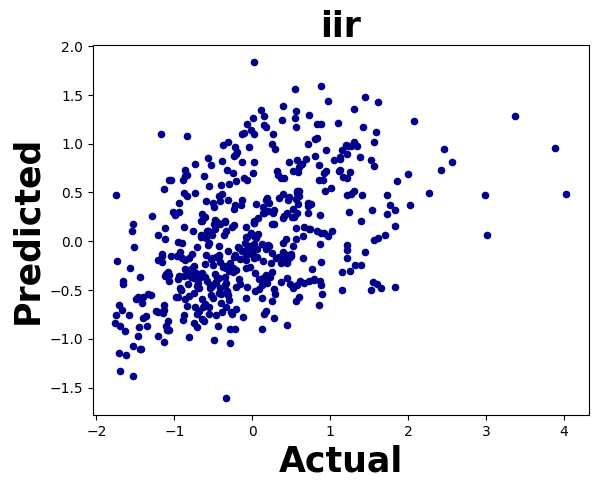}}
    \subfloat[\label{fig:bp_be}]{\includegraphics[width=0.2\columnwidth, valign=c]{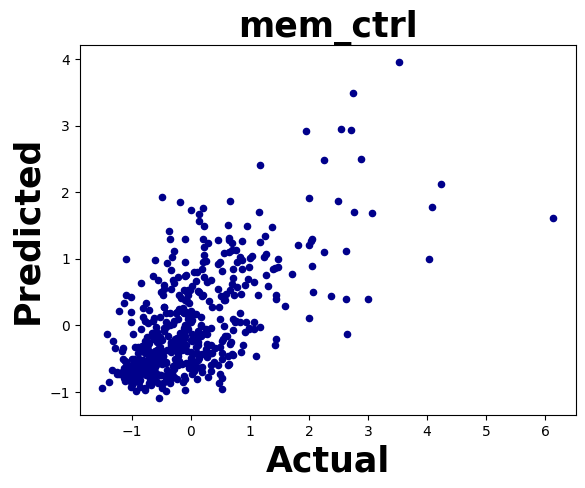}} \\\vspace*{-0.1in}
    \subfloat[\label{fig:ac97_ctrl}]{\includegraphics[width=0.2\columnwidth, valign=c]{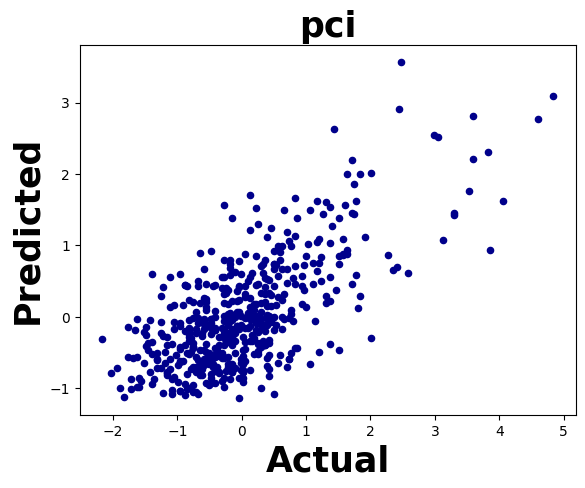}}
    \subfloat[\label{fig:aes_xcrypt}]{\includegraphics[width=0.2\columnwidth, valign=c]{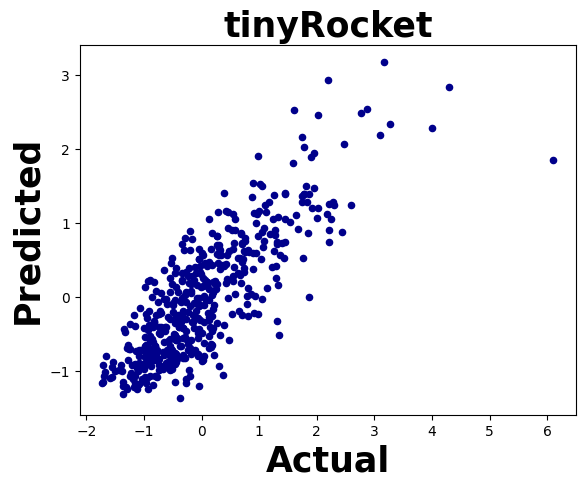}}
    \subfloat[\label{fig:aes}]{\includegraphics[width=0.2\columnwidth, valign=c]{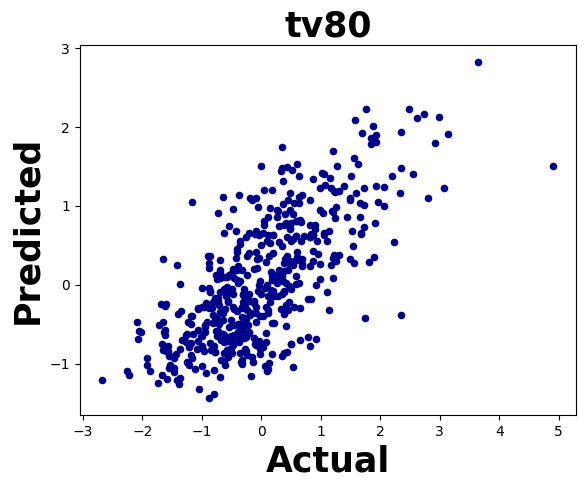}}
    \subfloat[\label{fig:bp_be}]{\includegraphics[width=0.2\columnwidth, valign=c]{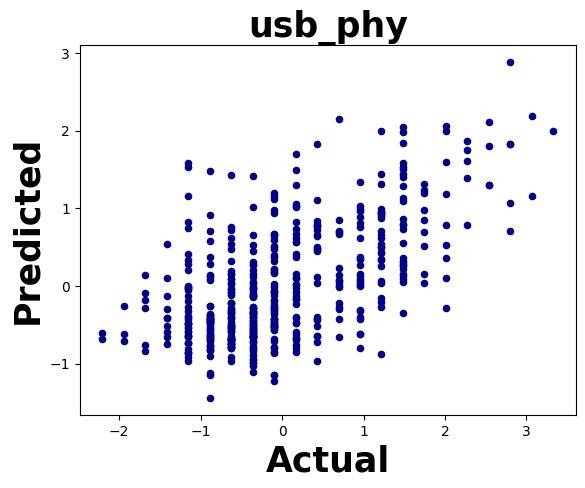}}
        \subfloat[\label{fig:bp_be}]{\includegraphics[width=0.2\columnwidth, valign=c]{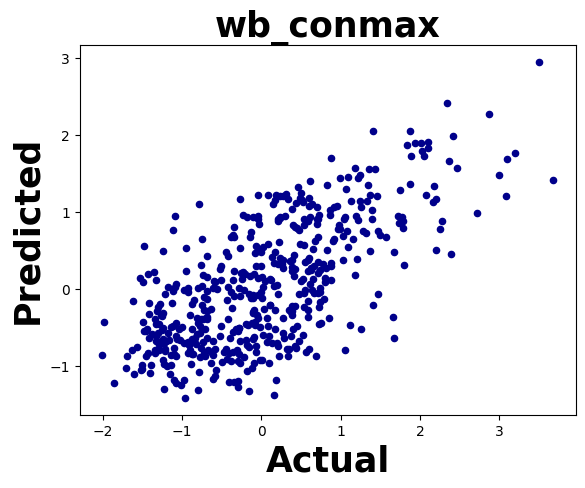}}
   \caption{Net 3 for QoR Task Variant 1 (Unseen Recipe)}
   \label{fig:net3_set1}
\end{figure}

%%%%%%%%%%%%%%%%%%%%% END of SET 1 Scatter plots %%%%%%%%%%%%%%%%%%%%%%%%%%%%

%%%%%%%%%%%%%%%%%%%%% END of SET 2 Scatter plots %%%%%%%%%%%%%%%%%%%%%%%%%%%%

\begin{figure}
    \centering
    \subfloat[\label{fig:ac97_ctrl}]{\includegraphics[width=0.2\columnwidth, valign=c]{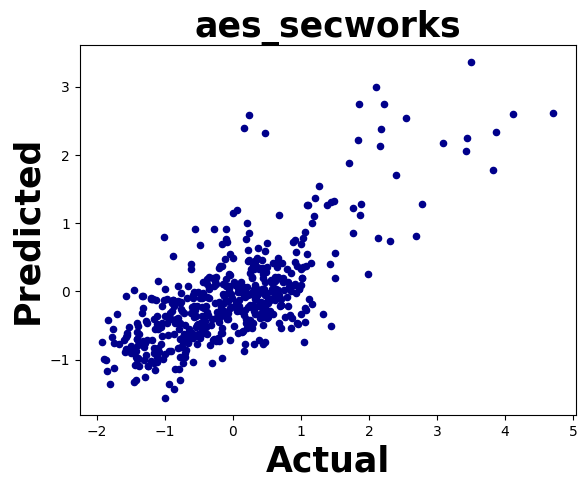}}
    \subfloat[\label{fig:aes_secworks}]{\includegraphics[width=0.2\columnwidth, valign=c]{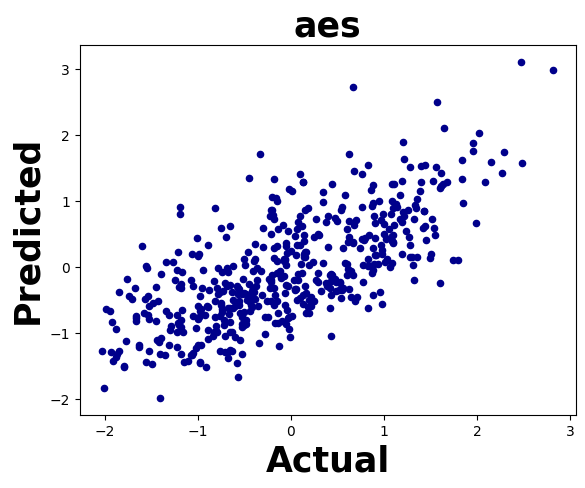}}
    \subfloat[\label{fig:aes_xcrypt}]{\includegraphics[width=0.2\columnwidth, valign=c]{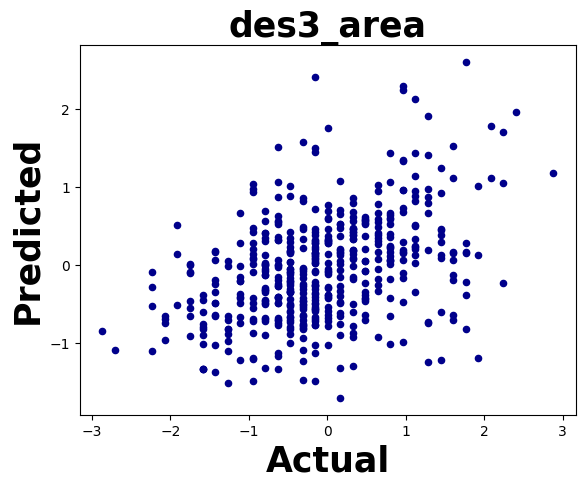}}
    \subfloat[\label{fig:aes}]{\includegraphics[width=0.2\columnwidth, valign=c]{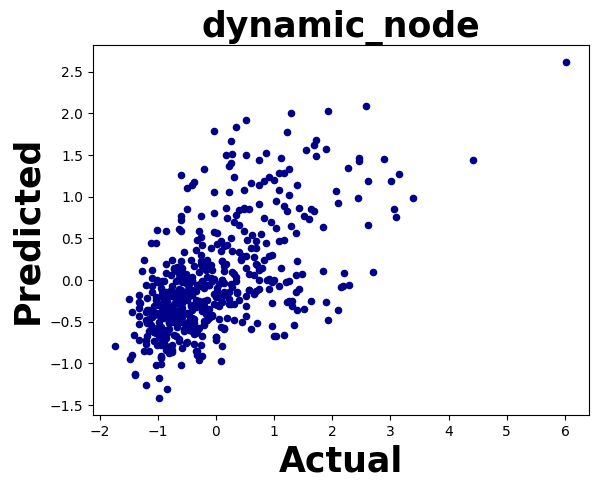}}
    \subfloat[\label{fig:bp_be}]{\includegraphics[width=0.2\columnwidth, valign=c]{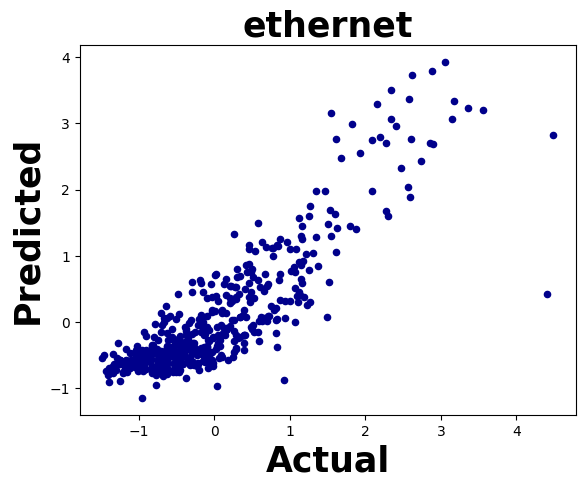}} \\ \vspace*{-0.1in}
    \subfloat[\label{fig:ac97_ctrl}]{\includegraphics[width=0.2\columnwidth, valign=c]{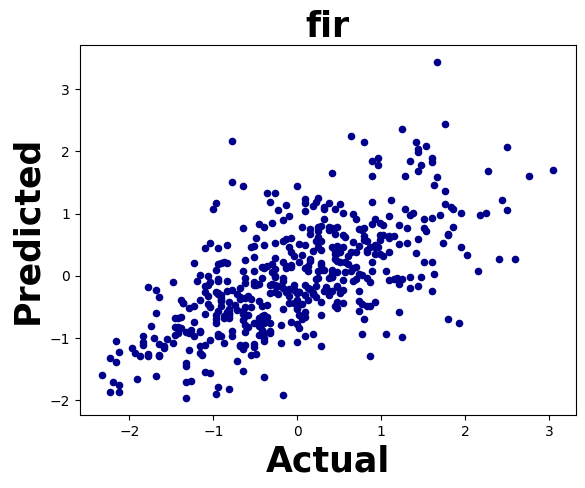}}
    \subfloat[\label{fig:aes_secworks}]{\includegraphics[width=0.2\columnwidth, valign=c]{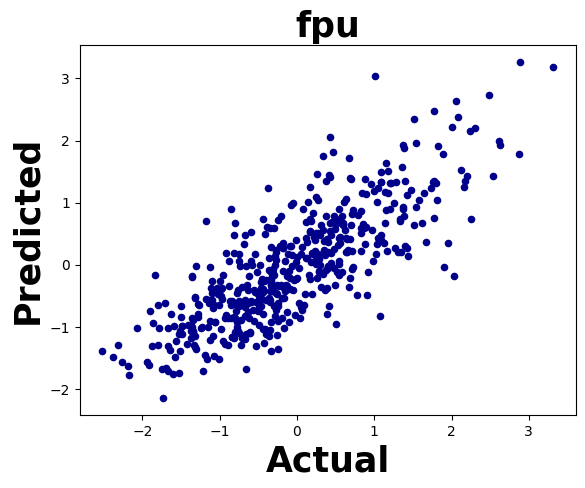}}
    \subfloat[\label{fig:aes_xcrypt}]{\includegraphics[width=0.2\columnwidth, valign=c]{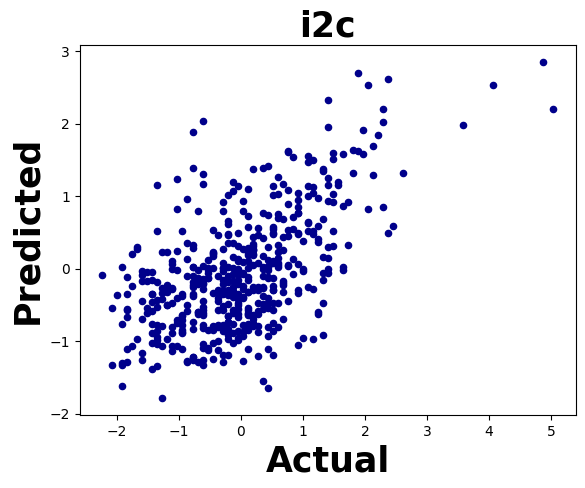}}
    \subfloat[\label{fig:aes}]{\includegraphics[width=0.2\columnwidth, valign=c]{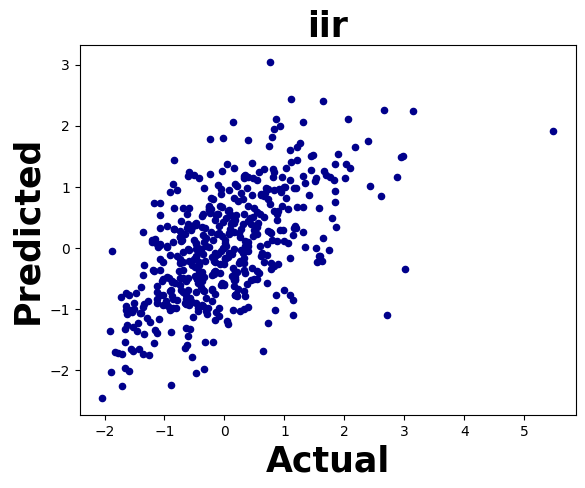}}
    \subfloat[\label{fig:bp_be}]{\includegraphics[width=0.2\columnwidth, valign=c]{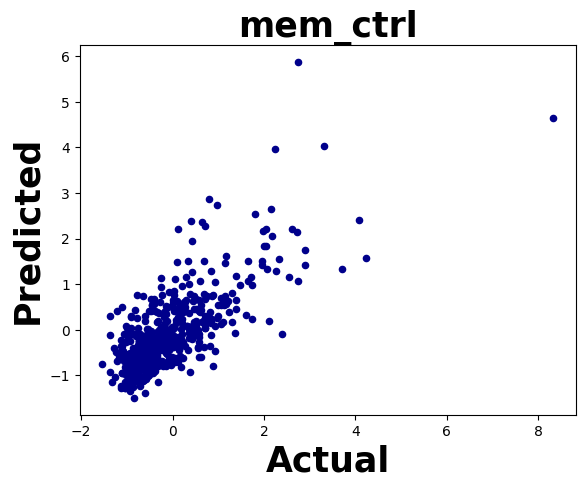}} \\\vspace*{-0.1in}
    \subfloat[\label{fig:ac97_ctrl}]{\includegraphics[width=0.2\columnwidth, valign=c]{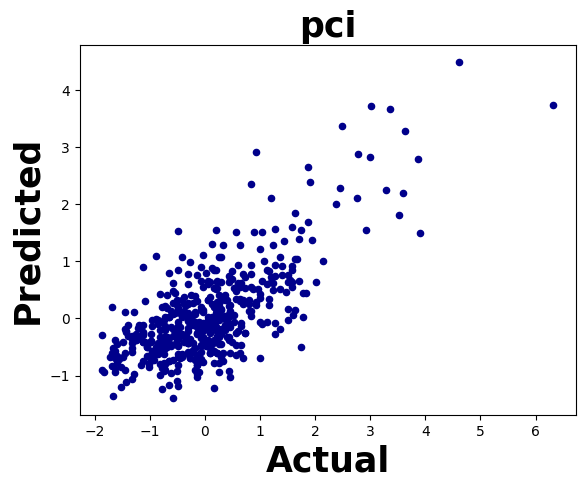}}
    \subfloat[\label{fig:aes_xcrypt}]{\includegraphics[width=0.2\columnwidth, valign=c]{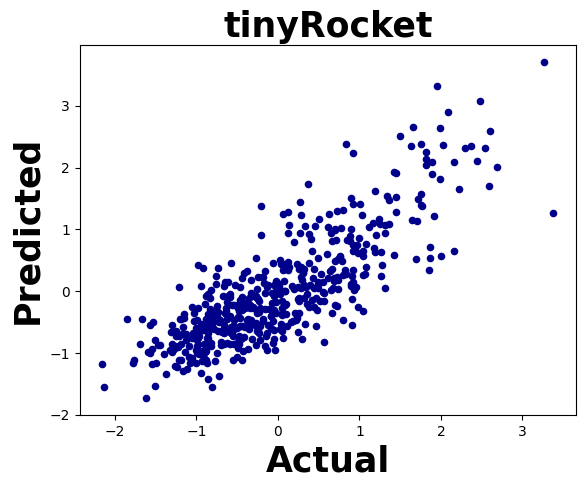}}
    \subfloat[\label{fig:aes}]{\includegraphics[width=0.2\columnwidth, valign=c]{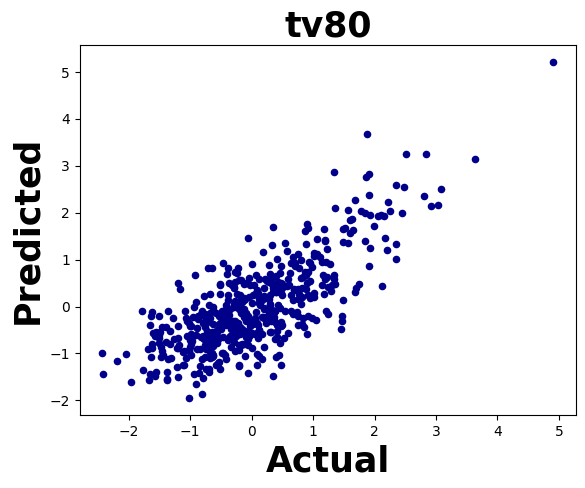}}
    \subfloat[\label{fig:bp_be}]{\includegraphics[width=0.2\columnwidth, valign=c]{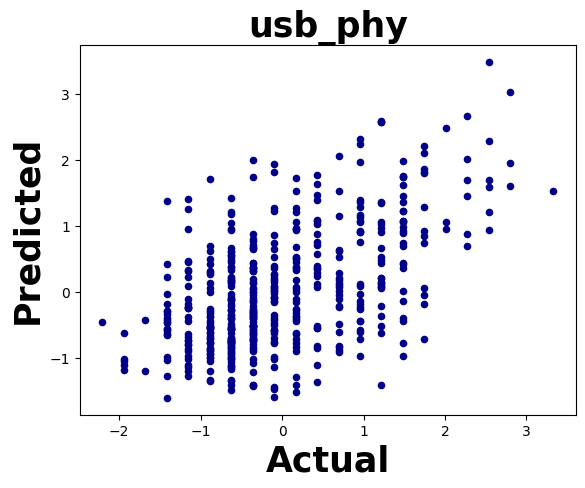}}
        \subfloat[\label{fig:bp_be}]{\includegraphics[width=0.2\columnwidth, valign=c]{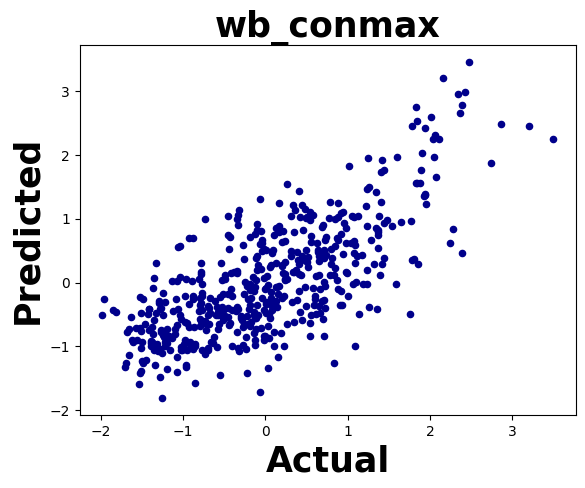}}
   \caption{Net 1 for QoR Task Variant 3 (Unseen IP-Recipe combination)}
    \label{fig:net1_set3}
\end{figure}

\begin{figure}
    \centering
    \subfloat[\label{fig:ac97_ctrl}]{\includegraphics[width=0.2\columnwidth, valign=c]{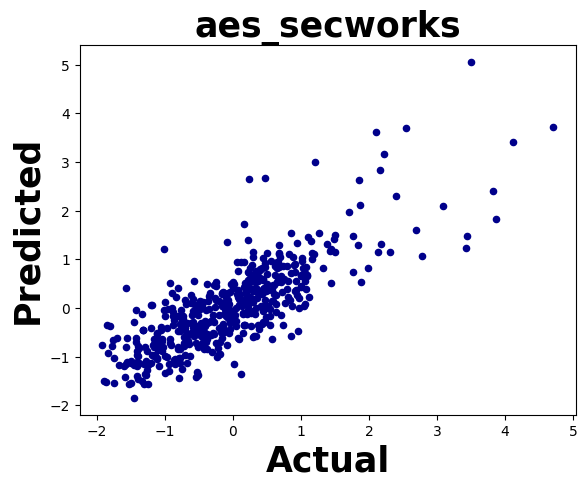}}
    \subfloat[\label{fig:aes_secworks}]{\includegraphics[width=0.2\columnwidth, valign=c]{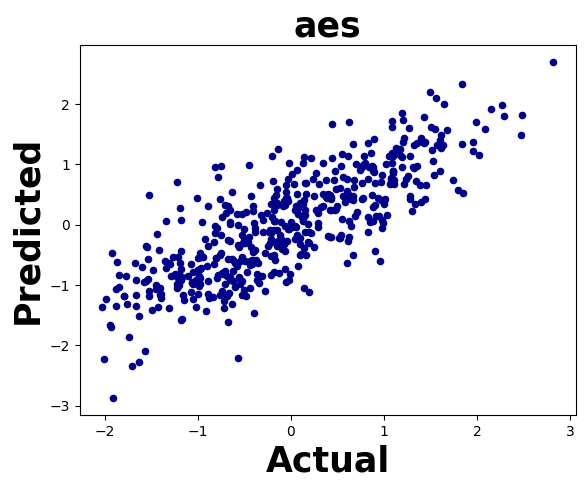}}
    \subfloat[\label{fig:aes_xcrypt}]{\includegraphics[width=0.2\columnwidth, valign=c]{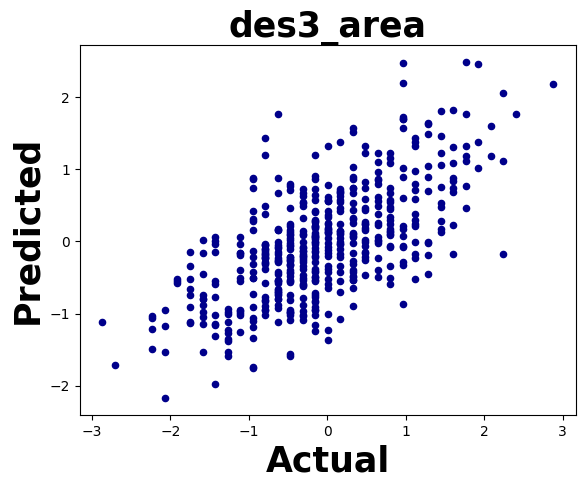}}
    \subfloat[\label{fig:aes}]{\includegraphics[width=0.2\columnwidth, valign=c]{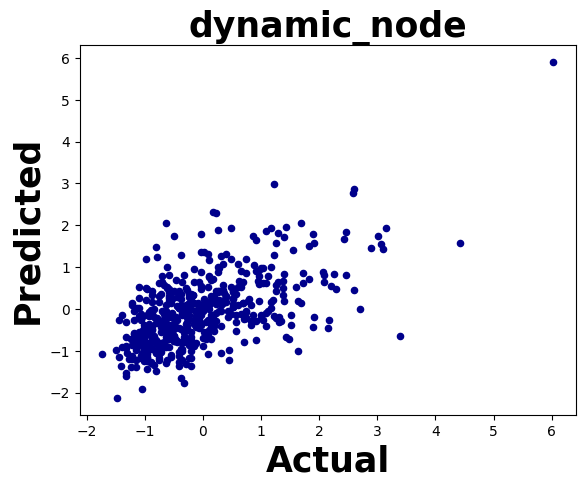}}
    \subfloat[\label{fig:bp_be}]{\includegraphics[width=0.2\columnwidth, valign=c]{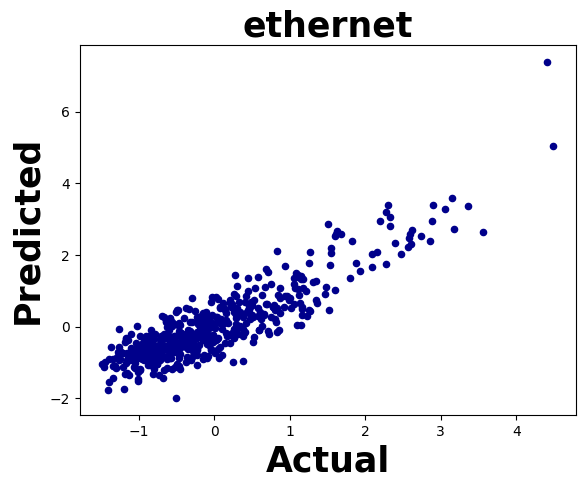}} \\ \vspace*{-0.1in}
    \subfloat[\label{fig:ac97_ctrl}]{\includegraphics[width=0.2\columnwidth, valign=c]{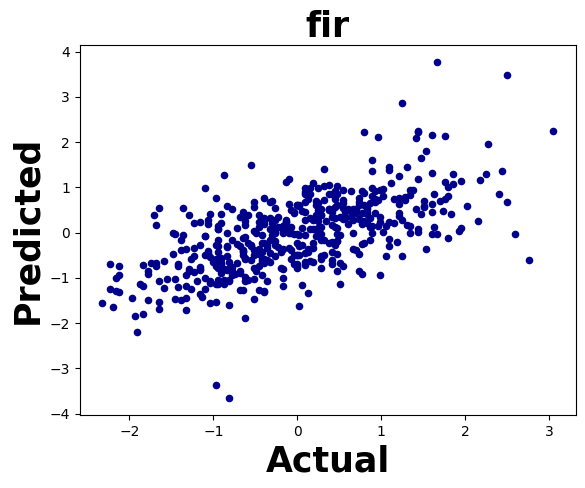}}
    \subfloat[\label{fig:aes_secworks}]{\includegraphics[width=0.2\columnwidth, valign=c]{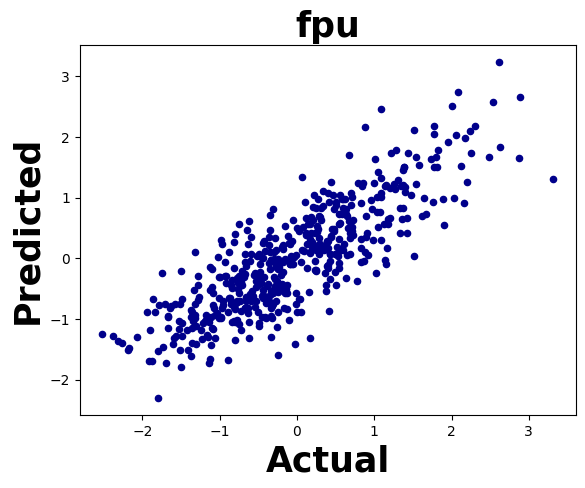}}
    \subfloat[\label{fig:aes_xcrypt}]{\includegraphics[width=0.2\columnwidth, valign=c]{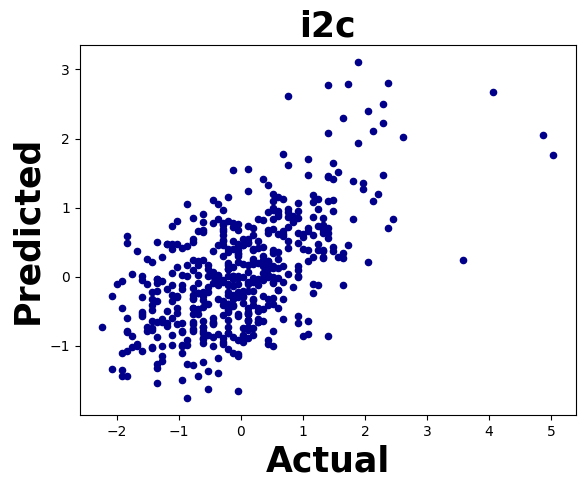}}
    \subfloat[\label{fig:aes}]{\includegraphics[width=0.2\columnwidth, valign=c]{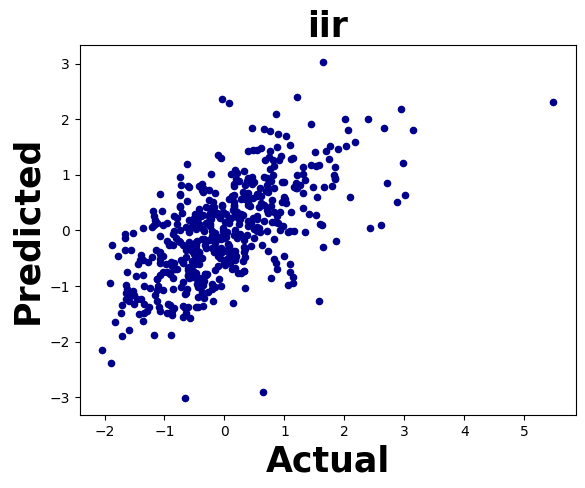}}
    \subfloat[\label{fig:bp_be}]{\includegraphics[width=0.2\columnwidth, valign=c]{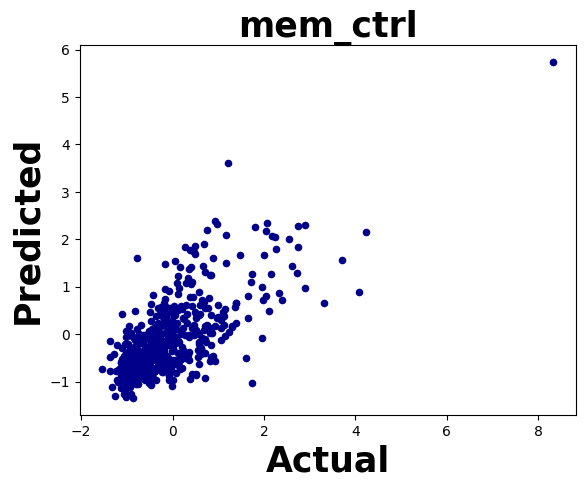}} \\\vspace*{-0.1in}
    \subfloat[\label{fig:ac97_ctrl}]{\includegraphics[width=0.2\columnwidth, valign=c]{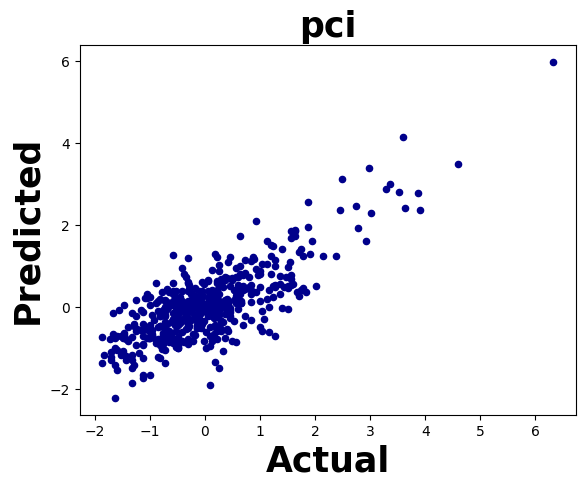}}
    \subfloat[\label{fig:aes_xcrypt}]{\includegraphics[width=0.2\columnwidth, valign=c]{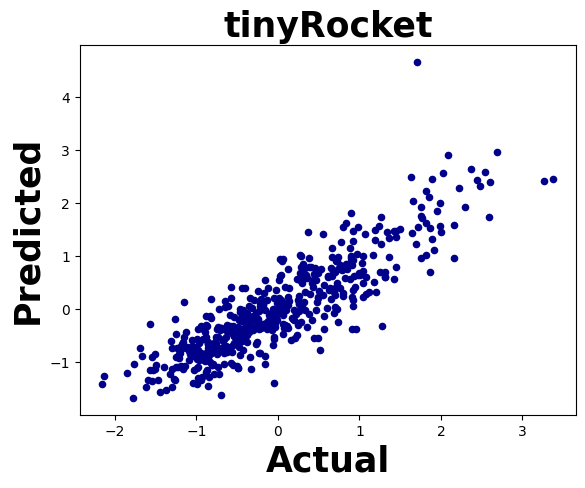}}
    \subfloat[\label{fig:aes}]{\includegraphics[width=0.2\columnwidth, valign=c]{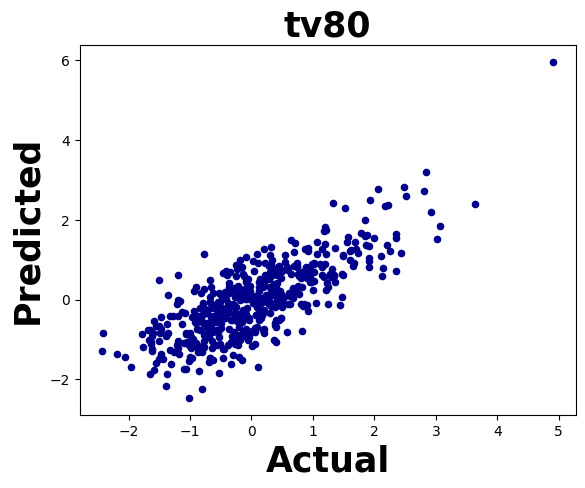}}
    \subfloat[\label{fig:bp_be}]{\includegraphics[width=0.2\columnwidth, valign=c]{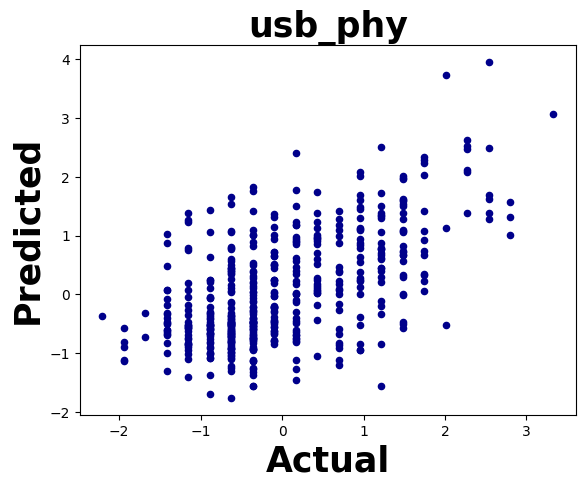}}
        \subfloat[\label{fig:bp_be}]{\includegraphics[width=0.2\columnwidth, valign=c]{figures/set1/syn16/scatterPlot_test_wb_conmax.png}}
   \caption{Net 2 for QoR Task Variant 3 (Unseen IP-Recipe combination)}
    \label{fig:net2_set3}
\end{figure}

\begin{figure}
    \centering
    \subfloat[\label{fig:ac97_ctrl}]{\includegraphics[width=0.2\columnwidth, valign=c]{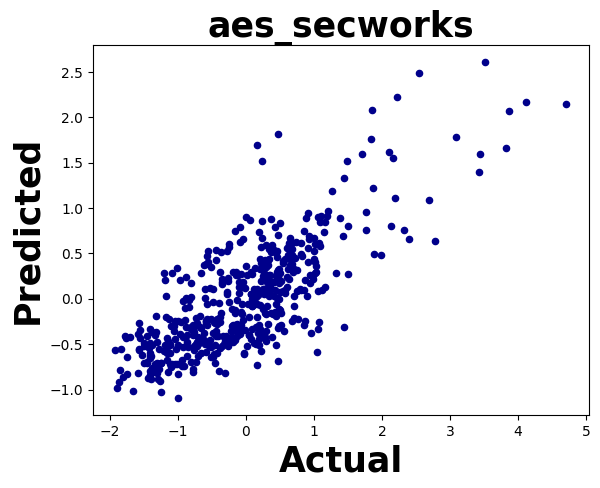}}
    \subfloat[\label{fig:aes_secworks}]{\includegraphics[width=0.2\columnwidth, valign=c]{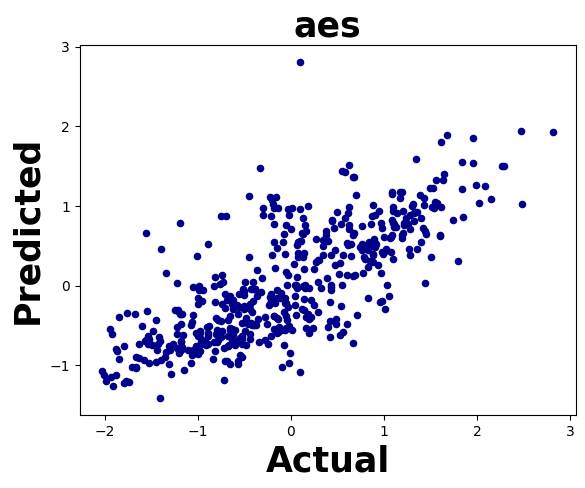}}
    \subfloat[\label{fig:aes_xcrypt}]{\includegraphics[width=0.2\columnwidth, valign=c]{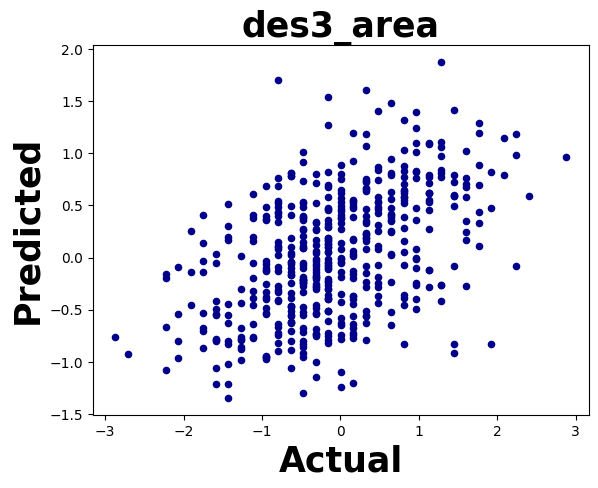}}
    \subfloat[\label{fig:aes}]{\includegraphics[width=0.2\columnwidth, valign=c]{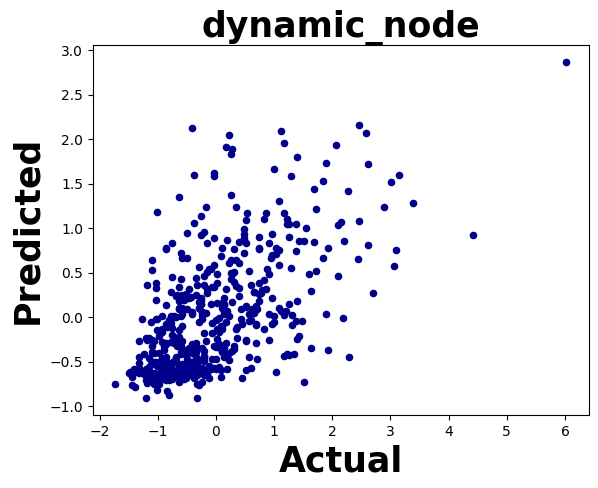}}
    \subfloat[\label{fig:bp_be}]{\includegraphics[width=0.2\columnwidth, valign=c]{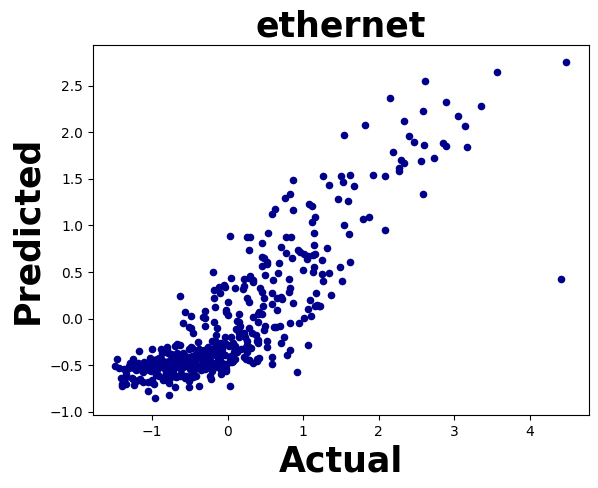}} \\ \vspace*{-0.1in}
    \subfloat[\label{fig:ac97_ctrl}]{\includegraphics[width=0.2\columnwidth, valign=c]{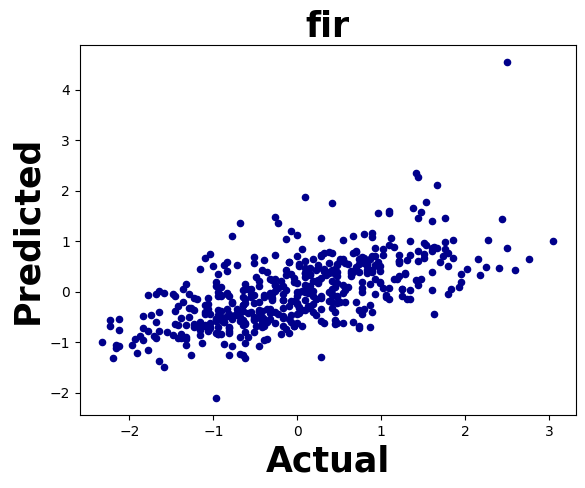}}
    \subfloat[\label{fig:aes_secworks}]{\includegraphics[width=0.2\columnwidth, valign=c]{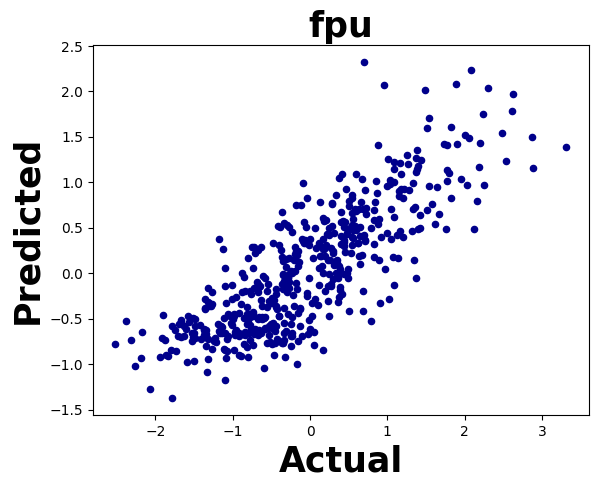}}
    \subfloat[\label{fig:aes_xcrypt}]{\includegraphics[width=0.2\columnwidth, valign=c]{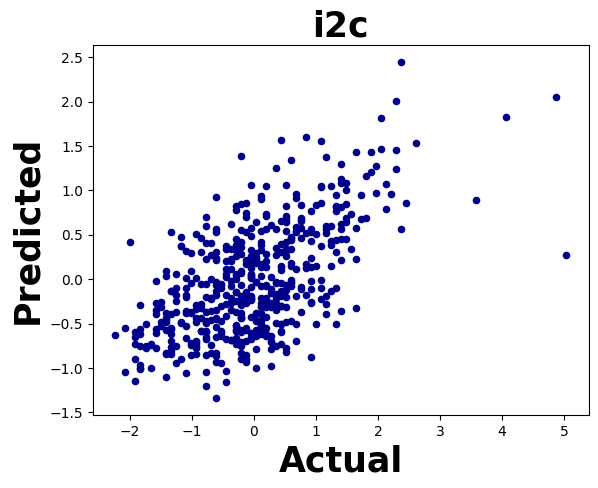}}
    \subfloat[\label{fig:aes}]{\includegraphics[width=0.2\columnwidth, valign=c]{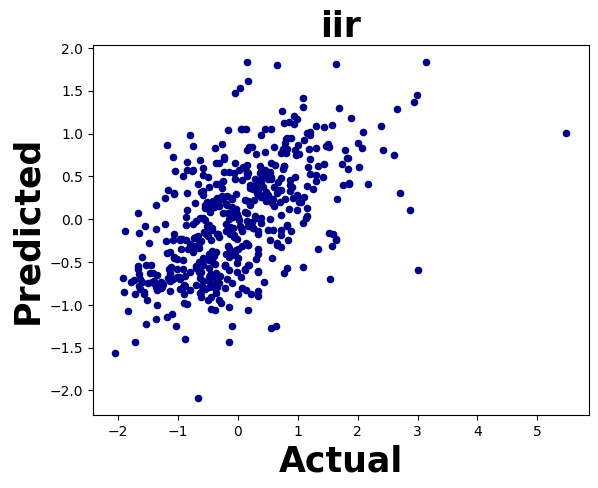}}
    \subfloat[\label{fig:bp_be}]{\includegraphics[width=0.2\columnwidth, valign=c]{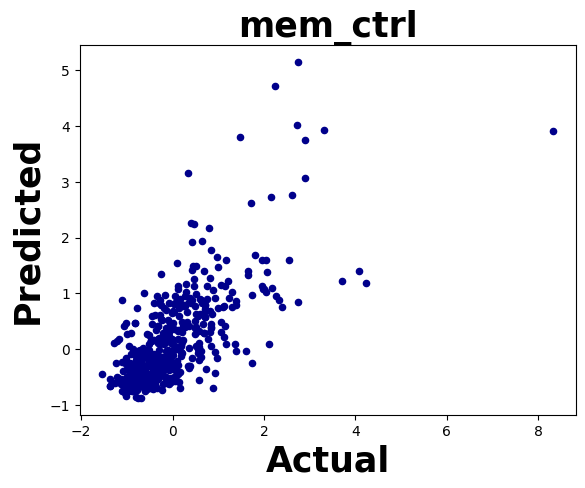}} \\\vspace*{-0.1in}
    \subfloat[\label{fig:ac97_ctrl}]{\includegraphics[width=0.2\columnwidth, valign=c]{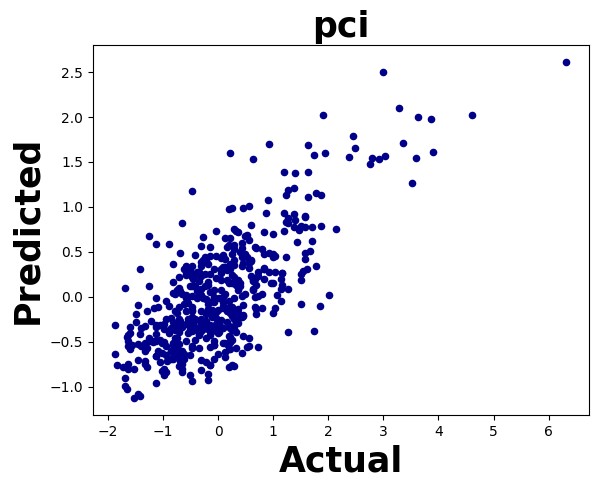}}
    \subfloat[\label{fig:aes_xcrypt}]{\includegraphics[width=0.2\columnwidth, valign=c]{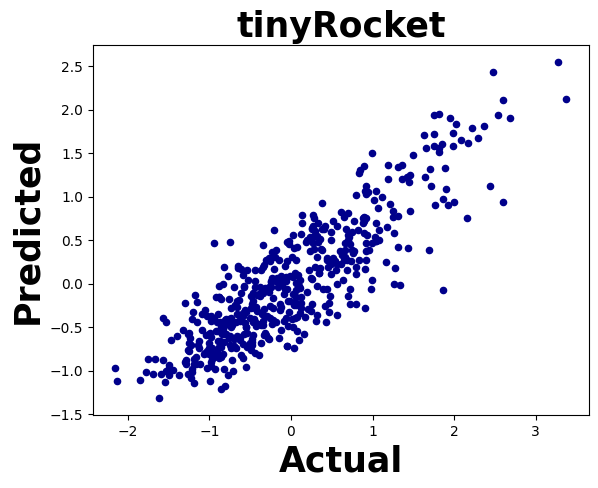}}
    \subfloat[\label{fig:aes}]{\includegraphics[width=0.2\columnwidth, valign=c]{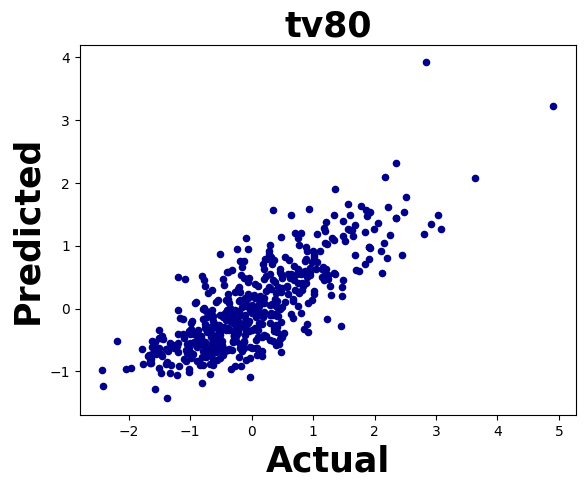}}
    \subfloat[\label{fig:bp_be}]{\includegraphics[width=0.2\columnwidth, valign=c]{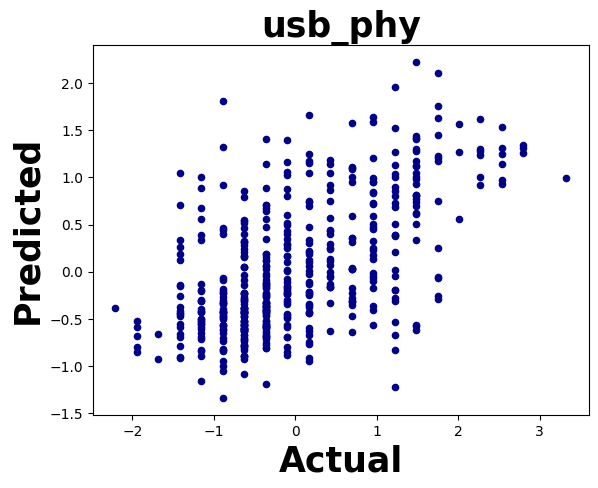}}
        \subfloat[\label{fig:bp_be}]{\includegraphics[width=0.2\columnwidth, valign=c]{figures/set1/syn20/scatterPlot_test_wb_conmax.png}}
   \caption{Net 3 for QoR Task Variant 3 (Unseen IP-Recipe combination)}
   \label{fig:net3_set3}
\end{figure}

\begin{table*}[]
\centering
\resizebox{0.6\columnwidth}{!}{%
\begin{tabular}{@{}cccc@{}}
\toprule
\multirow{2}{*}{QoR Task} & \multicolumn{3}{c}{Test MSE on baseline networks}\\
\cmidrule(lr){2-4}
& Net1 & Net2 & Net3 \\ \midrule
Variant1 & $0.648\pm0.05$  & $0.815\pm0.02$  & $\mathbf{0.579\pm0.02}$ \\%& $0.648\pm0.05$ & $0.815\pm0.02$ & \textbf{$0.579\pm0.02$} \\
Variant2 & $10.59\pm2.78$ & $1.236\pm0.15$ & $\mathbf{1.47\pm0.14}$ \\
Variant3 & $0.588\pm0.04$ & $0.538\pm0.01$ & $\mathbf{0.536\pm0.03}$ \\

\bottomrule
\end{tabular}
}
\caption{Benchmarking \ac{GCN} models for QoR prediction tasks}
\label{tab:qorsynthesisrecipe}
\end{table*}

\end{document}